\documentclass[5p,sort,compress]{elsarticle}
\usepackage{fancyhdr}
\usepackage{amsfonts}
\usepackage{url}
\usepackage{amsmath}
\usepackage[labelfont=bf,justification=raggedright,singlelinecheck=false]{caption}
\usepackage{array}
\usepackage{multirow}
\usepackage{graphicx}
\usepackage{amsmath}
\usepackage{algorithm}
\usepackage[noend]{algpseudocode}
\usepackage{pdflscape}
\usepackage{rotating}
\usepackage{wrapfig}
\usepackage[inline, shortlabels]{enumitem}
\usepackage{graphbox}
\usepackage{mathtools}
\usepackage{amsmath}

\newcolumntype{v}{>{\centering\arraybackslash}m{.18\linewidth} }
\usepackage{hyperref}
\hypersetup{
	colorlinks=true,
	linkcolor=blue,
	filecolor=magenta, 
	urlcolor=cyan,
}
\usepackage[labelfont=bf, justification=justified, format=plain]{caption} 
\begin{document}
	\journal{Arxiv}
	\title{EYNet: Extended YOLO for Airport Detection in Remote Sensing Images}
	\author[fum1]{Hengameh Mirhajianmoghadam}
	\ead{hengameh.mirhajian@mail.um.ac.ir}
	\author[fum2,MVLAB]{Behrouz Bolourian Haghighi\corref{cor1}}
	\ead{b.bolourian@mail.um.ac.ir}
	\cortext[cor1]{Corresponding author}
 	\address[fum1]{Electrical Engineering Department, Ferdowsi University of Mashhad, Mashhad, Iran }
	\address[fum2]{Computer Engineering Department, Ferdowsi University of Mashhad, Mashhad, Iran }
	\address[MVLAB]{Machine Vision Laboratory, Ferdowsi University of Mashhad, Mashhad, Iran}
	\begin{abstract}
	
	Nowadays, airport detection in remote sensing images has attracted considerable attention due to its strategic role in civilian and military scopes. In particular, uncrewed and operated aerial vehicles must immediately detect safe areas to land in emergencies. The previous schemes suffered from various aspects, including complicated backgrounds, scales, and shapes of the airport. Meanwhile, the rapid action and accuracy of the method are confronted with significant concerns. Hence, this study proposes an effective scheme by extending YOLOV3 and ShearLet transform. In this way, MobileNet and ResNet18, with fewer layers and parameters retrained on a similar dataset, are parallelly trained as base networks. According to airport geometrical characteristics, the ShearLet filters with different scales and directions are considered in the first convolution layers of ResNet18 as a visual attention mechanism. Besides, the major extended in YOLOV3 concerns the detection Sub-Networks with novel structures which boost object expression ability and training efficiency. In addition, novel augmentation and negative mining strategies are presented to significantly increase the localization phase's performance. The experimental results on the DIOR dataset reveal that the framework reliably detects different types of airports in a varied area and acquires robust results in complex scenes compared to traditional YOLOV3 and state-of-the-art schemes.

	\end{abstract}
	\maketitle
	\section{Introduction}	
	Remote sensing images are a way to acquire information about an object without direct contact. With the rapid pace of advancement in remote sensing technologies, providing and collecting numerous remote sensing images with superior quality, the high spatial or spectral resolution is effortlessly feasible\cite{ref1}. These images have been widely used in real-world applications; Hence, their interpretation plays a prominent role in this regard\cite{ref2}. Generally, remote object detection is the hot research topic in computer vision\cite{ref3}, which includes localization and identification of one or more objects\cite{ref4}. This topic has attracted many researcher's attention due to its importance in a vast range of practical applications, such as intelligent monitoring, urban planning, geological hazard detection, traffic management, precision agriculture, Geographic Information System (GIS) updating, searching, rescuing, environmental monitoring and other civil or military applications\cite{ref5, ref6}.					
	
	Satisfying recent advancements in object detection methods, detecting objects including cars, airplanes, ships, storage tanks, bridges, roads, and airports, in large-scale satellite imagery has been successful\cite{ref4}. However, tiny object detection in remote sensing image interpretation research is still a challenging task\cite{ref7}. On the other hand, the object detection phase faces significant challenges due to the different characteristics of remote images compared to natural ones. The accuracy and efficiency of methods are drastically reduced due to multiscale and multidirectional objects on a large scale. Also, the complex backgrounds and various interferences, such as illumination, occlusion, and geometric deformation, significantly challenge the detection schemes\cite{ref8}.
	
	Generally, object detection techniques can be categorized into two major groups: traditional machine learning-based and deep learning methods. This task in traditional methods includes two main steps: feature extraction and object classification\cite{ref4}. The common techniques used in the feature extraction phase were Histogram of Oriented Gradients(HOG), Bag-of-Words(BoW), and Sparse Representation(SR)-based features. On the other hand, Support Vector Machine(SVM), K-Nearest-Neighbor(KNN), and Sparse Representation-based Classification(SRC)\cite{ref10} were the principal algorithms in the classification phase. In recent years, end-to-end deep learning techniques have been extensively employed in computer vision applications because of their ability to extract features from big data automatically\cite{ref11}. With the significant breakthroughs made by deep learning methods in object detection, this field has experienced rapid development. 
	
	The usage of Deep Neural Network (DNN) as an object detecting method commenced in 2014 by proposing a Region-based Convolutional Neural Networks(R-CNN)\cite{ref12}. Overall, deep learning-based methods can be grouped into two main categories: one-stage and two-stage detectors\cite{ref1}.
	
	R-CNN is a two-stage detector\cite{ref12}, which extracts boxes or regions containing objects from an image by using selective research. Then Conventional Neural Network (CNN) generates a feature vector, and SVM is trained for classification\cite{ref6}. A Spatial Pyramid Pooling Network (SPP-Net)\cite{ref13} is proposed to improve the running time. Unlike R-CNN, it can take multiscale input by adding the SPP layer to CNN and extracting feature maps from the original image only once. 
	
	Fast R-CNN\cite{ref14} and Faster R-CNN\cite{ref15} are proposed to overcome multi-stage training and fine-tuning only fully connected layers. In these two methods, the R-CNN's structure adopts the advantages of SSP-Net. The significant difference between these methods is finding regions, selective search vs. Region Proposal Network (RPN) for Fast R-CNN and Faster R-CNN, respectively. Faster R-CNN is the first end-to-end object detection method\cite{ref6}. These object detection methods have two stages; First, generating regions of interest, second, classification in the second stage\cite{ref5}. These methods suffer from high computational costs and time-consuming time despite the high accuracy.
	
	For passing through this bottleneck, one-stage detectors were proposed for the first time in 2016\cite{ref16}. You Only Look Once (YOLO) model is the first attempt in the deep learning era to have a faster and simpler object detection algorithm. It is considered a fast real-time object detector due to predicting without any region proposal module. Although this property makes the detection task faster, the accuracy reduces compared to two-stage object detection methods. The later proposed methods, such as the subsequent versions of YOLO and Single Shot Multi-Box Detector (SSD)\cite{ref17} have relieved this shortage. SSD introduces multi-reference and multi-resolution detection techniques to increase the accuracy of one-stage object detection methods\cite{ref5}. Numerous deep learning techniques are proposed to further improve object detection algorithms in terms of the speed and precision of the detection process.
	
	As mentioned, unlike substantial advances in object detection in natural scenes, object detection from remote sensing images is still challenging. Particularly, detecting airports from remote sensing images is more complicated than others due to their specific characteristics. Airport detection has recently been gaining significant attention in this field. The importance of airports is because of their civil and military applications, such as aircraft takeoff and landing, communication, transportation, and energy supply\cite{ref18}. Having complexified and cluttered backgrounds with highly changing surroundings, multiple scales, similarity interference, and the diversity of shapes and illumination intensities are involved factors in the difficulty of airports detection\cite{ref19}. The following sub-section will briefly present the previous method's goal, superiority, weakness, and novelty in this field. 
		\subsection{Literature Review}
		Various traditional approaches have been proposed in airport detection from remote sensing images. They can be categorized into two main groups: 
		\begin{itemize}
			\item Edge-based detection\cite{ref20, ref21, ref22}
			\item Region-based segmentation\cite{ref23, ref24}
		\end{itemize}

		In this way, many efforts have been made to improve the traditional approaches for detecting airports.
		
		The authors\cite{ref25} applied a novel coarse-to-fine model to detect airports from remote sensing images. This model consisted of two layers in its hierarchical architecture, the coarse layer for rapid airport candidate localization and the fine layer for detecting the target. A Conditional Random Field(CRF) was adopted in this layer to learn sparse features. It provided more accurate and robust detection to target scale variations.
		
		The authors\cite{ref26} utilized a two-layer saliency model to detect airports and recognize aircraft. First Layer Saliency(FLS), with the help of spatial-frequency visual saliency analysis, was designed to detect airport regions. The Second Layer Saliency(SLS) model using Edge Feature Preserving Wavelet Transform(EFP-WT) and high-frequency wavelet coefficient were proposed to extract aircraft candidates from candidate regions of the previous layer. In the next step, SVM was used as a classifier. This method indicated its reliability in detecting a target in high-resolution broad-area remote-sensing images and its robustness in complex scenes.
		
		A saliency model was also adopted with some modifications for this purpose in \cite{ref27}. A novel saliency model was proposed based on hierarchical reinforcement learning to reinforce the difference between background and airport. This method had the adaptive ability to be applied to high-resolution and low-resolution remote sensing images with a higher detection rate.
		
		A new method to detect airports from Synthetic Aperture Radar(SAR) images was proposed with the help of saliency analysis in \cite{ref28}. The proposed methods consisted of line segment grouping to extract airport support regions. The selective non-maximum suppression and controlled false-alarm were also employed to extract potential airport regions. The experimental results showed that this method improved the performance and efficiency of airport detection algorithms.
		
		In \cite{ref29}, a novel method based on Salient Line Segment Detector (SLSD) and Edge-Oriented Region Growing (EORG) was proposed for detecting airports in SAR images which increased the precision of detection results. The adopted SLSD in this method was able to detect line segments and highlight the edges of airport runways. The proposed plan had more accurate locations and better contours for airports.
		
		Authors in \cite{ref43} presented a method by combining spectral features and geometric features of airports to cope with complex background challenges. In this way, a decision tree algorithm was developed based on these features to extract the main concrete areas within the whole RSI. In general, the proposed method provides a high-accuracy detection rate compared to previous schemes.

		All proposed methods in these two categories have not achieved the desired efficiency and accuracy in detecting airports. Hence, recently, drawing from unique deep learning capabilities in image processing, particularly transfer learning, new methods have been provided in object detection. Deep learning-based object detection methods revealed their superiority in detecting objects compared to conventional ones.
		
		The introduction of CNNs for airport detection was firstly made by \cite{ref30}. This work employed the transfer learning ability of deep neural networks; For this aim, the trained CNN on natural images is applied for detecting objects in satellite images. Besides, the prior knowledge about airports, such as having a long parallel runway, was used in the region proposal stage. This method improved airport detection approaches with a higher detection rate in seconds’ computational time. The detection rate reached 88.8\% in experimental results.
		
		To enhance the efficiency of the training phase, a hard examples mining layer and an improved weight-balanced loss function in a unified CNN structure were proposed in \cite{ref31}. Reducing false positives rate was addressed by offering a cascade design of RPN and object detection in this network. The method had a good performance in detecting airports from the multiscale dataset with complicated background.
		
		To overcome illumination intensities and contextual information in remote sensing images, researchers in \cite{ref32} applied a multiscale fusion feature using a new GoogleNet light feature model. In this approach, an improved SVM with a hard negative mining method was used for classification; Finally, a simplified bounding-box regression optimizes the locations. It was shown that employing these three strategies improves the precision and optimize the region locations.
		
		The automatic fast airport detection is proposed in \cite{ref33}. Here, two CNNs in the Faster R-CNN framework were used for detecting airports. The first one was for obtaining candidate regions and the second one was for the detection based on those extracted features. This approach has better performance with less pre-processing compared to other airport detection algorithms.
		
		To improve the pace and precision of airport detection, transfer learning on the basis of end-to-end Faster R-CNN was employed in \cite{ref8} and \cite{ref19}. The proposed method architecture in \cite{ref8} was equipped with cascade RPN’s structure and multiple threshold detection strategies to improve the quality of candidate boxes and loss function, respectively. However, the method in \cite{ref19} benefited from a cascade region proposal network, multilayer feature fusion, and hard example mining which improved airport detection performance.
		
		Another airport detection method based on faster R-CNN was proposed in \cite{ref34}. This algorithm had multiscale training and a modified multi-task loss function to improve robustness and accuracy. Furthermore, the online hard sample mining strategy was employed to satisfy a balance between positive and negative samples. The proposed method in this paper showed an outstanding performance in detecting airports under complex backgrounds with higher detection rates and lower false alarm rates in a short running time.
		
		In \cite{ref35}, cycling, offline learning, and online representation (COLOR) framework was proposed to address needing a large number of training examples and their annotating. This framework includes three modules: cycling by example refinement (C) for refining coarse examples, offline learning (OL) for tuning CNN, and online representation (OR). The OR module was based on a coarse-to-fine cascaded CNN for airport classification and airplane detection. This method reached good accuracy as other approaches in this regard while it is much faster than others due to proposing a coarse-to-fine cascaded OR module.
		
		Authors in \cite{ref42} establish a framework for detecting unknown airport distributions in a wide area based on deep network and geographic analysis. In this way, correct points are extracted from an existing airport dataset (Google image). A classification model was trained to classify and extract candidate airport regions. Finally, the airport confidence was computed to extract the exact location of airports in the selection area. It should be noted that geographical data such as road networks and water systems were employed to analyze the detection results. The framework increased the discovery rate of airports with various confidence in the target area.
		
\begin{figure}[t!]
	\center
	\includegraphics[width=0.39\textwidth]{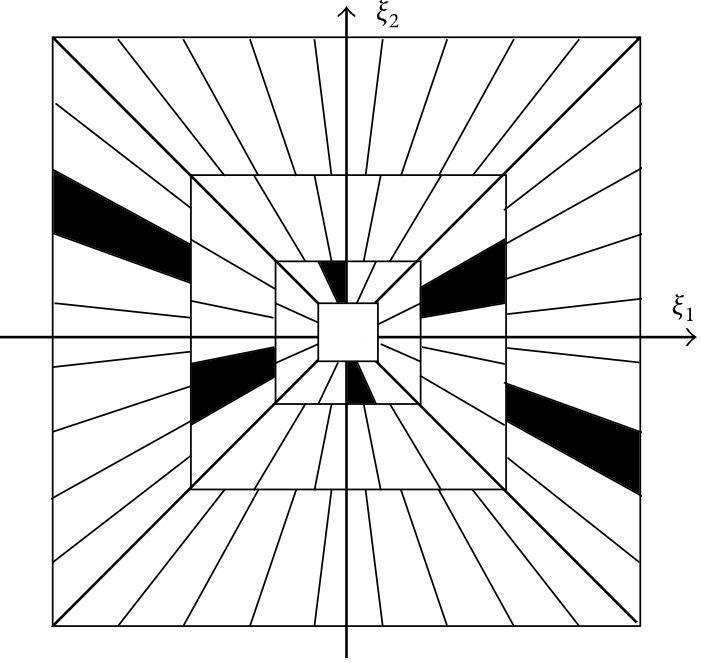} 
	\caption{The tiling of frequency planes based on Shearlet transform\cite{ref41}.}
	\label{fig:tiling}
\end{figure}
\begin{figure}[t!]
	\center
	\setlength{\tabcolsep}{2pt}
	\begin{tabular}{cc}
		\includegraphics[width=0.23\textwidth]{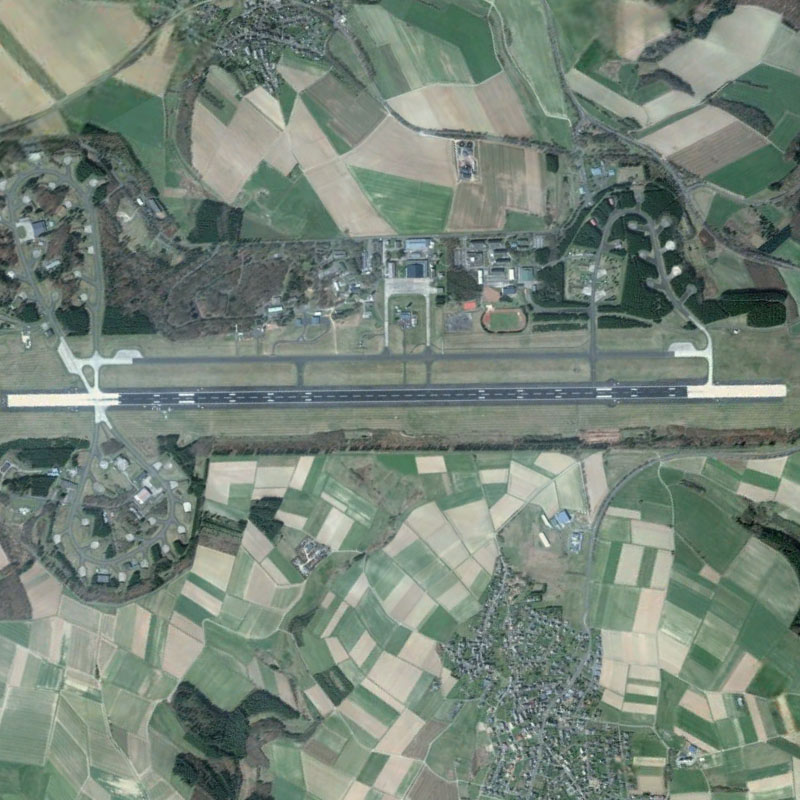} &
		\includegraphics[width=0.23\textwidth]{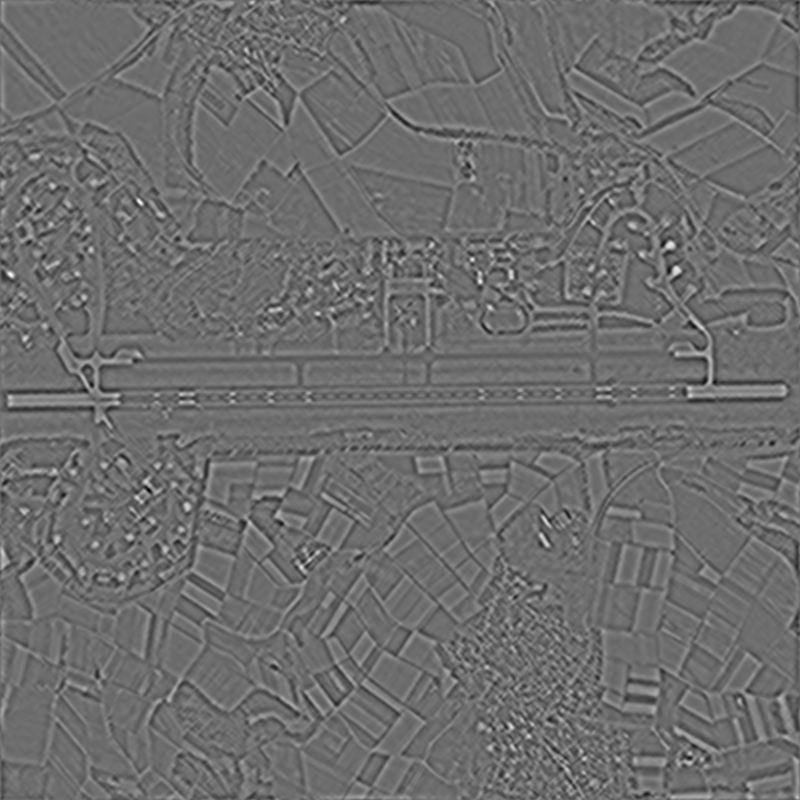} \\
		(a)&(b)\\
		\includegraphics[width=0.23\textwidth]{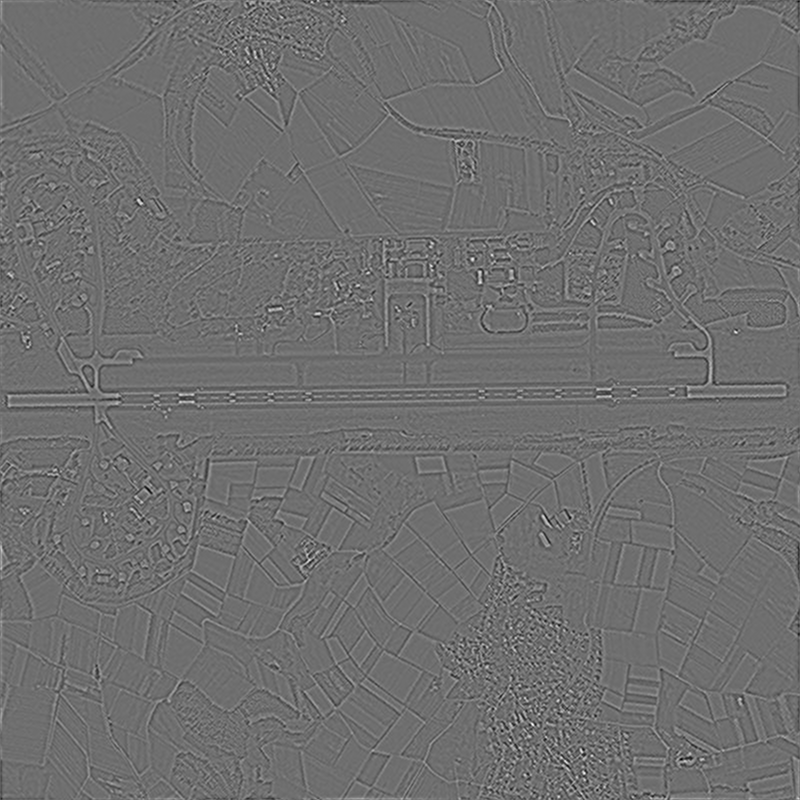} &
		\includegraphics[width=0.23\textwidth]{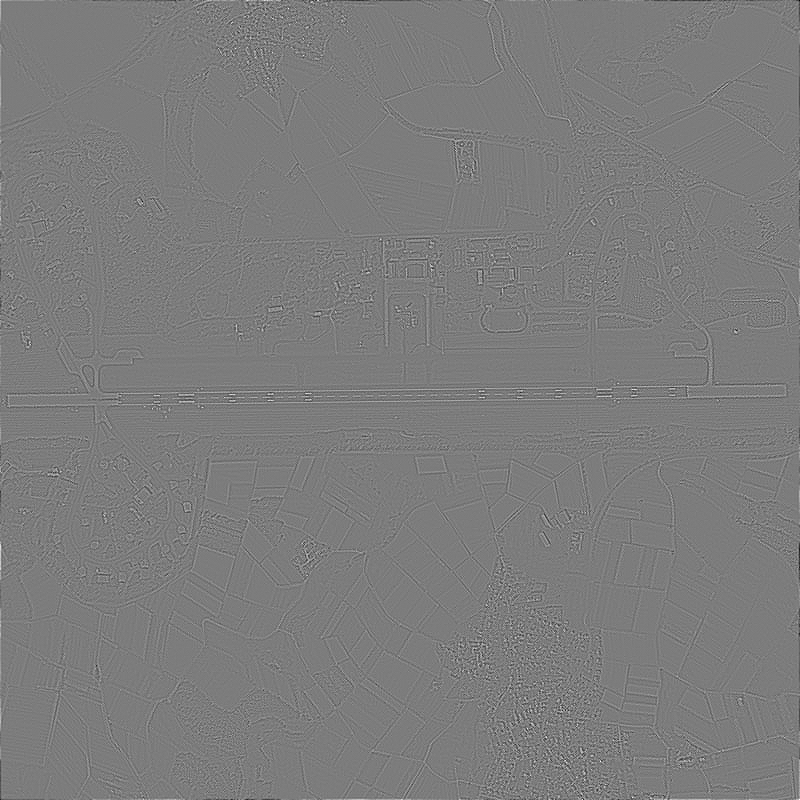} \\
		(c)&(d)\\
	\end{tabular}
	\caption{The high-frequency decomposition based on ShearLet.(a) Original image, (b-d) Combined details based on ShearLet (lower to higher scales, respectively).}		
	\label{fig:shearoutput}
\end{figure}
		\subsection{Key Contributions of EYNet}
		As mentioned, airport detection research has extended dramatically in the last few years. However, it still requires advancement, particularly in detecting tiny airports in the high-resolution image and tackling the complex backgrounds images. Hence, this study establishes a novel scheme by extending YOLOV3 named EYNet for airport detection in remote sensing images. 
		
		To do so, two small and low-latency networks, including ResNet18(a residual network) and MobileNet18, are chosen as the based network of YOLO. Instead of the ImageNet dataset, the mentioned networks are retrained on a similar dataset as classification problems to boost the model's performance and decrease the training time of the goal problem. So, the candidate networks' whole parameters are finetuned in the relevant dataset. Moreover, the high-frequency filters of Discrete ShearLet in different directions and scales are utilized in the first convolution layers of ResNet18 to decompose the low and high frequencies of the input image. 
		
		Due to airports' unique characteristics that contain parallel or diagonal lines, this visual attention mechanism can suppress irrelevant lines and effectively highlight the edges of airport runways. Besides, novel modifications and improvements are employed in the topology of YOLOV3 to improve the detection accuracy of airports. For this aim, the structure of detection subnetworks and the connections between the detection network sources are upgraded. Something else that should be mentioned is a novel augmentation and hard example mining strategies that effectively increase the proposed scheme's performance. Briefly, the main contributions of EYNet are itemized in the following aspects:
		\begin{itemize}
			\item Applying two parallel pre-trained networks
			\item Extending YOLOV3 structure and other customization
			\item Novel augmentation and negative mining strategies
			\item ShearLet Transform as attention mechanism
		\end{itemize}
		The experimental results demonstrate that EYNet can enhance the effectiveness and the detection robustness compared to other state-of-the-art schemes.	
		\subsection{Road map}
		The remaining parts of this paper are organized as follows. In Section \ref{sec:Preliminaries}, the essential components of YOLOV3 as real-time object detection and ShearLet transform are briefly explained. Next, Section \ref{sec:Proposed} provides an overview and makes detailed descriptions of several steps of the presented framework. Section \ref{sec:Experimental} analyzes and evaluates the experimental results of the proposed method on real Airport images in detail. Also, a comparison with traditional YOLOV3 and state-of-the-art schemes will be covered in this section. Finally, the conclusions and the future works are presented in Section \ref{sec:Conclusion}.

	\begin{figure*}[t!]
	\center
	\setlength{\tabcolsep}{2pt}
	\begin{tabular}{cccccc}
		\includegraphics[width=0.13\textwidth]{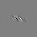} &
		\includegraphics[width=0.13\textwidth]{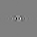} &
		\includegraphics[width=0.13\textwidth]{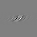} &
		\includegraphics[width=0.13\textwidth]{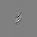} &
		\includegraphics[width=0.13\textwidth]{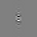} &
		\includegraphics[width=0.13\textwidth]{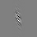} \\
		
		\includegraphics[width=0.13\textwidth]{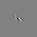} &
		\includegraphics[width=0.13\textwidth]{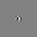} &
		\includegraphics[width=0.13\textwidth]{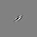} &
		\includegraphics[width=0.13\textwidth]{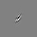} &
		\includegraphics[width=0.13\textwidth]{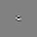} &
		\includegraphics[width=0.13\textwidth]{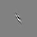} \\
		
		\includegraphics[width=0.13\textwidth]{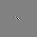} &
		\includegraphics[width=0.13\textwidth]{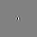} &
		\includegraphics[width=0.13\textwidth]{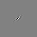} &
		\includegraphics[width=0.13\textwidth]{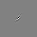} &
		\includegraphics[width=0.13\textwidth]{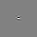} &
		\includegraphics[width=0.13\textwidth]{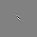} \\
		
	\end{tabular}
	\caption{Visualizing weights of ShearLet filters. Three different scales of ShearLet filters with six directions for each scale.}
	\label{fig:ShearLet}

	\end{figure*}

	\begin{figure*}[t!]
		\center
		\setlength{\tabcolsep}{2pt}
		\begin{tabular}{cccccc}
			\includegraphics[width=0.15\textwidth]{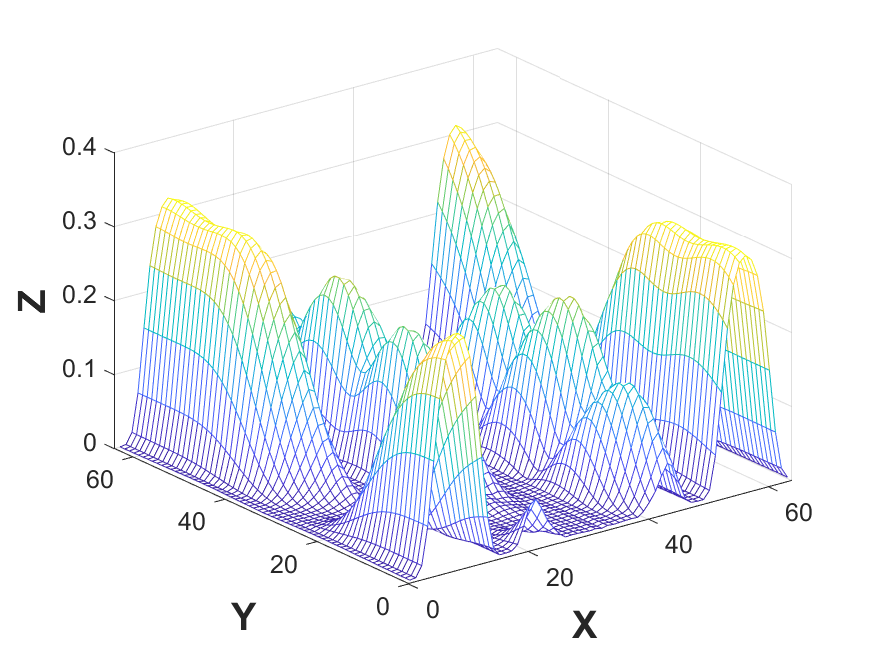} &
			\includegraphics[width=0.15\textwidth]{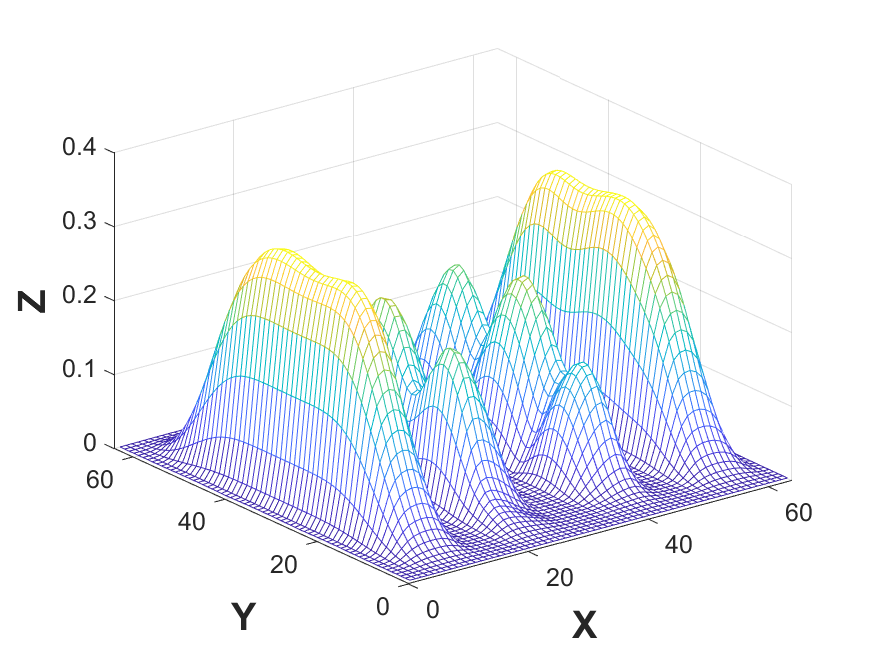} &
			\includegraphics[width=0.15\textwidth]{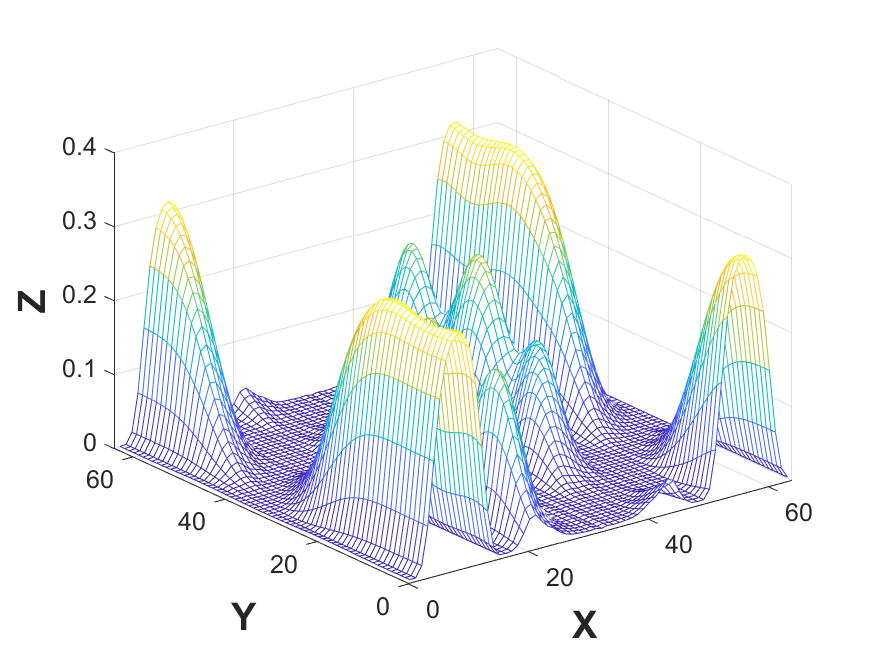} &
			\includegraphics[width=0.15\textwidth]{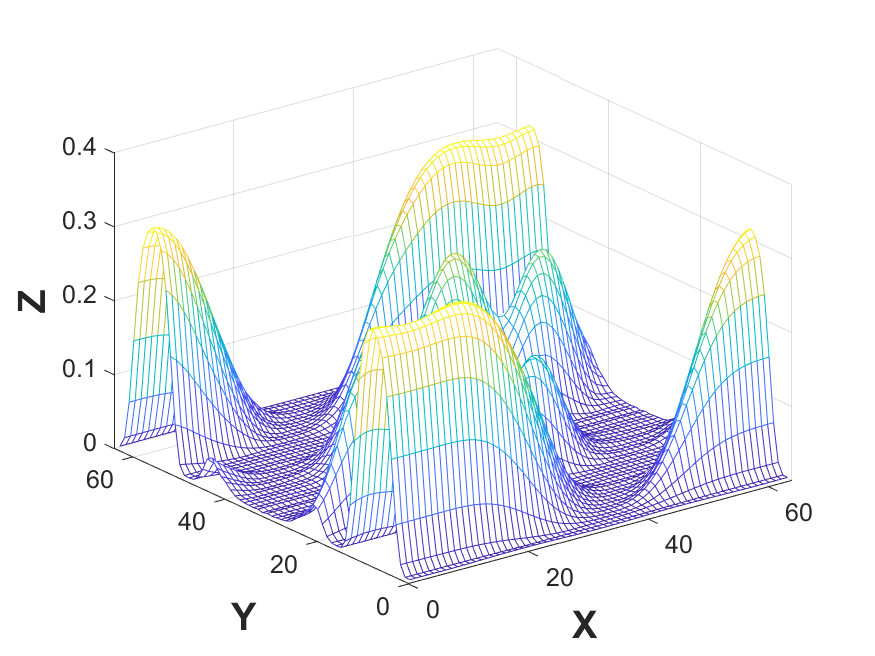} &
			\includegraphics[width=0.15\textwidth]{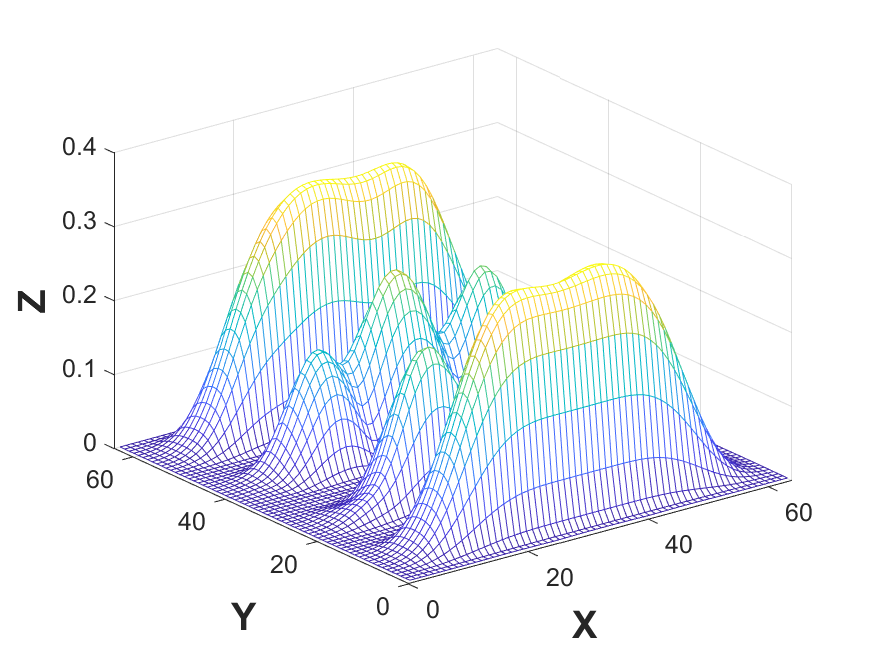} &
			\includegraphics[width=0.15\textwidth]{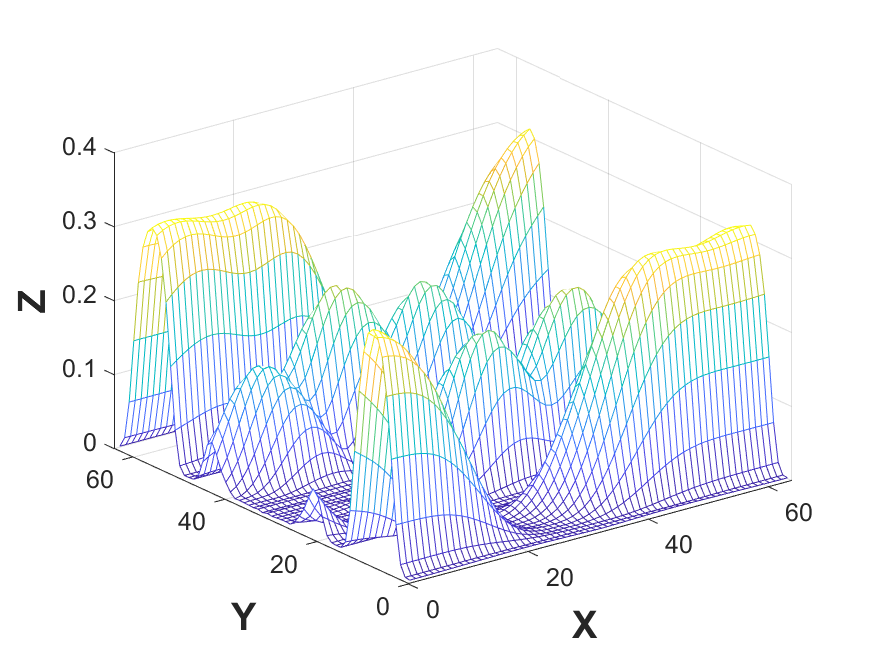} \\
			
			\includegraphics[width=0.15\textwidth]{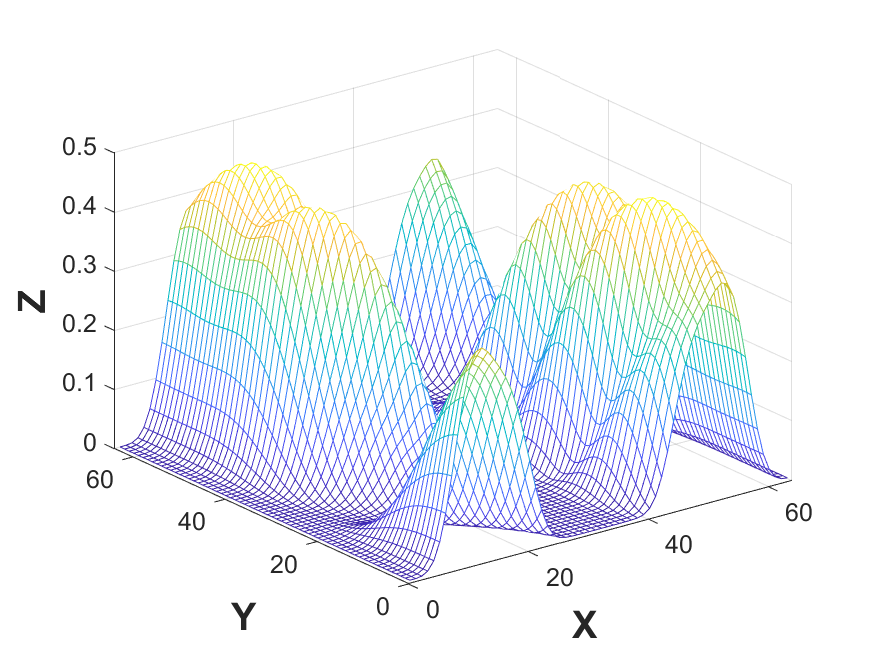} &
			\includegraphics[width=0.15\textwidth]{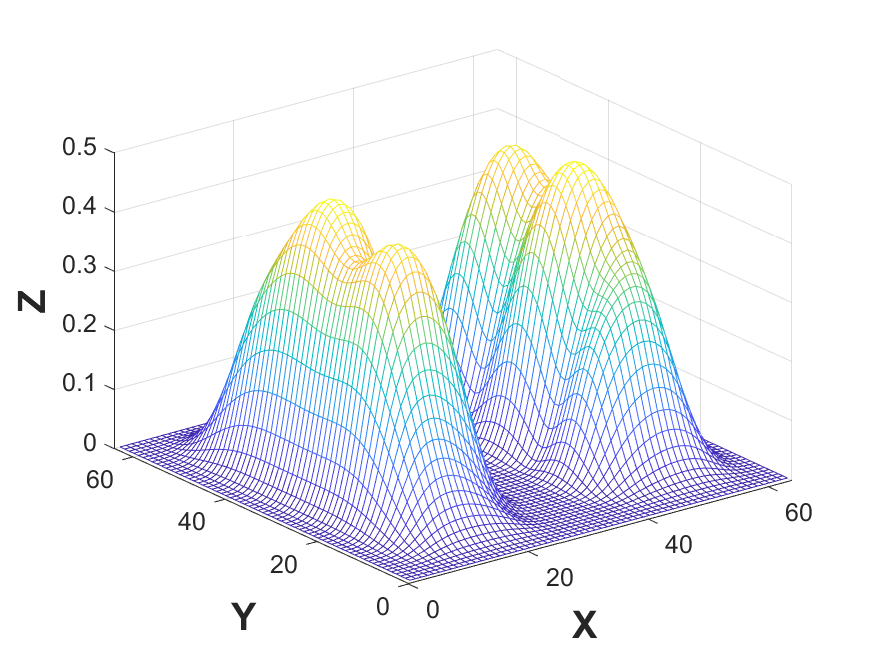} &
			\includegraphics[width=0.15\textwidth]{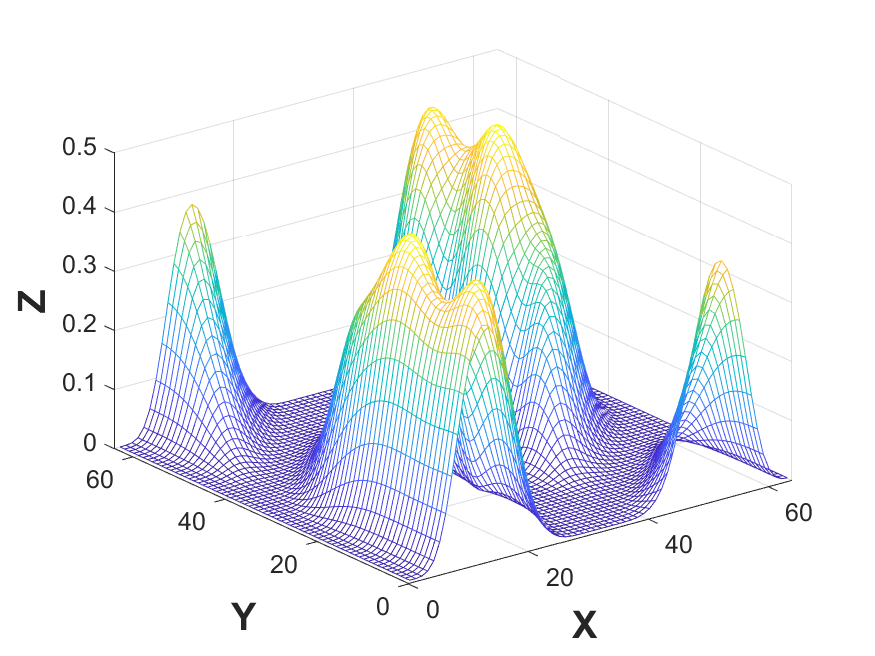} &
			\includegraphics[width=0.15\textwidth]{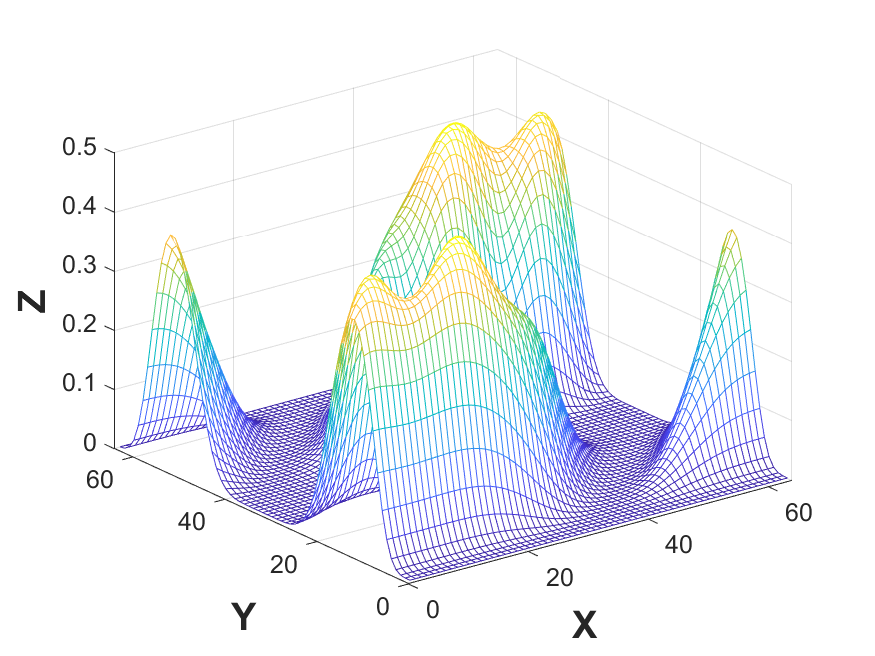} &
			\includegraphics[width=0.15\textwidth]{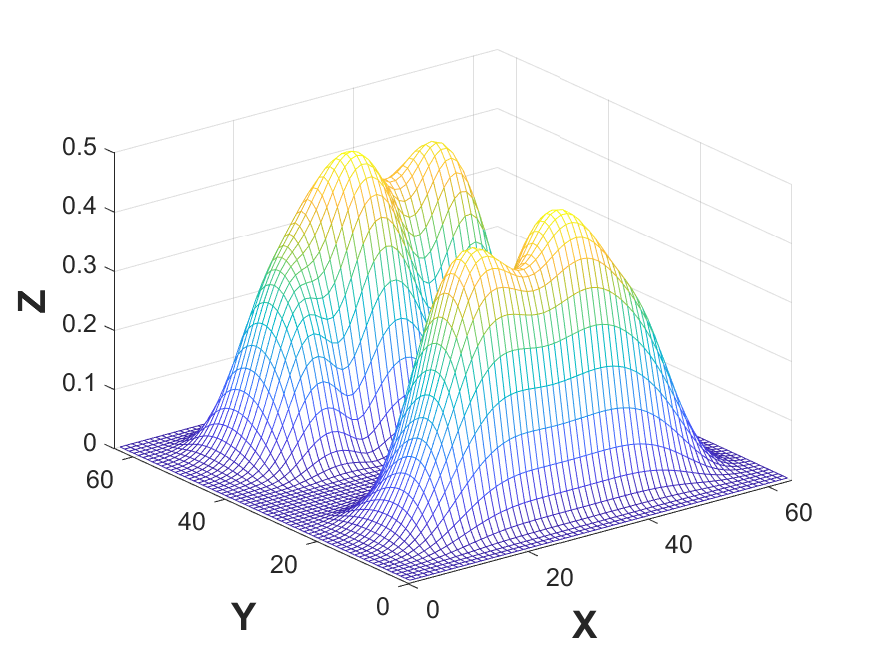} &
			\includegraphics[width=0.15\textwidth]{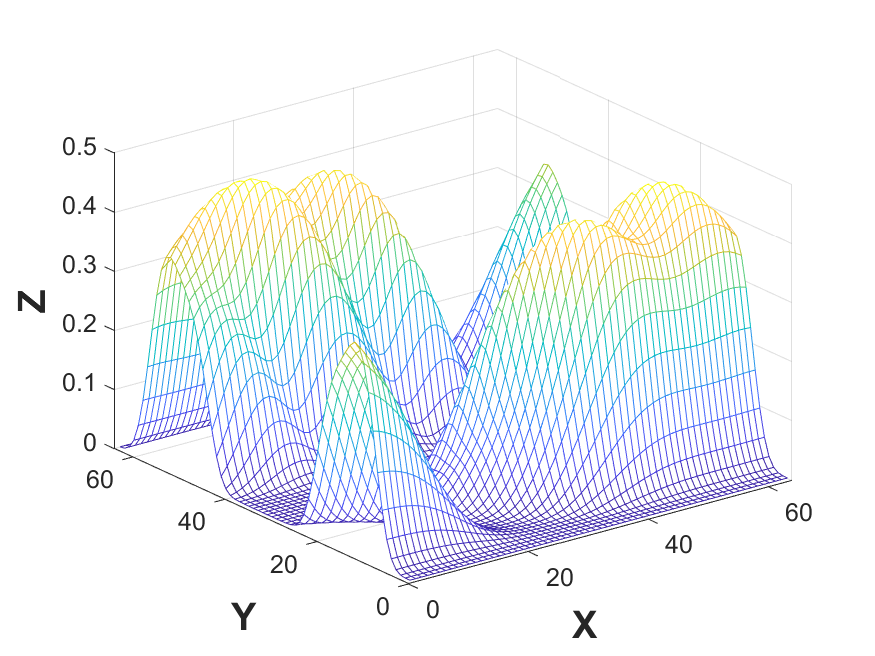} \\
			
			\includegraphics[width=0.15\textwidth]{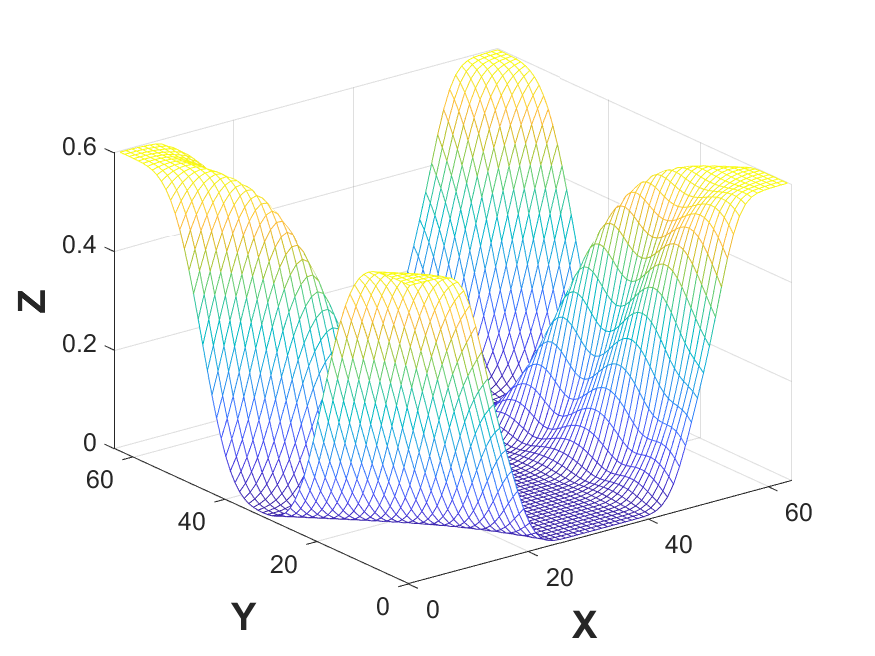} &
			\includegraphics[width=0.15\textwidth]{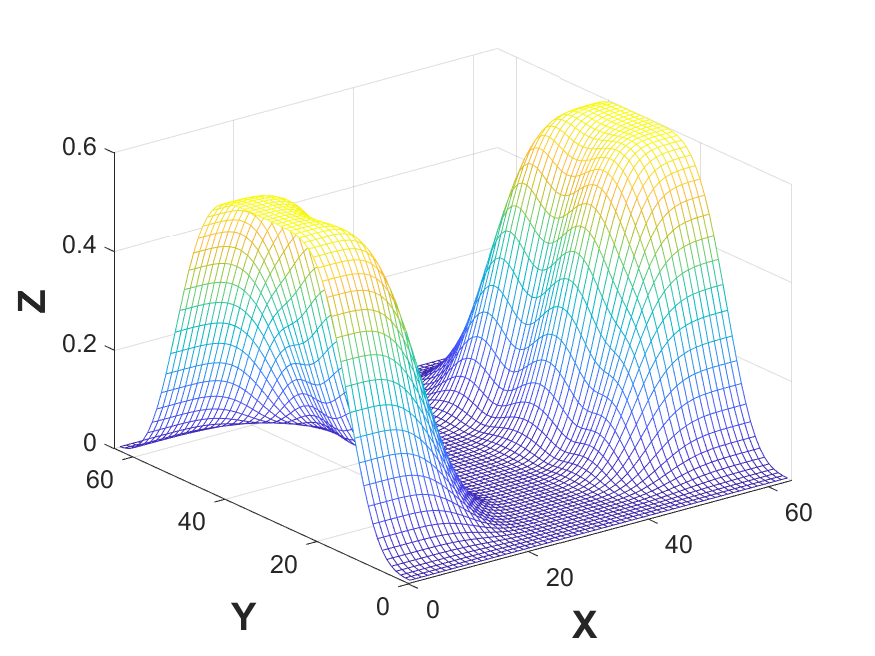} &
			\includegraphics[width=0.15\textwidth]{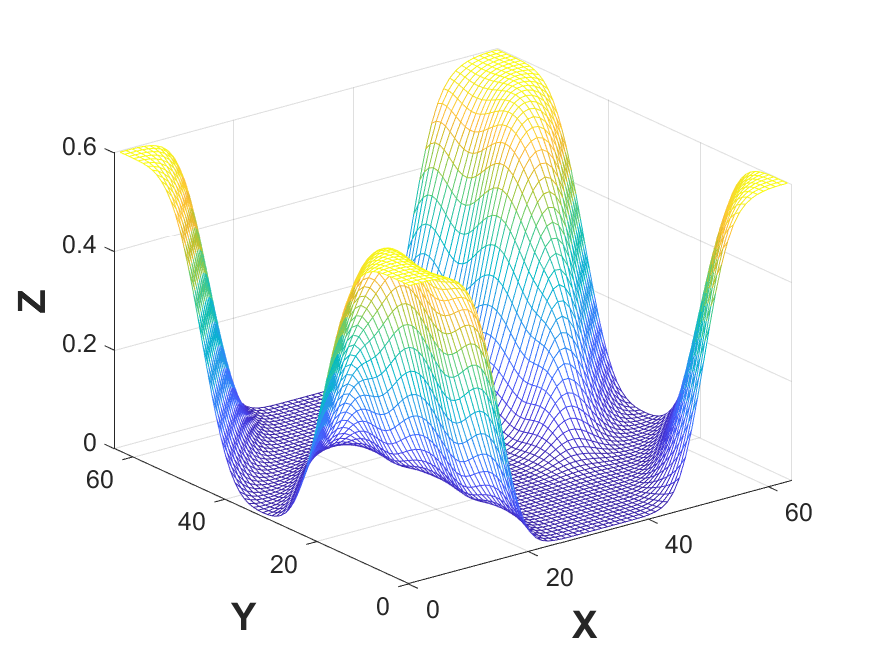} &
			\includegraphics[width=0.15\textwidth]{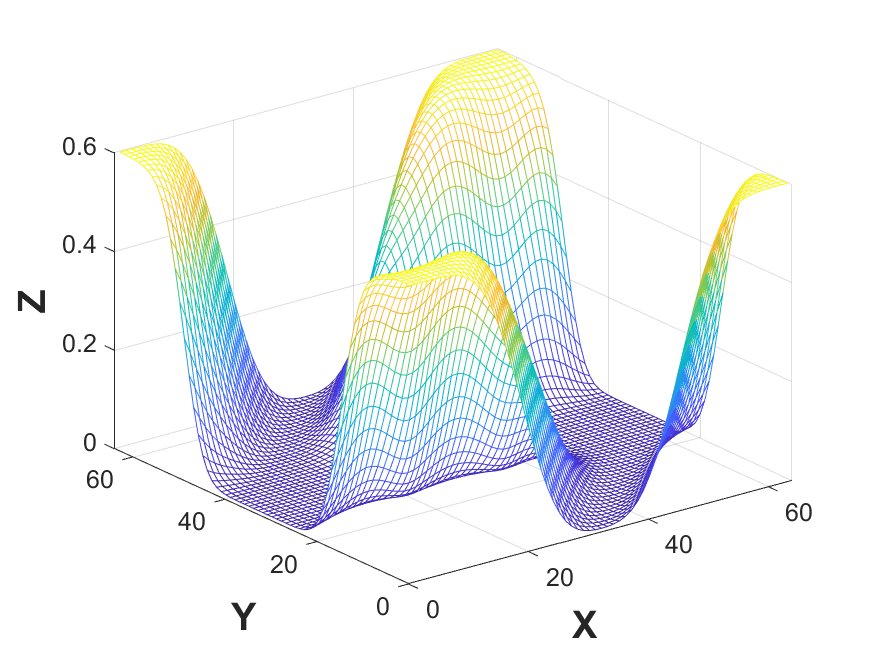} &
			\includegraphics[width=0.15\textwidth]{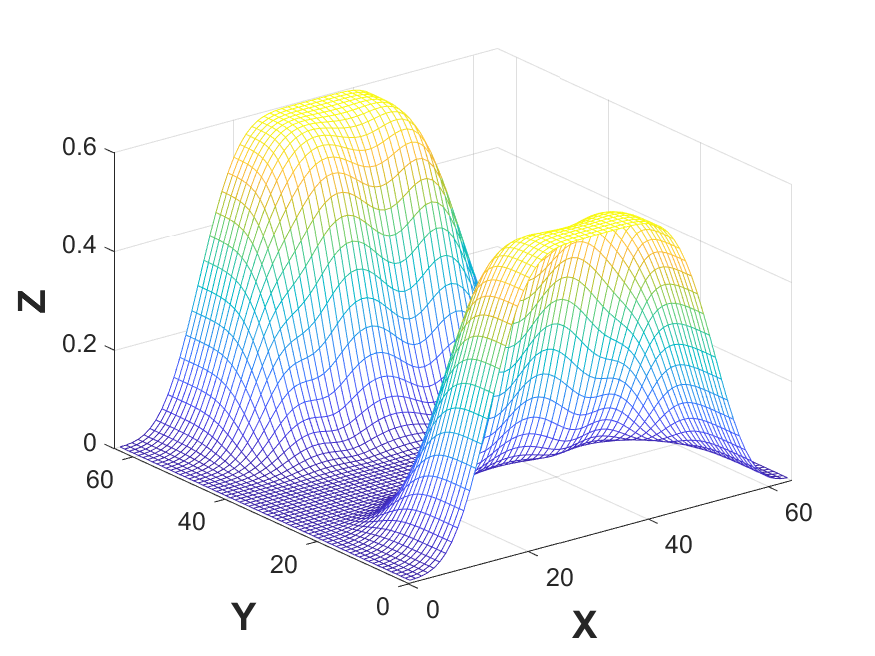} &
			\includegraphics[width=0.15\textwidth]{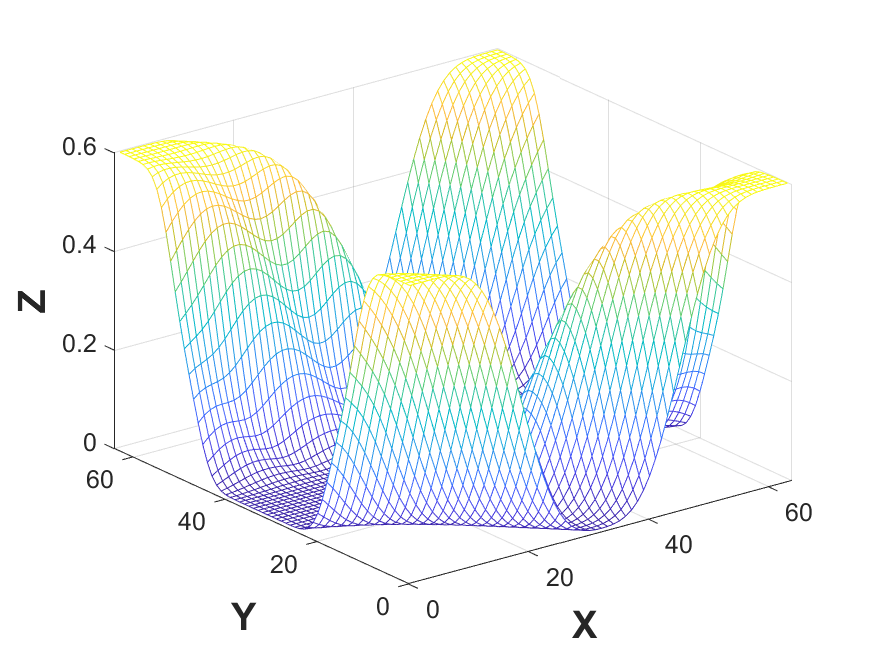} \\
			
		\end{tabular}
		\caption{ 
			The frequency response (band's weights) of three different scales of ShearLet filters. Six direction for each scales. Axis x, y, and z shows Fx, Fy, and Magnitude, respectively.}
		\label{fig:frequency}
	\end{figure*}

	\begin{figure*}[t!]
		\includegraphics[width=1\textwidth,trim=2cm 10cm 8cm 4cm,clip]{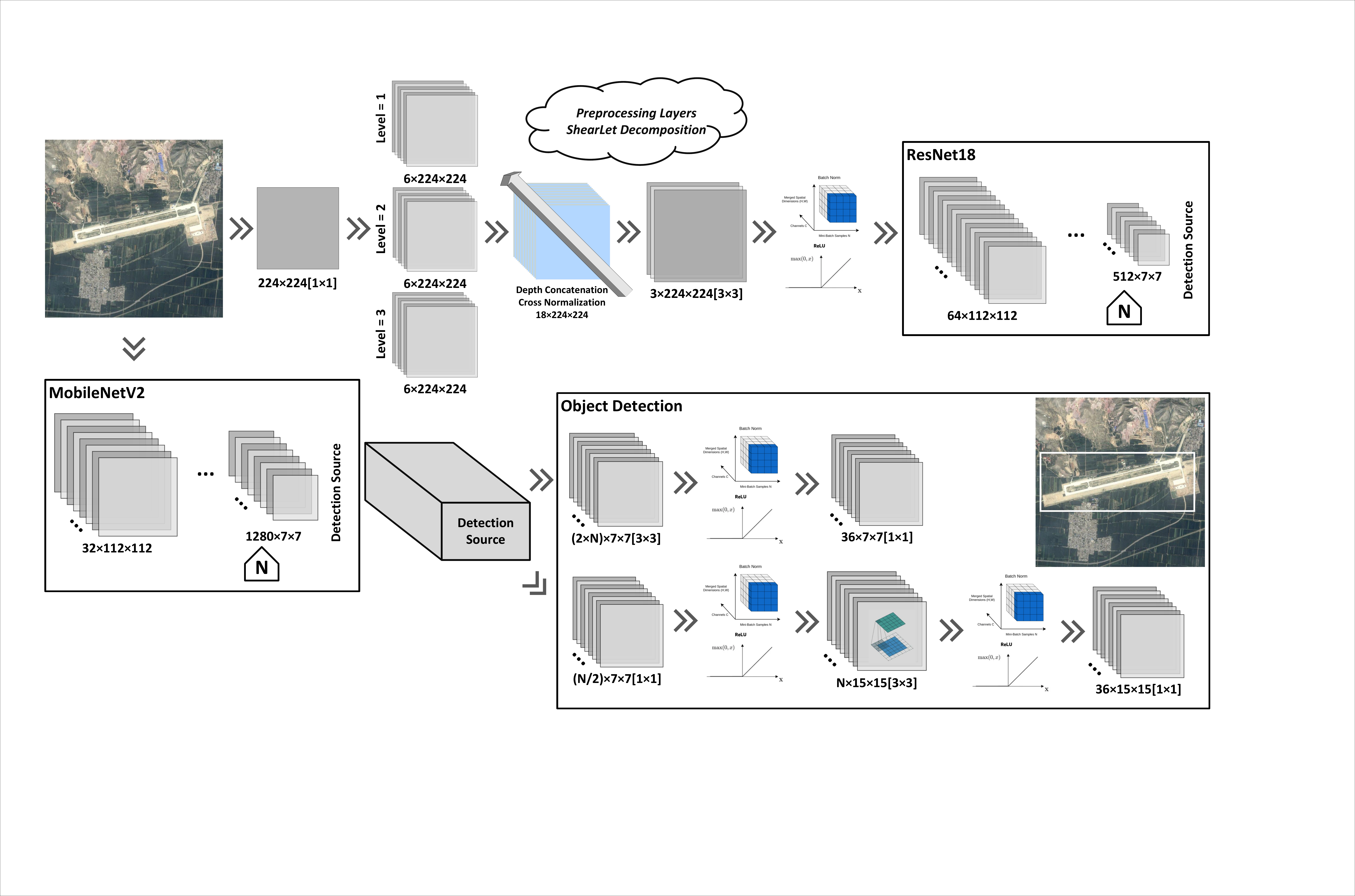} 
		\caption{The structure of the proposed network(EYNet). N$\times$X$\times$Y represents the number of filters and width and height of output, respectively. Also, [m$\times$n] is the size of the filter in the corresponding layer.}
		\label{fig:digram}
	\end{figure*}

	\section{Preliminaries}
	\label{sec:Preliminaries}
	This section describes the principal background material of EYNet. In this way, first, the ShearLet transform is briefly introduced. Then, a short review of YOLO (particularly YOLOV3) is explained. For more information about these concepts, refer to \cite{ref38, ref39, ref41} and \cite{ref16, ref44, ref45}, respectively.	
	\subsection{ShearLet Transform}
	
	A short introduction is given about the ShearLet transform in this sub-section. ShearLet transform is an extended version of the wavelet family with a simple mathematical model. Although wavelet is appropriate for approximation of one-dimensional signals, it is not efficient for two or multi-dimensional data the representation of curves. A shearLet framework is multi-directional and multi-scale, which is presented to overcome the last transform's crucial weaknesses, such as edge, contour, curve, and boundary analysis, particularly on a small scale. In other words, it is high directionality and representation of mentioned salient features of the signal in a better way. The geometry of objects can be significantly preserved with the help of it. In detail, this transform is an affine scheme including a mother ShearLet function with three parameters: scale, shear, and translation. Moreover, it is able to produce a collection of basic functions using the scale, transform, and rotation functions. Besides, it can be divided step by step in frequency space, which increases their efficiency. These advantages are the superiority of the Shearlet transform over the traditional wavelet. Figure. \ref{fig:tiling} illustrated the tiling of the frequency planes based on the ShearLet transform.
	
	Recently, ShearLet has been widely used in considerable image processing problems such as edge analysis, inpainting, image separation, inverse scattering, image interpolation, image denoising, medical image, etc. In this work, the ShearLet filters are adapted as the weight of the first convolution layers of ResNet18. To the best of our knowledge, it is the first time which employed these filters in such way. Figure. \ref{fig:shearoutput} demonstrates the result of shearlet decomposition of the candidate instance. As can be expected, it can effectively represent the details of the image in three different scales. The candidate filters of three scales and the corresponding frequency response(band's weights) are visually illustrated in the Figures. \ref{fig:ShearLet} and \ref{fig:frequency}, respectively.
	\subsection{You Only Look Once (YOLOV3) }
	Most traditional object detection schemes like R-CNNs use regions proposal networks to localize the objects. In other words, the network had to look at the image parts that have a high chance of containing objects. These models include two output layers: class probability distribution and bounding box prediction. 

	You Only Look Once(YOLO) is an end-to-end real-time object detection scheme that is extremely different compared to the mentioned approaches. In general, a single-stage architecture simultaneously extracts features and predicts the object's bounding box(s) by taking the entire input. Obviously, the primary purpose of YOLO is to separate an instance into grid cells and predict the objectness probability using anchor boxes. In this way, each cell must predict a bounding box and confidence scores for those boxes. Also, the anchor boxes, a set of predefined bounding boxes, improve the speed and efficiency of the object detection portion. On the other hand, YOLO combines feature extraction and object localization steps into a single network. Moreover, the localization and classification heads were also united.
	
	In summary, the significant advantages of YOLO compared to previous schemes are its super speed, scanning the entire image during training and testing times, and learning generalizable representations. In contrast, the poor detection of small objects is the crucial weakness of YOLO.
	
	The YOLOV3 architecture (Darknet-53) is a nominee for airport detection in this research. This model adopts the detection phase into three different scales. The detection is done by only utilizing 1$\times$1 kernel on the feature maps. It also uses binary cross-entropy loss instead of the Mean Squared Error(MSE). The probability of objects in the image and the class predictions are done using logistic regression. Moreover, three prediction heads are employed to process the image at different spatial scales. The following section will describe the significant modifications applied to YOLOV3.
	
	\section{Proposed Method}
	\label{sec:Proposed}
	This work proposed an effective airport detection scheme (EYNet) based on extending YOLOV3 and ShearLet Transform to overcome the mentioned challenge in the previous section. Three main phases organize EYNet: Transfer Learning, Preprocessing, and Object Detection Phases. First, two pre-trained networks MobileNetV2 and ResNet18 are retrained on a similar dataset compared to the main problem. Afterward, the extended YOLOV3, using ShearLet filters and modification in layers structure and training strategy, is employed in the object detection phase. These phases are detailed in the following subsections. The overall architecture of the proposed network (EYNet) is illustrated in Figure. \ref{fig:digram}.
	
	\subsection{Transfer Learning}
	Recently, transfer learning science has been widely employed in computer vision tasks. For this aim, a pre-trained network model (e.g., ImageNet pre-trained model) is chosen to be the initial weight of the new problem. In other words, with the help of this science, a source problem directly affects the inductive bias of the target one. In this way, a set of training examples in the target domain are considered for fine-tuning the model. Generally, this is a typical way and one of the most popular strategies among neural network applications, which conducts to achieve high accuracy and generalization in optimum time.
	
	In this work, MobileNetV2 and ResNet18 models, trained on ImageNet for the classification task, are utilized in the transfer learning phase. A few parameters and depth of networks compared to the rest are some significant characteristics of the mentioned models. On the other hand, the type of objects used in the previous version of the model significantly affects the performance of the new models. Hence, in this work, the mentioned models are retrained on some relevant classes of the NWPU-RESISC45 dataset \cite{ref36} which are similar to the main target task. Something which should be mentioned here is that some modifications are employed as preprocessing on input images in ResNet18. These customizations are completely described in the next sub-section.  
	
	\subsection{Preprocessing}
	This subsection explains the main preprocessing steps employed in this work. This phase of EYNet is overall classified into three units: ShearLet Filters, Detection Source, and Estimate Anchors. These modifications effectively promote the object representation capacity and increase the accuracy rate.
	\begin{figure}[t!]
		\center
		\setlength{\tabcolsep}{2pt}
		\begin{tabular}{ccccc}
			\includegraphics[width=0.45\textwidth]{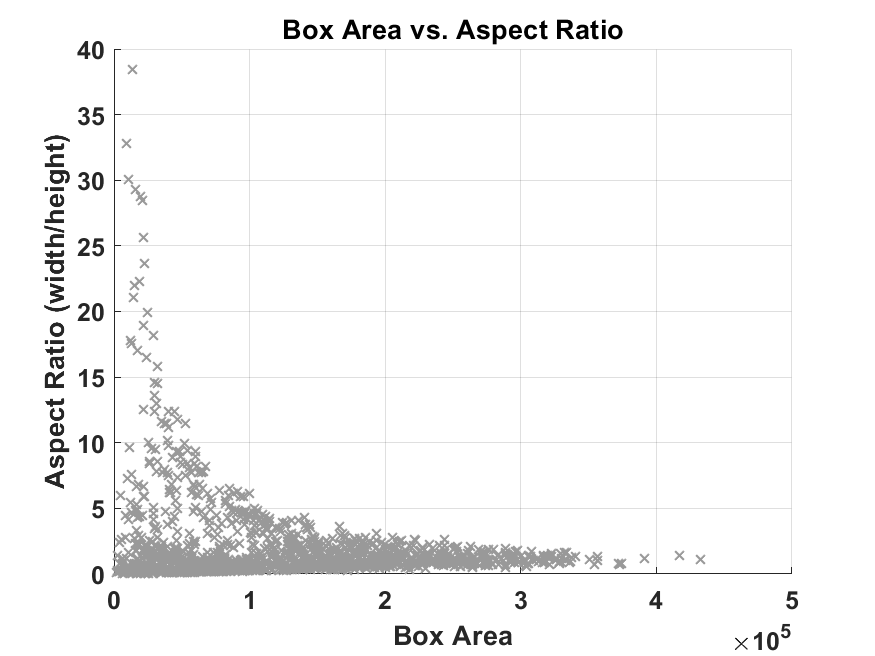} \\
			(a) \\
			\includegraphics[width=0.45\textwidth]{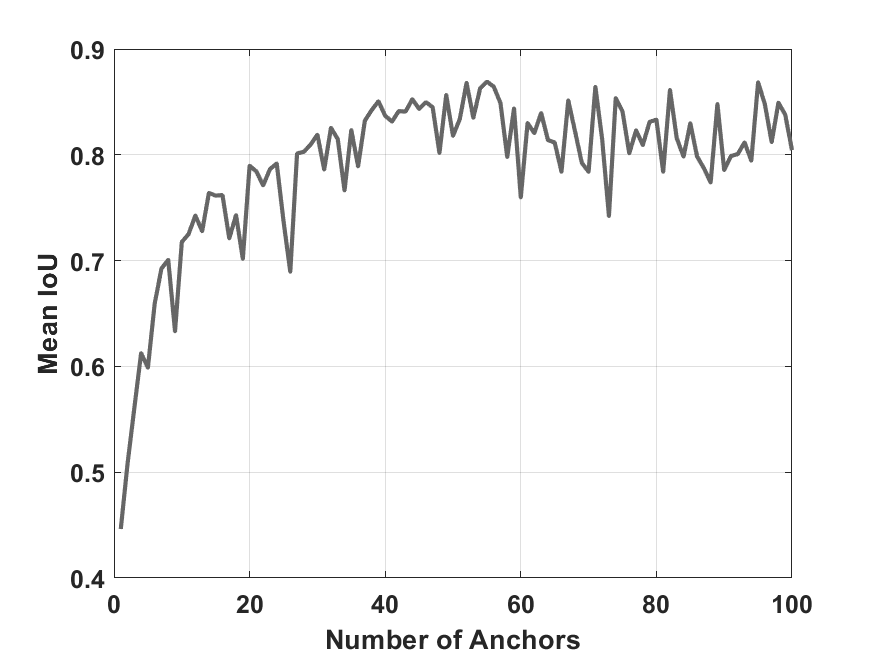} \\
			(b) \\
		\end{tabular}
		\caption{Estimate Anchors. (a) Box Area vs. Aspect Ratio, (b) Number of Anchors vs. Mean IoU for Airport objects.}
		\label{fig:anchor}
	\end{figure}
	
	\begin{figure}[t!]
		\center
		\includegraphics[width=0.45\textwidth]{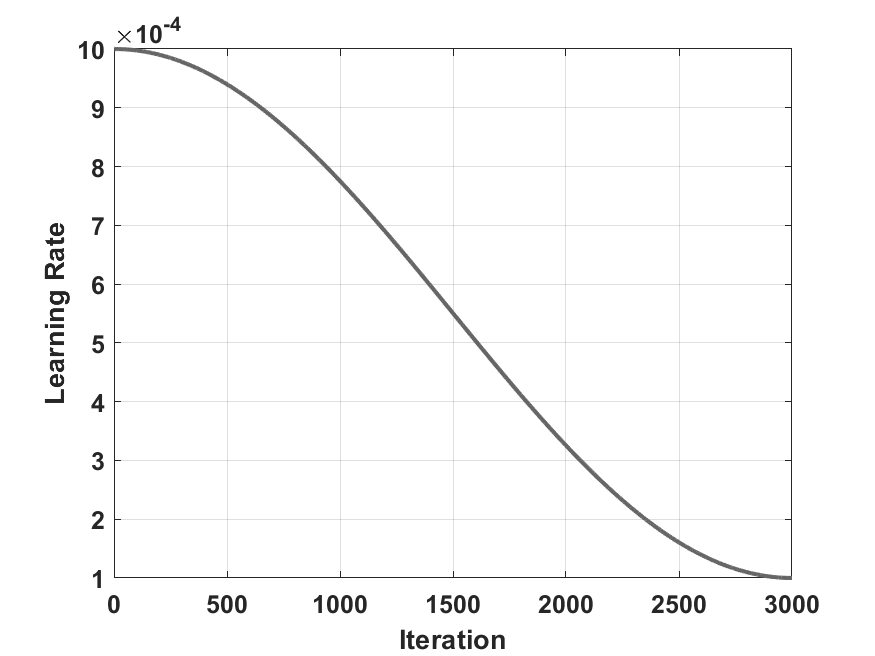} 
		\caption{The form of learning rate schedule with the help of Warm Restart (Cyclical) mechanism.}
		\label{fig:warm}
	\end{figure}
	
	\subsubsection{ShearLet Filters}				
	In this work, the ShearLet filters \cite{ref38, ref39, ref41}, in different directions and scales, are adopted to preserve the object's geometry and improve airport detection accuracy. For this aim, ShearLet filters with three scales and six directions are considered for forming the initial weights of three convolution layers. In detail, three parallel convolution layers with the size of $5\times5$ and weight/bias learning rate equal to zero are generated. The initial weights are computed by Equation. \ref{eq:shear}:
	
	\begin{equation}
		x(t) = f\bigg(X_s^d(\omega)\bigg)^{-1}, \forall s \in [1, 3], d \in [1, 6]
		\label{eq:shear}
	\end{equation}	
	where $f(n)^{-1}$, $s$, and $d$ are inverse Fourier transform, scale and direction, respectively. Also, $x(t)$ and $X(\omega)$ are coefficients in spatial and frequency domains, respectively.
	
	Something which should be noticed here is that these convolution layers are employed as the first layers of ResNet18 in both the transfer learning and object detection phases of EYNet. The significant characteristic of Airport objects with the help of these filters should be passed and preserved through whole layers of the network. Skip-layer connection (shortcuts) of ResNet18 effectively fuses multilayer features, promotes the object representation capacity, and prevents the degradation problem. Moreover, it can overcome the vanishing gradient problem. Figures. \ref{fig:ShearLet} and \ref{fig:frequency} demonstrate these filters and frequency response of them, respectively.						 
	
	\subsubsection{Detection Source}			
	The detection network source(s) play a significant role in YOLO object detection. Hence, a dynamic strategy should be employed to select the best layer(s). In this work, based on the experiments that resulted using Support Vector Machine(SVM) and mentioned models as feature extractor, the last convolution layers (after normalization) of each model has higher accuracy compared to the rest layers. In detail, the mentioned five classes are used to extract learned image features and use them for training classier.
	
	Although original YOLO can overcome the small objects by selecting the middle and last layers of the model, EYNet only used the last layers of models. In other words, predicting boxes at different scales not only prevents boosting the performance but also reduces the average precision rate. Hence, the last layers of mentioned networks are considered as detection layers to extract more robust and distinguishing features. In the following, EYNet utilizes these potential layers as inputs for fine airport detection and localization.

	\begin{figure*}[t!]
		\center
		\setlength{\tabcolsep}{2pt}
		\begin{tabular}{cccccc}
			\includegraphics[width=0.12\textwidth]{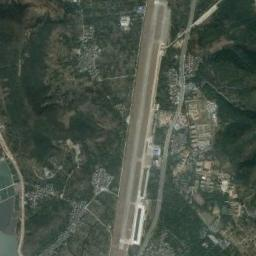} &
			\includegraphics[width=0.12\textwidth]{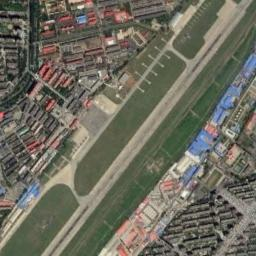} &
			\includegraphics[width=0.12\textwidth]{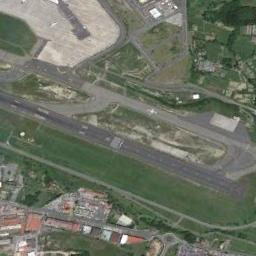} &
			\includegraphics[width=0.12\textwidth]{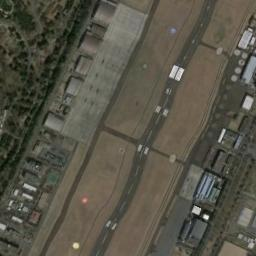} &
			\includegraphics[width=0.12\textwidth]{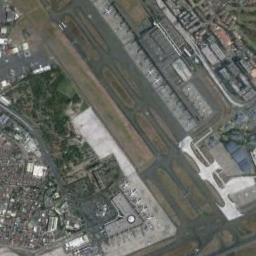} &
			\includegraphics[width=0.12\textwidth]{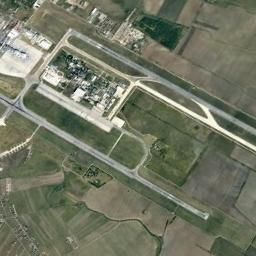} \\
			
			\includegraphics[width=0.12\textwidth]{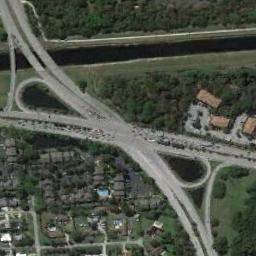} &
			\includegraphics[width=0.12\textwidth]{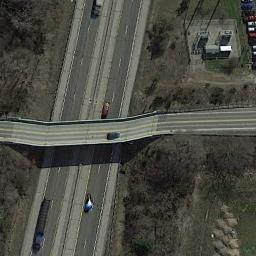} &
			\includegraphics[width=0.12\textwidth]{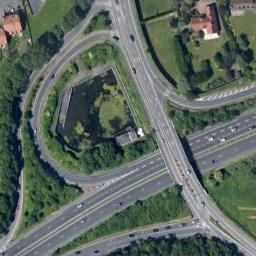} &
			\includegraphics[width=0.12\textwidth]{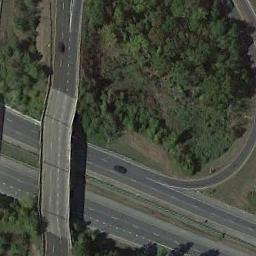} &
			\includegraphics[width=0.12\textwidth]{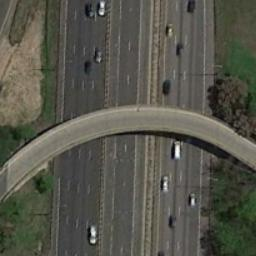} &
			\includegraphics[width=0.12\textwidth]{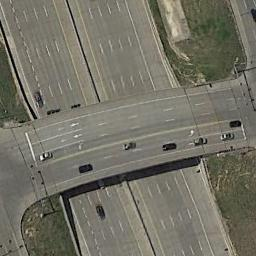} \\
			
			\includegraphics[width=0.12\textwidth]{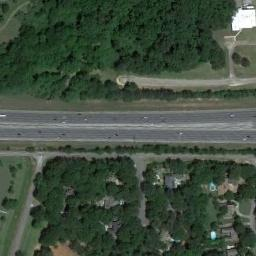} &
			\includegraphics[width=0.12\textwidth]{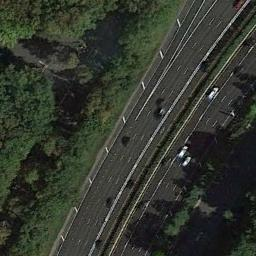} &
			\includegraphics[width=0.12\textwidth]{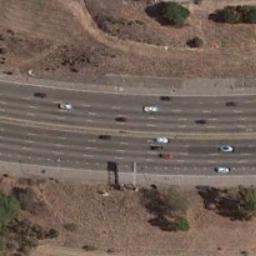} &
			\includegraphics[width=0.12\textwidth]{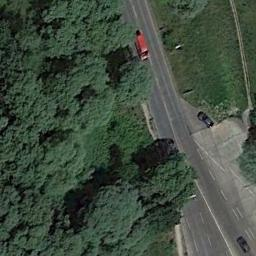} &
			\includegraphics[width=0.12\textwidth]{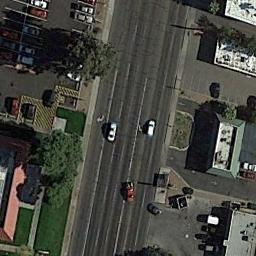} &
			\includegraphics[width=0.12\textwidth]{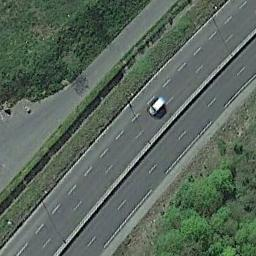} \\
			
			\includegraphics[width=0.12\textwidth]{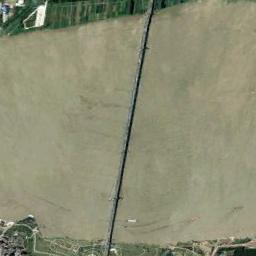} &
			\includegraphics[width=0.12\textwidth]{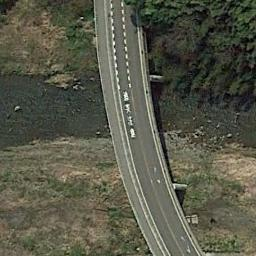} &
			\includegraphics[width=0.12\textwidth]{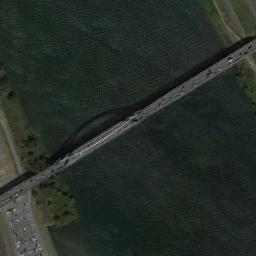} &
			\includegraphics[width=0.12\textwidth]{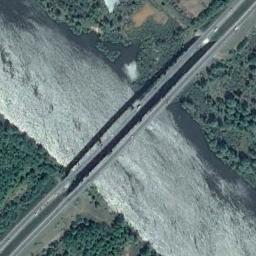} &
			\includegraphics[width=0.12\textwidth]{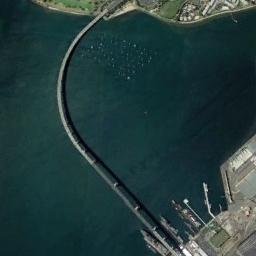} &
			\includegraphics[width=0.12\textwidth]{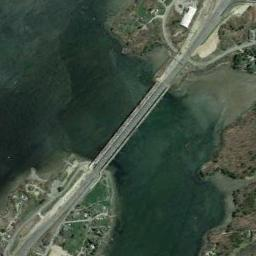} \\
			
			\includegraphics[width=0.12\textwidth]{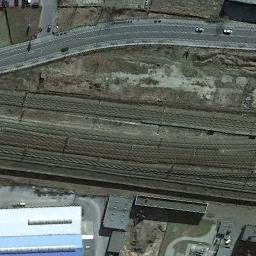} &
			\includegraphics[width=0.12\textwidth]{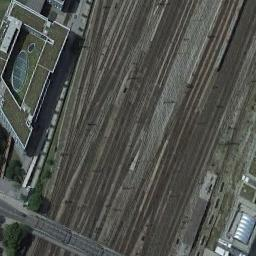} &
			\includegraphics[width=0.12\textwidth]{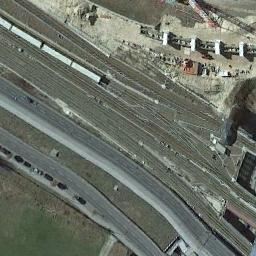} &
			\includegraphics[width=0.12\textwidth]{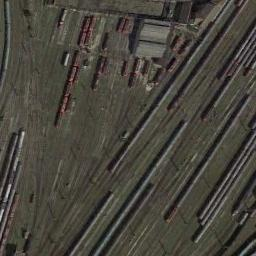} &
			\includegraphics[width=0.12\textwidth]{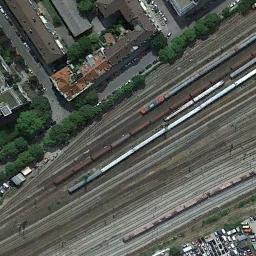} &
			\includegraphics[width=0.12\textwidth]{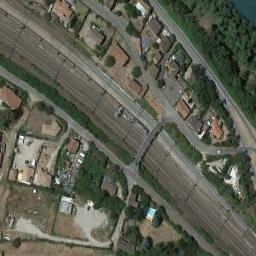} \\
			
		\end{tabular}
		\caption{Six samples of five categories of NWPU-RESISC45 dataset \cite{ref36}. Top to bottom illustrate Airport, Overpass, Freeway, Bridge, and Railway.}
		\label{fig:NWPU}
	\end{figure*}

	\subsubsection{Estimate Anchors}	
	In the object detection task, multi-scale processing must be considered to detect objects of varying sizes. In YOLO object detection, determining the characteristic of anchor boxes, including number, shape, and scale, dramatically affects the final results. In other words, it is a training hyper-parameter that requires empirical analysis. In this work, the k-means clustering algorithm and the Intersection-Over-Union ($IOU$) distance metric, invariant to the size of boxes, are used for this aim. The $IOU$ metric is calculated by Eq.  \ref{eq:iou}:
	\begin{equation}
		IOU = \frac{Area \: of \: Overlap}{Area \: of \: Union} = \frac{area(A \cap B)}{area(A \cup B)}
		\label{eq:iou}
	\end{equation}
	where $A$ and $B$ are bounding boxes of candidate anchors and grand-truth, respectively.
		
	The mean $IOU$ of the boxes in each cluster is a metric to judge the quality of anchor boxes. The boxes of similar aspect ratios and sizes are clustered together with the help of this technique. In summary, the number of anchor boxes provides a trade-off between detection speed and accuracy.
	
	Figure. \ref{fig:anchor}.(a) demonstrates box area versus aspect ratio for Airport bounding boxes. It can be seen from the chart that some groups have similar shapes and sizes, but the majority of them are spread out. Hence, instead of manually choosing the anchor boxes, a mentioned strategy is employed to categorize similar boxes with the help of a meaningful metric. The result of this process for different anchor boxes (different $k$) is illustrated in Figure. \ref{fig:anchor}.(b). This indicates that the mean IOU value greater than 0.5 guarantees that the grand truth has admissible overlap with anchor boxes. Although the mean IOU measure can be improved, the computation cost and over-fitting challenges appear by increasing the number of anchor. Nevertheless, the optimum number of anchor boxes is a value between two and ten. The mean IOU greater than 0.5 appears for two anchors, and more than ten anchor boxes yield slight improvement. Hence, six anchor boxes are considered in this work based on experimental results and evaluations. Finally, the specified anchor boxes are candidates in both pre-trained models' detection heads.
	
		\begin{figure*}[t!]
		\center
		\setlength{\tabcolsep}{2pt}
		\begin{tabular}{ccccc}
			\includegraphics[width=0.15\textwidth]{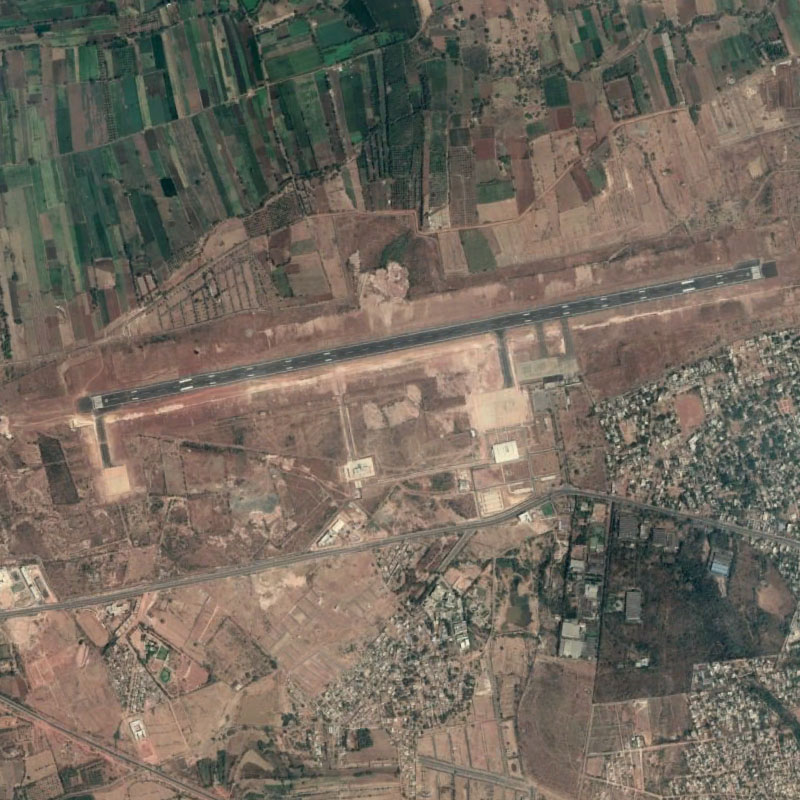} &
			\includegraphics[width=0.15\textwidth]{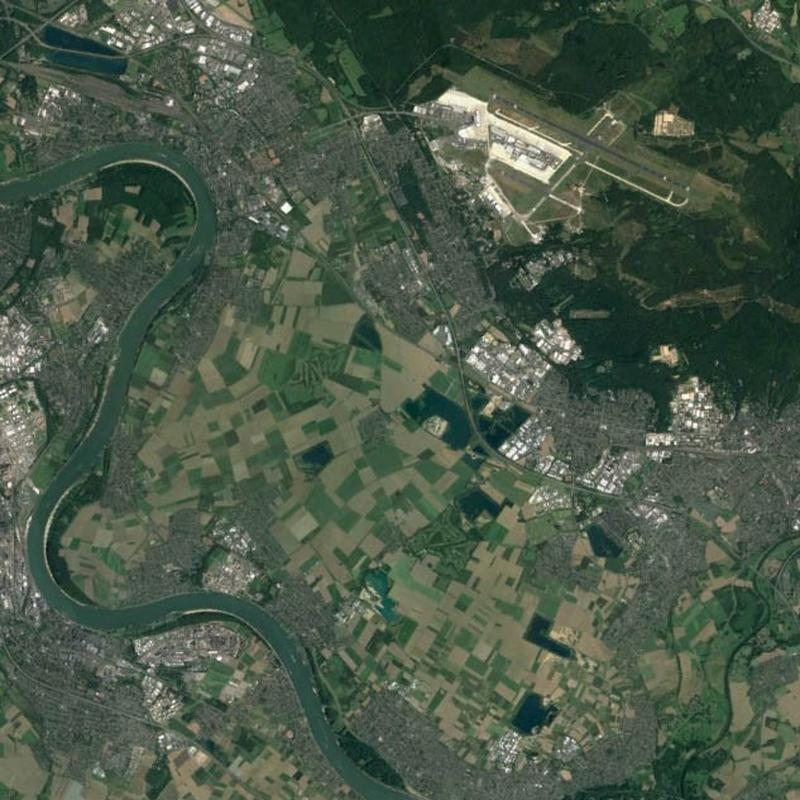} &
			\includegraphics[width=0.15\textwidth]{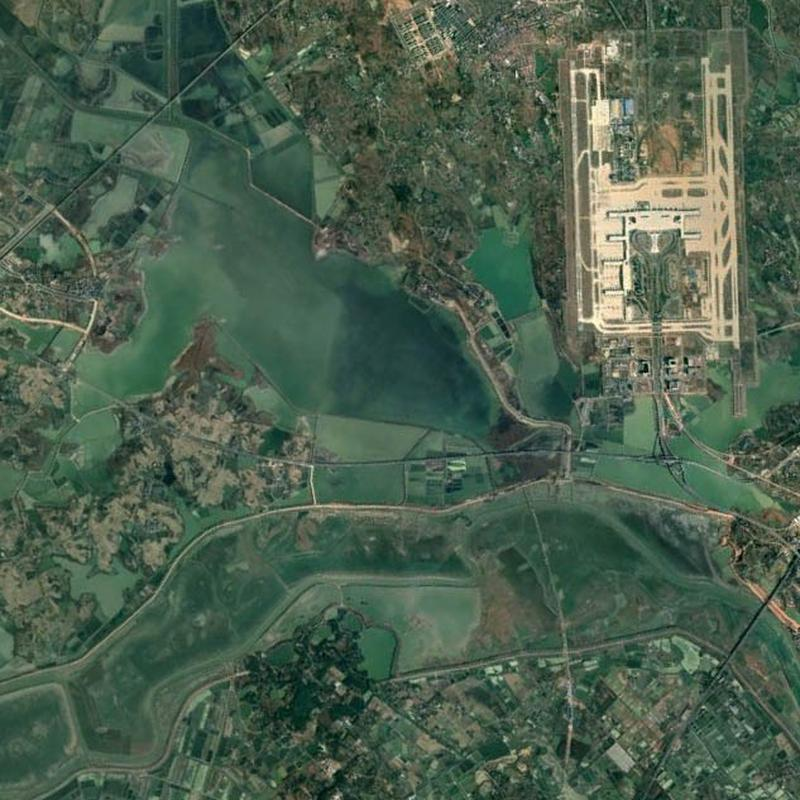} &
			\includegraphics[width=0.15\textwidth]{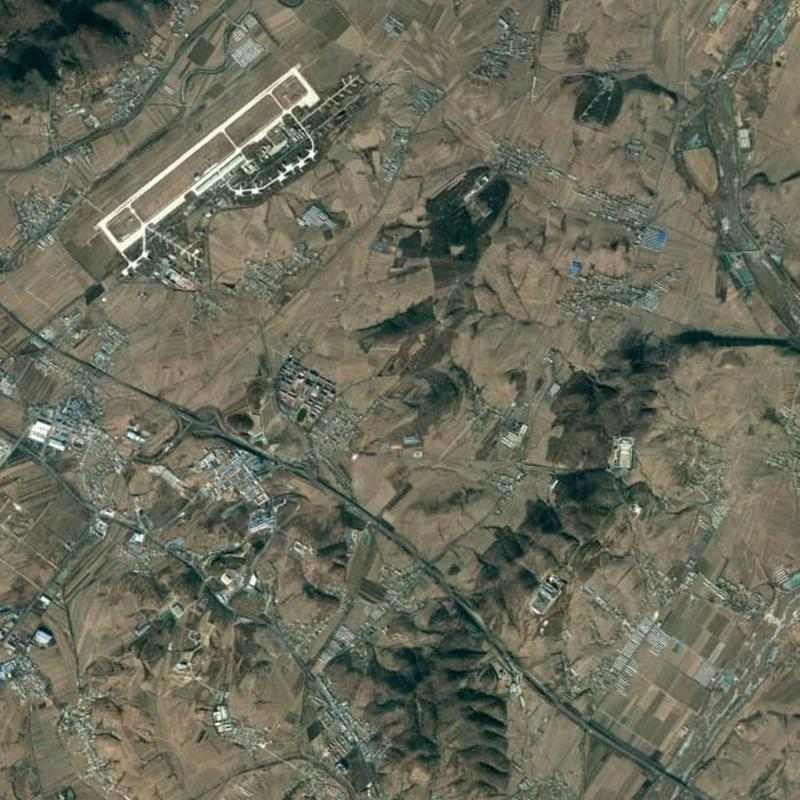} &
			\includegraphics[width=0.15\textwidth]{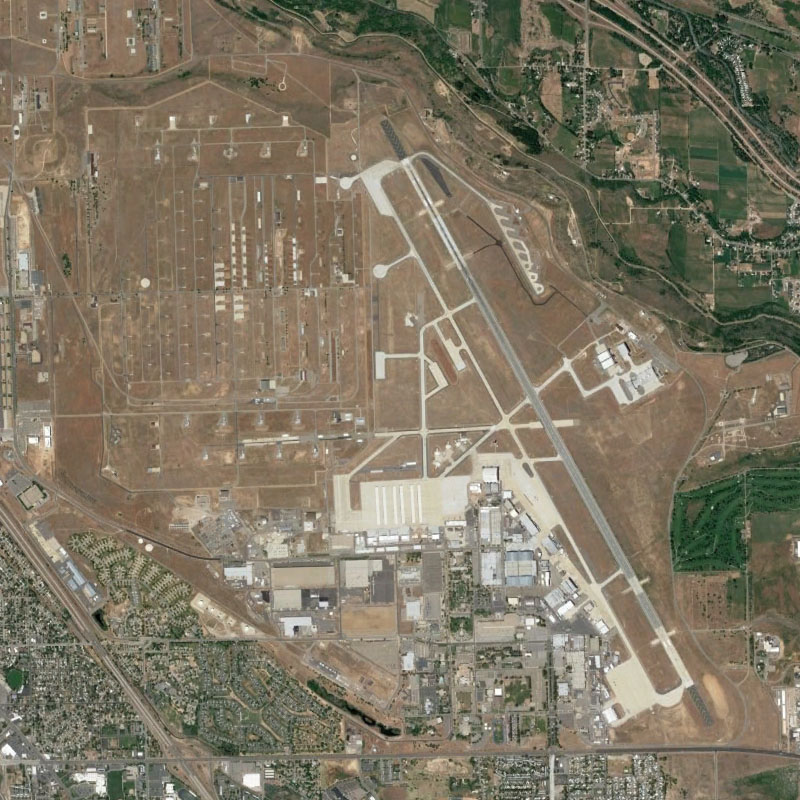} \\
			\includegraphics[width=0.15\textwidth]{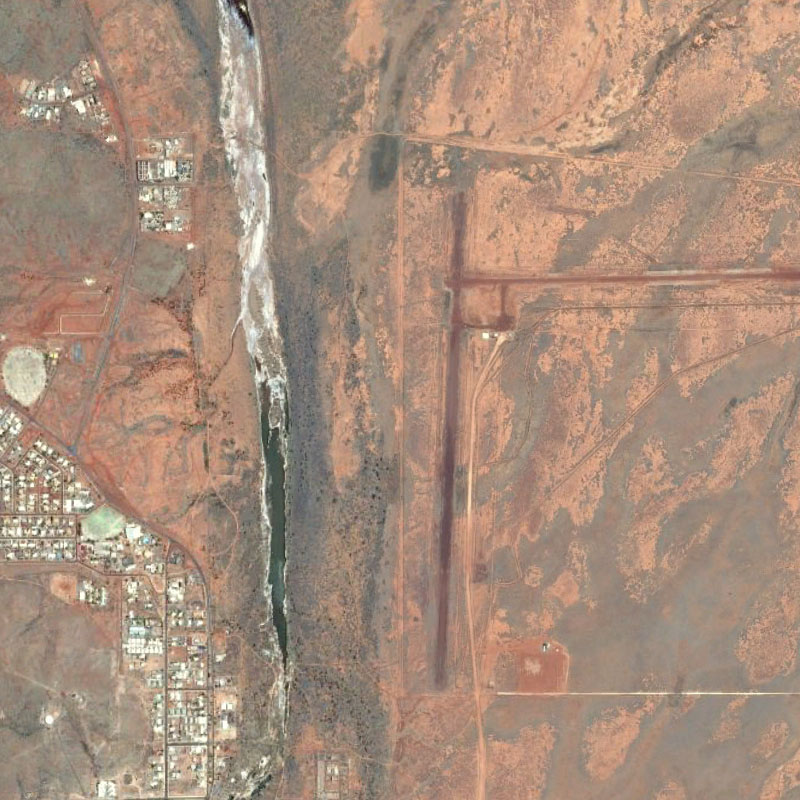} &
			\includegraphics[width=0.15\textwidth]{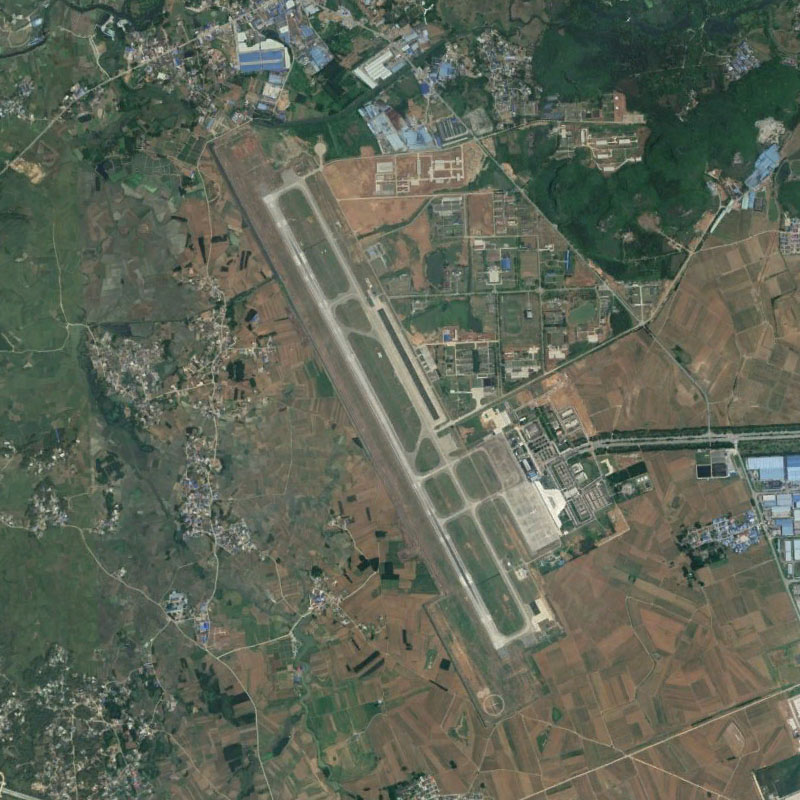} &
			\includegraphics[width=0.15\textwidth]{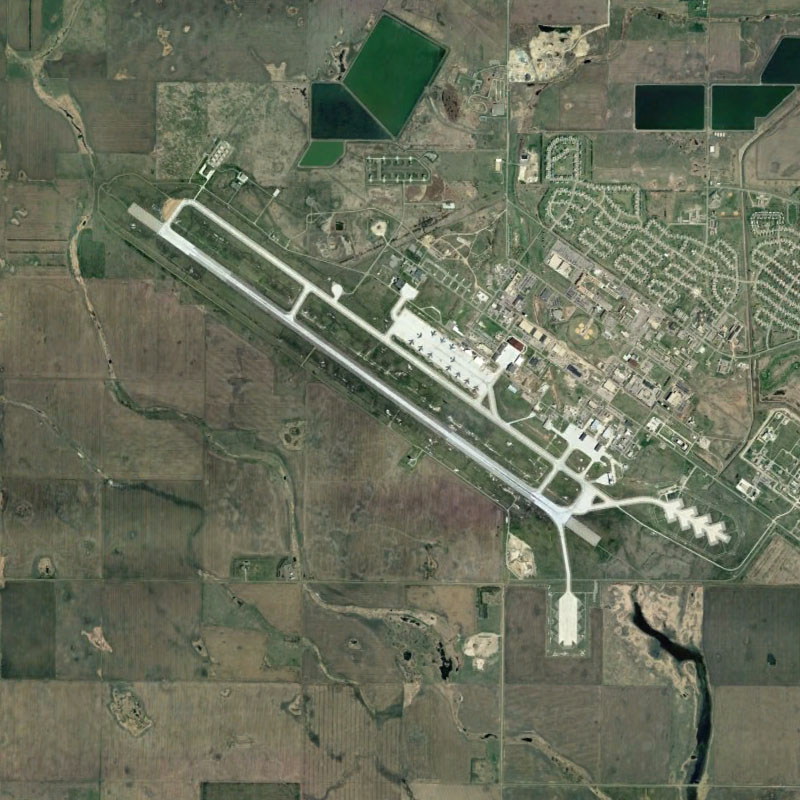} &
			\includegraphics[width=0.15\textwidth]{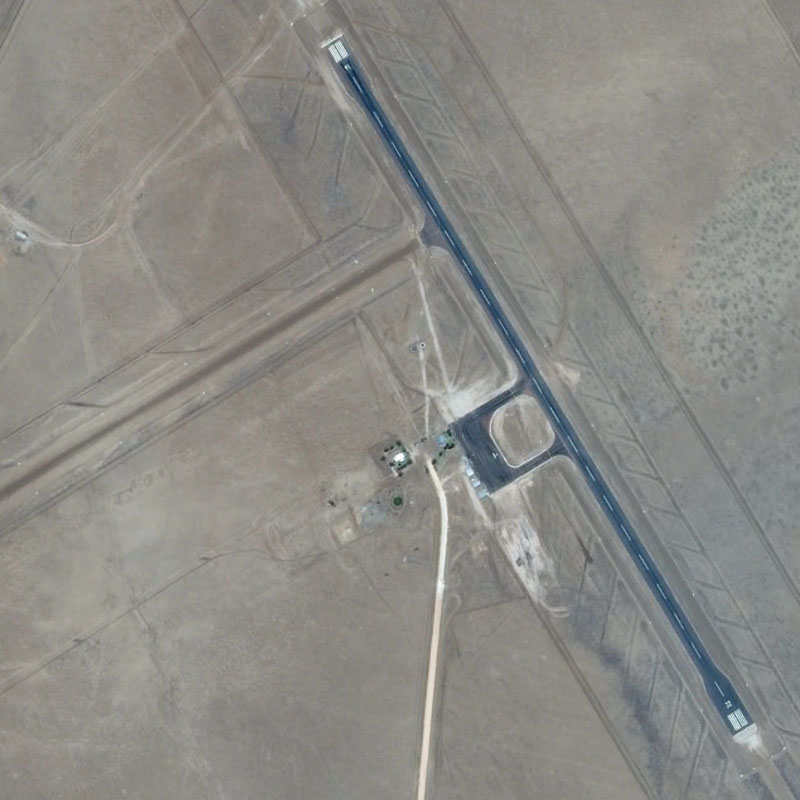} &
			\includegraphics[width=0.15\textwidth]{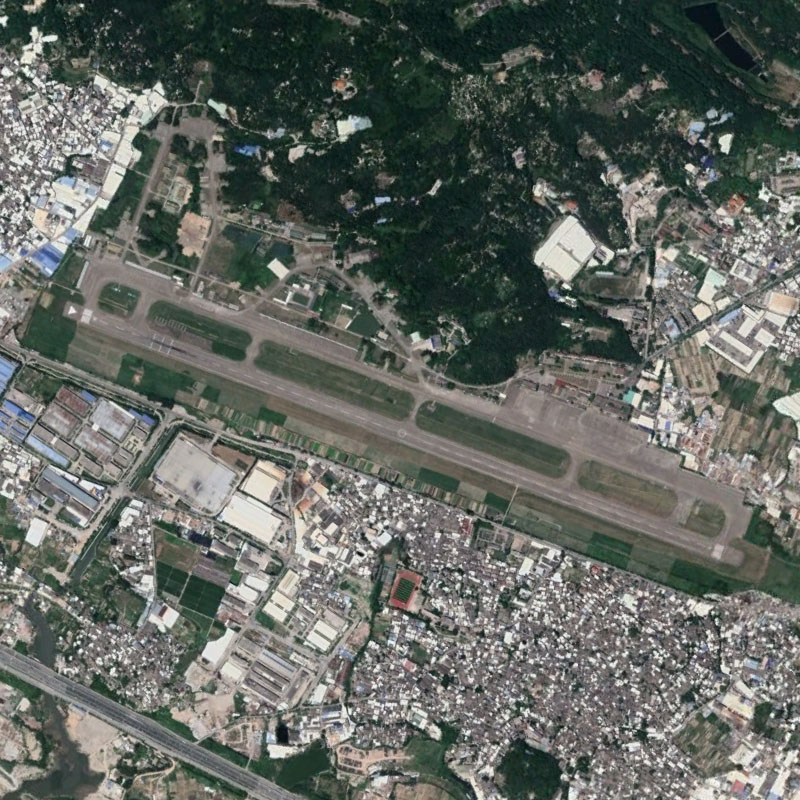} \\
			\includegraphics[width=0.15\textwidth]{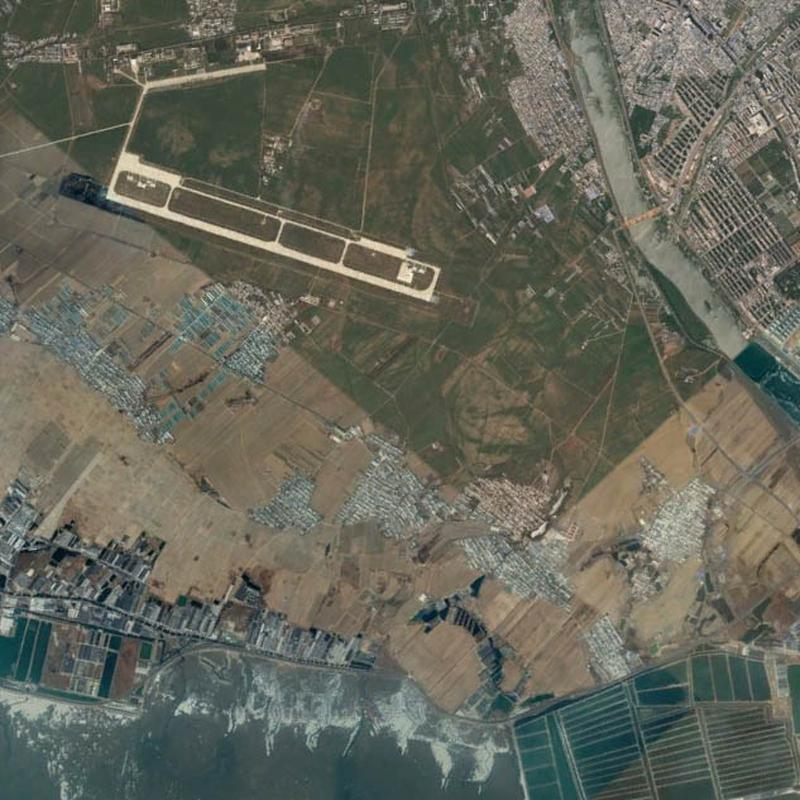} &
			\includegraphics[width=0.15\textwidth]{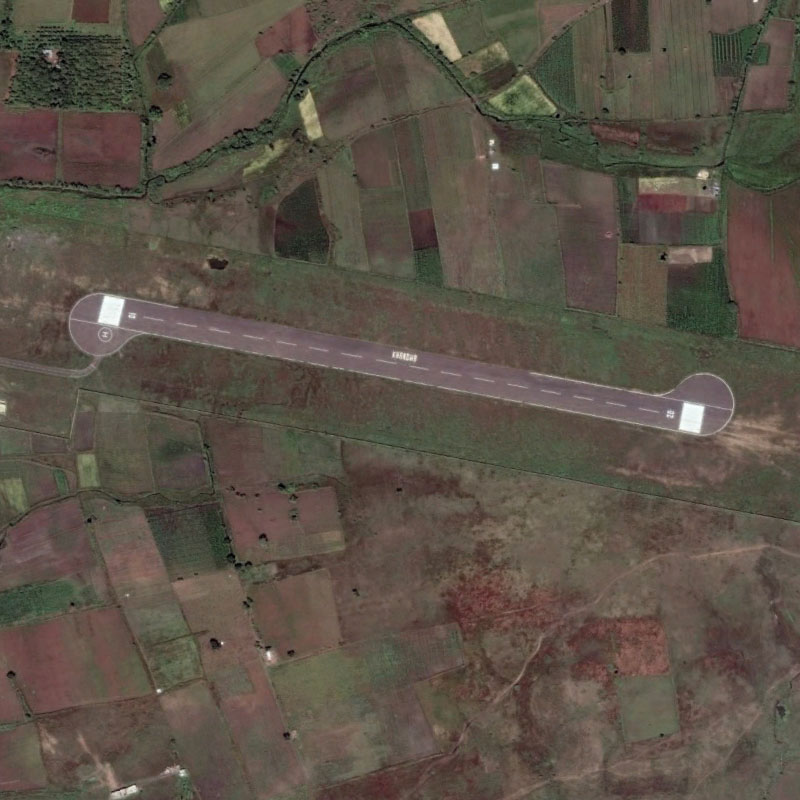} &
			\includegraphics[width=0.15\textwidth]{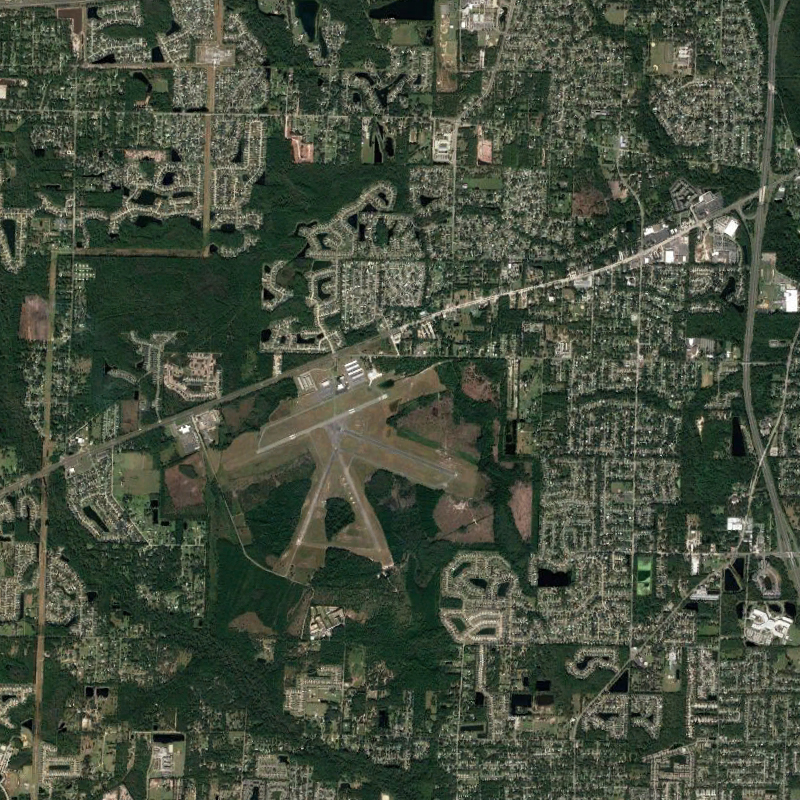} &
			\includegraphics[width=0.15\textwidth]{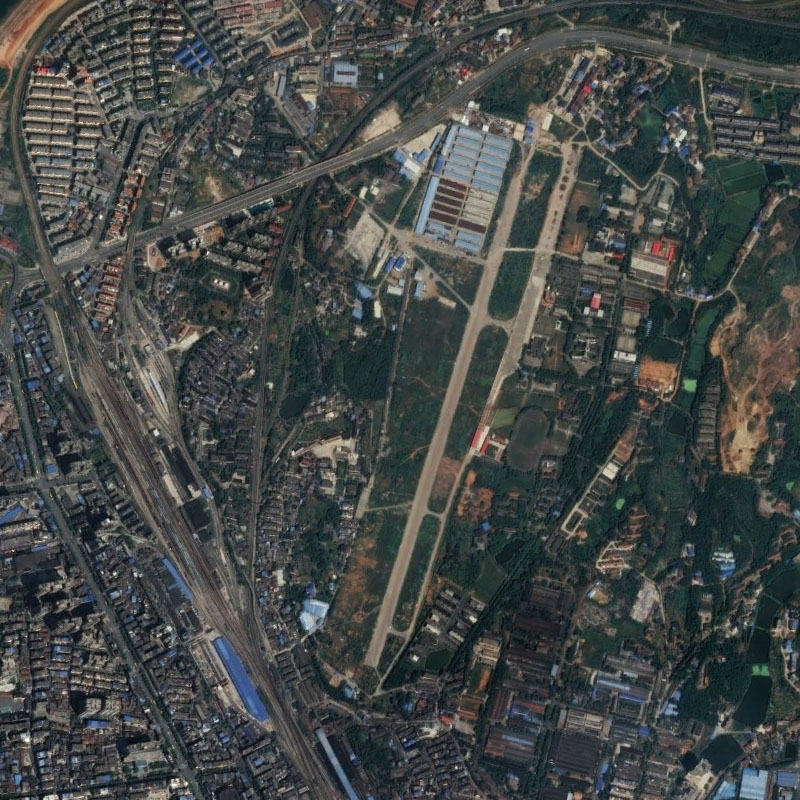} &
			\includegraphics[width=0.15\textwidth]{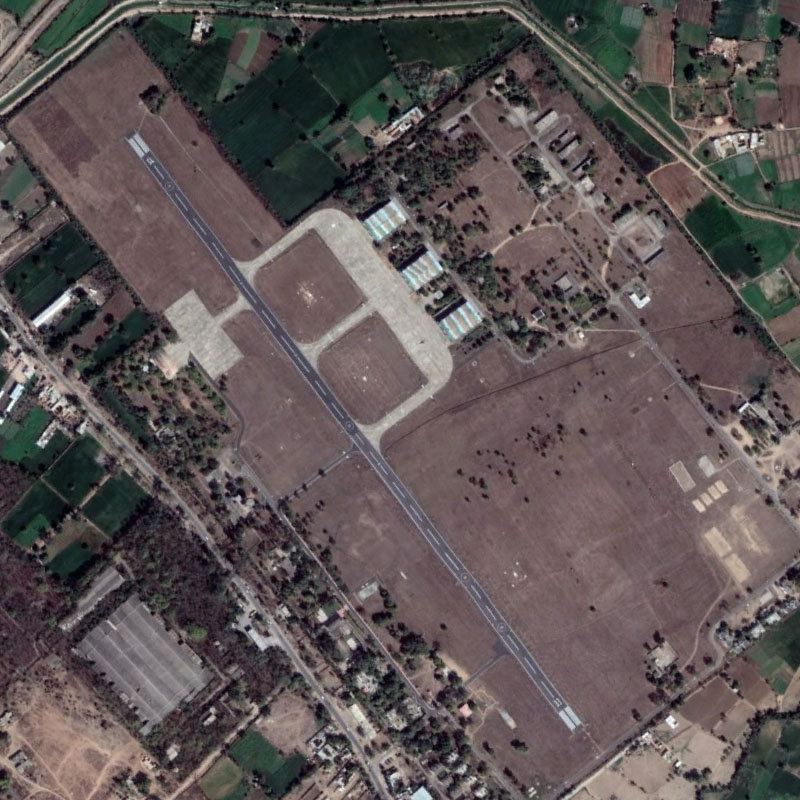} \\
		\end{tabular}
		\caption{Fifteen examples of airport images which randomly selected from DIOR Dataset\cite{ref5}.}
		\label{fig:dior}
	\end{figure*}
	\begin{figure}[t!]
		\center
		\includegraphics[width=0.5\textwidth]{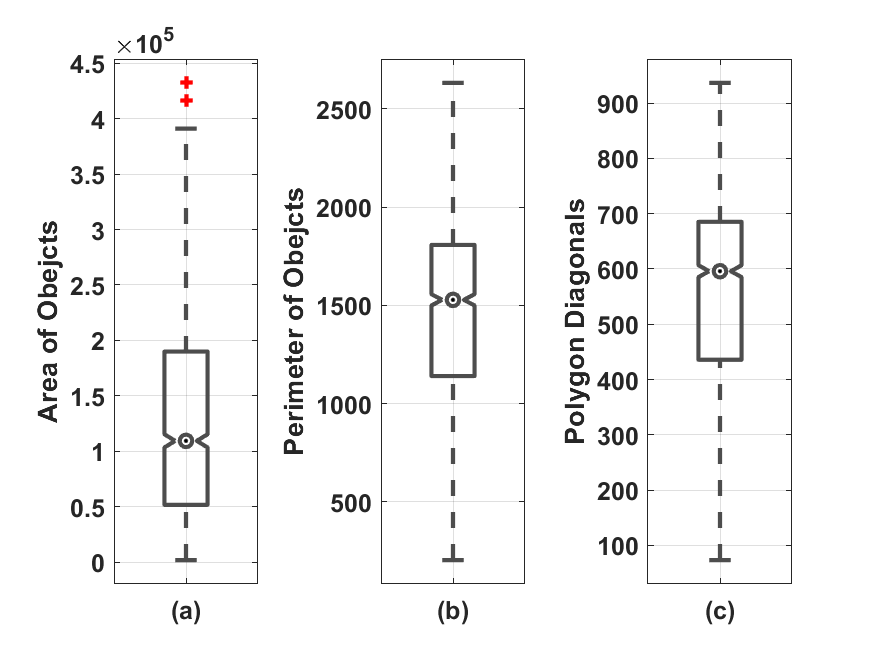} 
		\caption{The graphical distribution of airports using box-plot. (a) Locality, (b) Spread, and (c) Skewness.}
		\label{fig:boxplot}
	\end{figure}
		
		\begin{figure*}[t!]
		\center
		\setlength{\tabcolsep}{2pt}
		\begin{tabular}{ccccc}
			\includegraphics[width=0.15\textwidth]{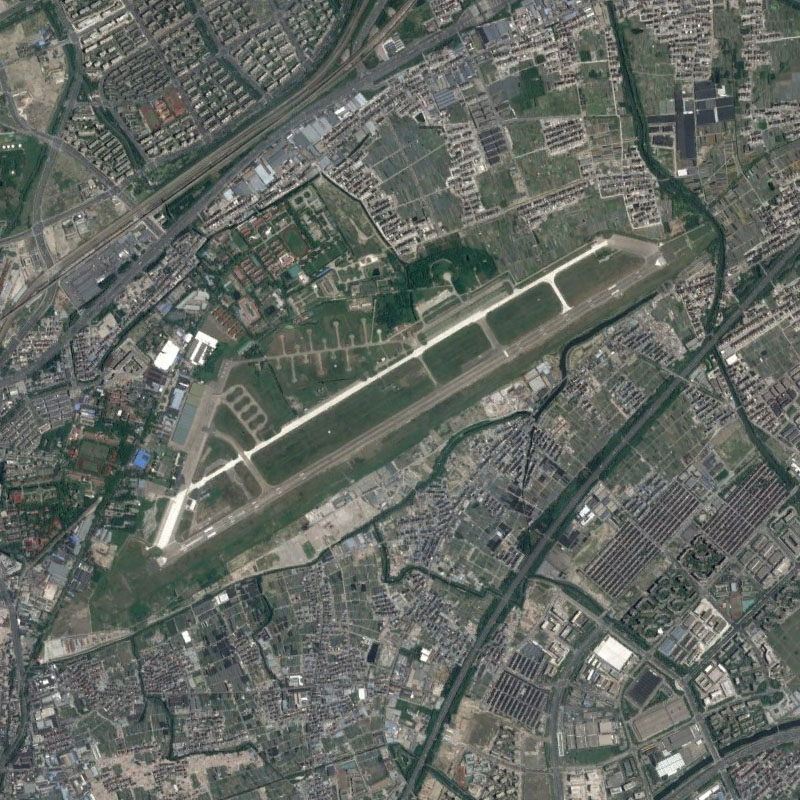} &
			\includegraphics[width=0.15\textwidth]{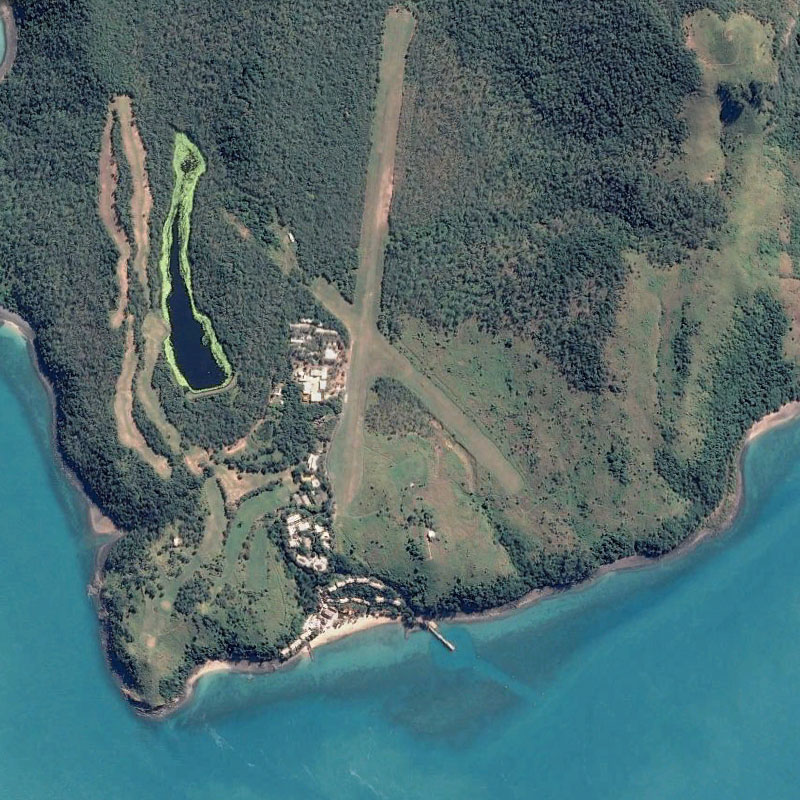} &
			\includegraphics[width=0.15\textwidth]{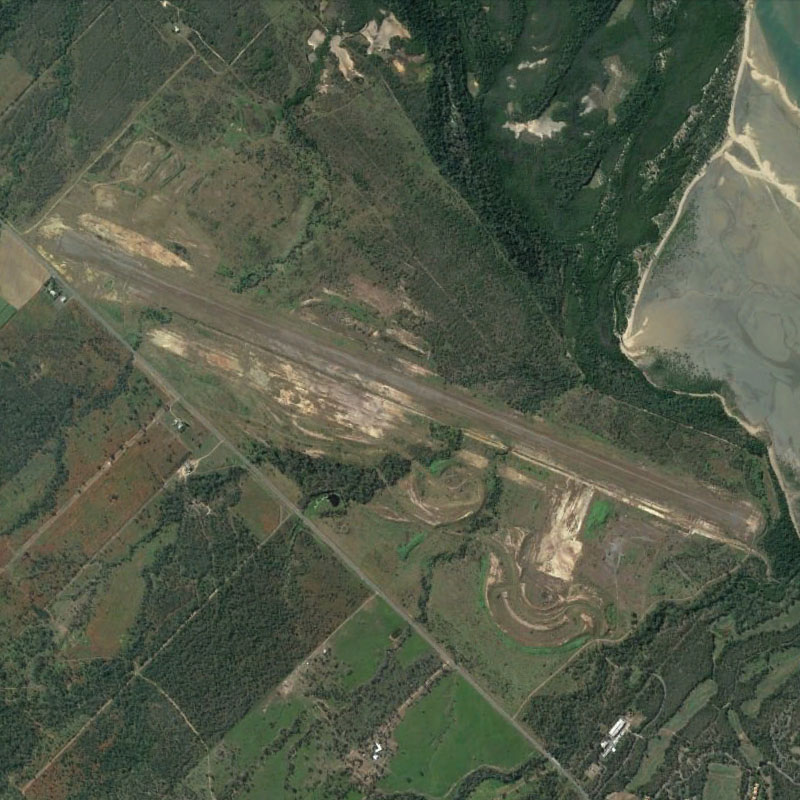} &
			\includegraphics[width=0.15\textwidth]{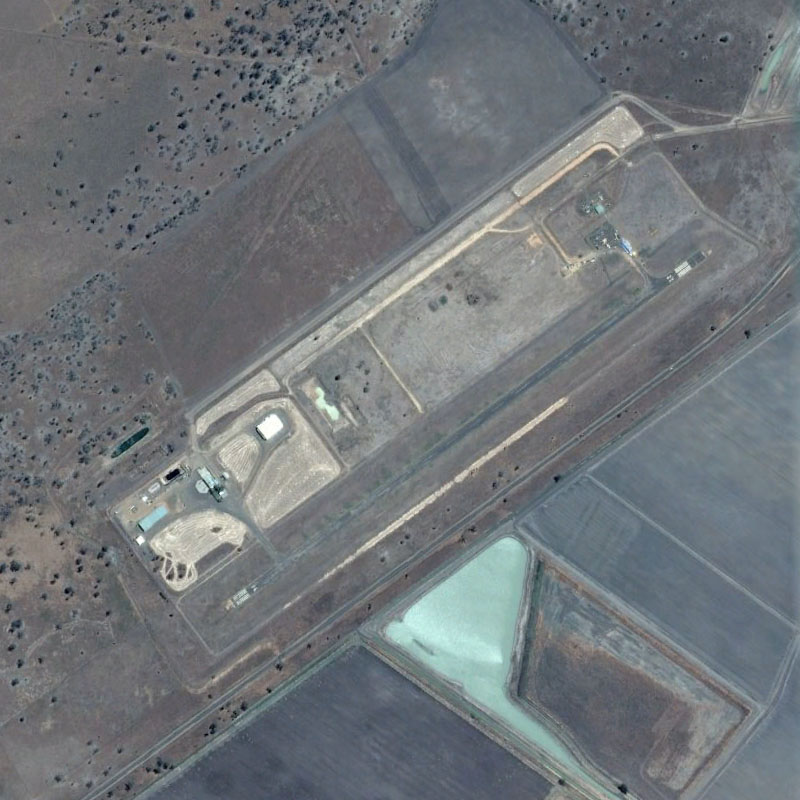} &
			\includegraphics[width=0.15\textwidth]{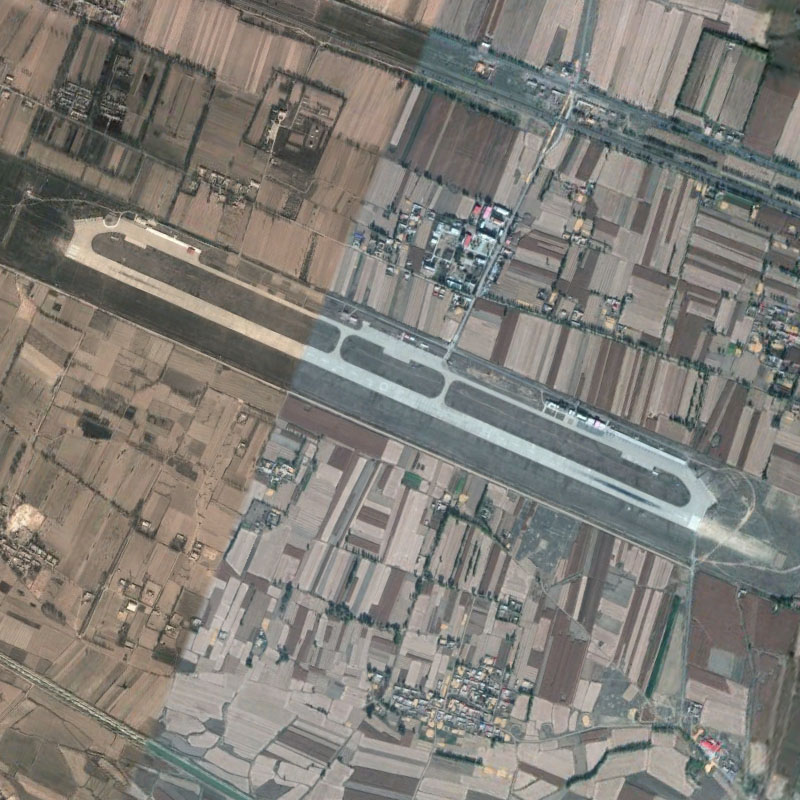} \\
			\includegraphics[width=0.15\textwidth]{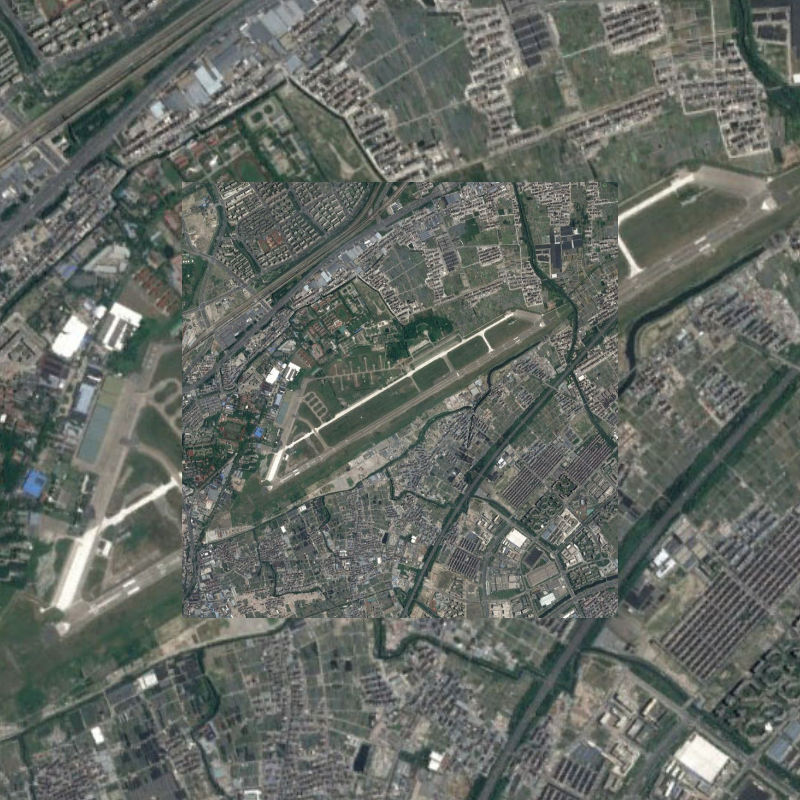} &
			\includegraphics[width=0.15\textwidth]{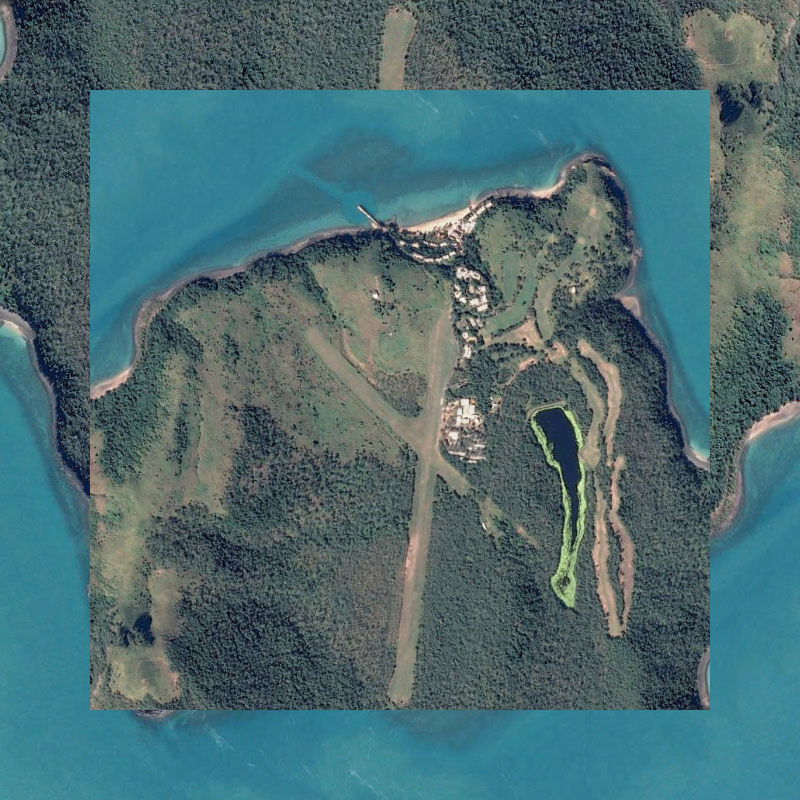} &
			\includegraphics[width=0.15\textwidth]{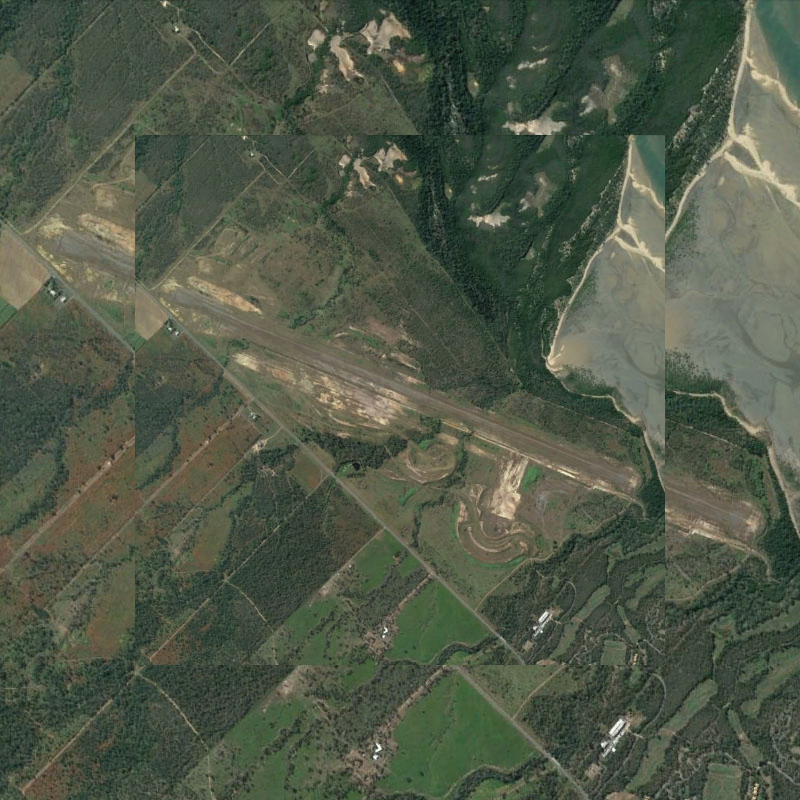} &
			\includegraphics[width=0.15\textwidth]{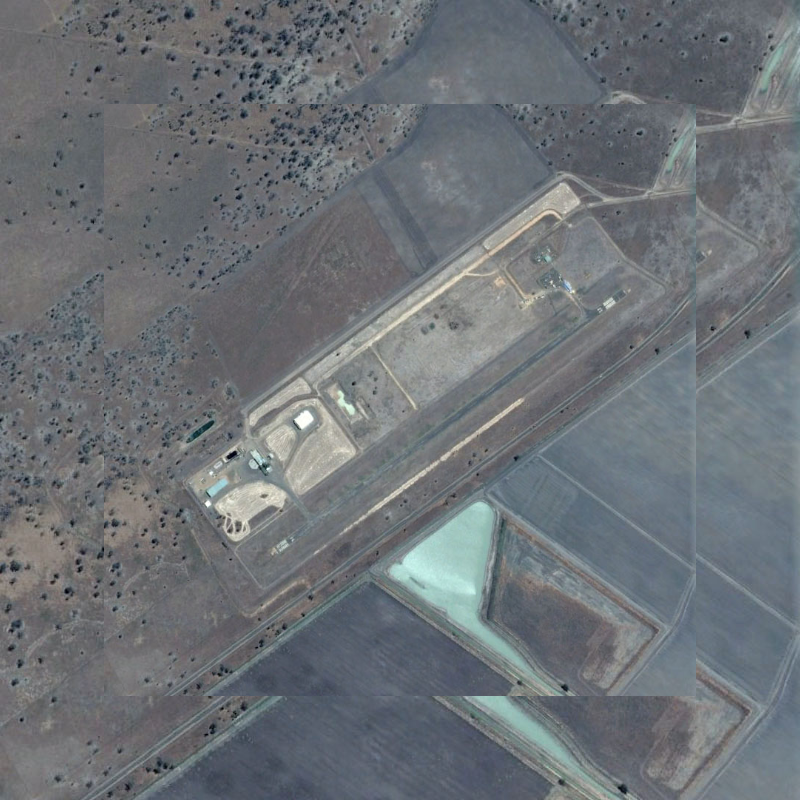} &
			\includegraphics[width=0.15\textwidth]{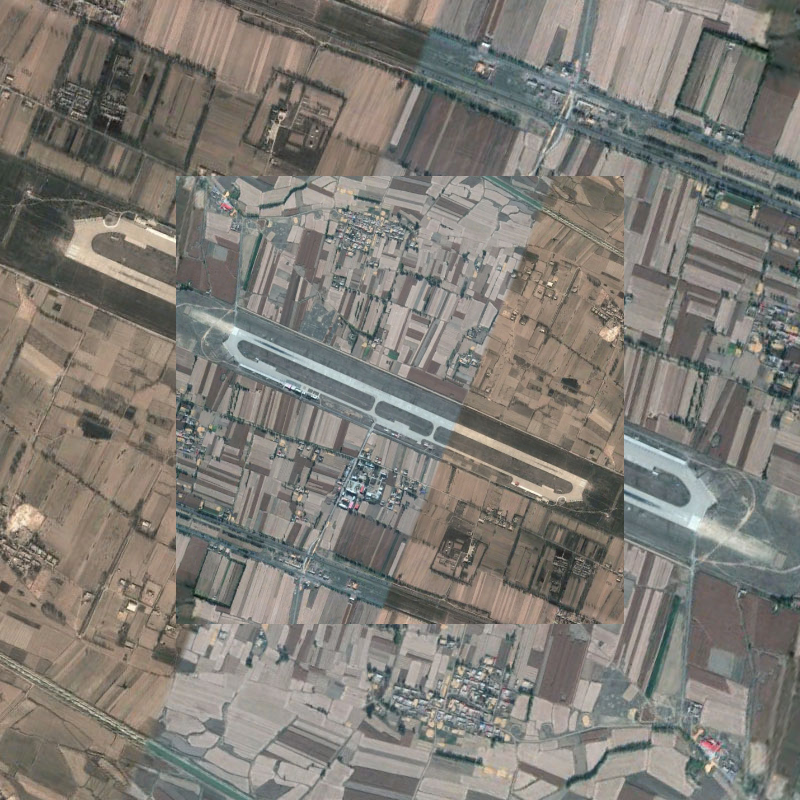} \\
		\end{tabular}
		\caption{The results of the presented image augmentation strategy in terms of downscale.}
		\label{fig:augmentation}
	\end{figure*}
	
	\subsection{Object Detection Phase}		
	After preprocessing and constructing the architect of the network, the main phase of EYNet is detailly explained in this sub-section. In this way, the presented augmentation and hard example mining strategies play essential roles.

	\subsubsection{Main Core}	
	The central core of EYNet consists of training and validation stages and their details. In this way,  rmsprop optimizer is selected due to the effective performance compared to Stochastic Gradient Descent (SGD) and Adam solvers. Also, data permutation is shuffled in each epoch to form mini-batches. Moreover, as illustrated in Figure. \ref{fig:digram}, the connection status and the structure of the detection source of EYNet are designed in a way to gain more effective and useful features from both Models. The transpose convolution layer, which upsamples the feature map, boosts the performance compared to the usual convolution in standard YOLO.
	
		\begin{algorithm}[t]
		\caption{The process of image augmentation.}
		\label{ALG:augmentation}
		\textbf{Input:} Mini-Batch, Cyclical Value, Type\\
		\textbf{Output:} Augmented Image
		\begin{algorithmic}[1]
			\Procedure{AugmentData}{$x$, $\gamma_t$, $\lambda$}													
			\If {$\lambda == 0$} \Comment Normal Mining  
			\If {$\zeta - \gamma_t \leq rand(0, 1)$} 
			\\ \Comment $\gamma_t$ is WarmRestart(cyclical) $min=0, max=1$
			\\ \Comment $\zeta$ is an adjustment parameter for augmentation
			\State 	$\psi$ = RandAffine2d($x$) 
			\State 	\Return FillMissing($\psi$) \Comment Ex. Figure. \ref{fig:augmentation}
			\Else
			\State 	\Return $x$	\Comment No Augmentation	
			\EndIf		
			\Else \Comment Hard (Negative) Example Mining                     
			\State 	\Return RandAdjustsColor($x$)	
			\EndIf					
			\EndProcedure
		\end{algorithmic}
	\end{algorithm}	
	
	In this work, the Huber loss function\cite{ref40} is employed for computing the loss of bounding boxes instead of using Mean Squared Error(MSE). This modification in standard YOLO extremely boosts the performance of final results. Simple computing and being less sensitive to outliers are noteworthy characteristics of this loss. This function for robust regression tasks is computed by Equation. \ref{eq:huber}:
	\begin{equation}
		L_i =
		\begin{cases}
			\frac{1}{2}(Y_i-Y'_i)^2 & \quad \text{if} |Y_i-Y'_i|\leq \delta \\
			\delta |Y_i-Y'_i|-\frac{1}{2}\delta^2 & \quad \text{otherwise}
		\end{cases}    
		\label{eq:huber}
	\end{equation}    
	where $Y_i$ and $Y'_i$ are the target value and the corresponding value in prediction step, respectively. Also, $\delta$ is a transition point where the loss transitions from a quadratic function to a linear function.
			
	On the other hand, a Warm Restart (Cyclical) \cite{ref37} learning rate strategy is employed in the EYNet training stage. In this way, a Cosine function is utilized for the learning rate schedule. In EYNet, the optimizer (solver) uses the learning rate given by shifted Cosine function for each iteration. EYNet only used a cycle in the whole training stage. In other words, the process starts with a large learning rate and reaches the minimum value at the beginning and ending steps of the learning stage, respectively. In this period, the learning rate decays monotonically in each iteration. The Warm Restart (Cyclical) is computed by Equation. \ref{eq:warm}:
	\begin{equation}
		\eta_t	= \eta_{min} + \frac{1}{2}\big(\eta_{max} - \eta_{min}\big)\big(1+cos(\frac{T_{c}}{T}\pi)\big)
		\label{eq:warm}
	\end{equation}
	where $T$ is the number of iterations per snapshot and $T_{c}$ represents the number of epochs since the last restart(Current iteration). Moreover, $\eta_{min}$ and $\eta_{max}$  are ranges for the learning rate. It should be noted, although with the help of this kind of strategy, the method does not require a separated validation set, around 10\% of data are considered for this aim. Also, a small learning rate guarantees that the common features previously learned in the transfer learning phase can be involved in the current phase. The form of a decrease in the learning rate is shown in Figure. \ref{fig:warm}. 
	
	\begin{table}[t!]
		\footnotesize
		\caption{The type and details of augmentation used in both the transfer learning phase and EYNet.}
		\label{TABLE:augmentation}
		\renewcommand{\arraystretch}{1.5}
		\scalebox{1} {
			\begin{tabular*}{\columnwidth}{@{\extracolsep{\fill}}l@{}c@{}c}
				\cline{1-3}
				Type&Value&Description\\
				\cline{1-3}
				Saturation&[-0.5, 0.5]&EYNet\\
				Random Rotation&[-180, 180]&Transfer Learning\\
				Random XReflection&True&Both phases\\
				Random YReflection&True&Both phases\\
				Random XScale&[0.5, 1.5]&Both phases\\
				Random YScale&[0.5, 1.5]&Both phases\\					
				\cline{1-3}
		\end{tabular*}}
	\end{table}		
	\begin{table}[t!]
		\footnotesize
		\caption{The hyperparameters used in the experiments for both MobileNet and ResNet18 in transfer learning phase.}
		\label{TABLE:hyperparameterstrans}
		\renewcommand{\arraystretch}{1.5}
		\scalebox{1} {
			\begin{tabular*}{\columnwidth}{@{\extracolsep{\fill}}l@{}c@{}c}
				\cline{1-3}
				Hyper parameters&Value&Description\\
				\cline{1-3}
				(Train, Validation, Test)&(80\%, 20\%, 0\%)&-\\
				Mini-Batch Size & 16 &-\\				
				Epoch & 50 & -\\
				Solver Name & sgdm & momentum(0.9)\\
				Initial Learning Rate & 0.001 &-\\
				L2Regularization& 0.0001& -\\
				Learning Rate Schedule & Piecewise &-\\
				Learning Rate Drop Period& 10& -\\
				Learning Rate Drop Factor& 0.9&- \\
				Shuffle & Every Epoch &-\\	
				Validation Frequency&350& iteration\\				
				Validation Patience&10&stopping criteria\\
				ShearConv(in ResNet18)&5$\times$5 (6 Filters)&learning rate = 0\\					
				\cline{1-3}
		\end{tabular*}}
	\end{table}

	\begin{table}[t!]
		\footnotesize
		\caption{The hyperparameters used in the experiments for object detection phase based on EYNet.}
		\label{TABLE:hyperparameters}
		\renewcommand{\arraystretch}{1.5}
		\scalebox{1} {
			\begin{tabular*}{\columnwidth}{@{\extracolsep{\fill}}l@{}c@{}c}
				\cline{1-3}
				Hyper parameters&Value&Description\\
				\cline{1-3}					
				(Train, Validation, Test)&(60\%, 10\%, 30\%)&-\\					
				Input Layer Size&[224 224]&color\\
				Input Layer Normalization&Z-Score&-\\																				
				Epoch & 30 & stopping criteria\\					
				Mini-Batch Size & 8 &-\\								
				Learning Rate Schedule&Cosine Annealing(Cyclical)& snapshots = 1\\								
				Solver Name & RMSprop & momentum(0.9)\\						
				Initial Learning Rate & 0.001 & [0.0001, 0.001]\\										
				Validation Frequency&100& iteration\\				
				L2Regularization& 0.00001& -\\					
				Shuffle & Every Epoch &-\\	
				Bias Learning Rate&0&all layers\\
				Learning Rate Factor&10&detection layers\\
				Weights Initializer&He&detection layers\\
				Objectness Loss&Cross-Entropy&-\\
				Class Confidence Loss&Cross-Entropy&-\\					
				Box Offset Loss&Huber&-\\							
				Number of anchors&6&-\\
				Penalty Threshold&0.5&-\\							
				Detection Threshold&0.5&-\\							
				\cline{1-3}
		\end{tabular*}}
	\end{table}

	\subsubsection{Augmentation}	
	Data augmentation (enhancement) plays a notable role in object localization, detection, and recognition problems. This step is adapted to increase the accuracy and tolerate the low number of data in the training phase. Generally, more varieties of data can be generated with the help of this mechanism. The augmentation techniques can be broadly categorized into three groups: geometric transform, color adjustment, and image filtering. The employed techniques should be intelligently selected in this process based on the type of image. Some augmentations not only increase the complexity but also lead the poor result.
	
	\begin{figure*}[t!]
		\center
		\setlength{\tabcolsep}{2pt}
		\begin{tabular}{cc}
			\includegraphics[width=0.39\textwidth]{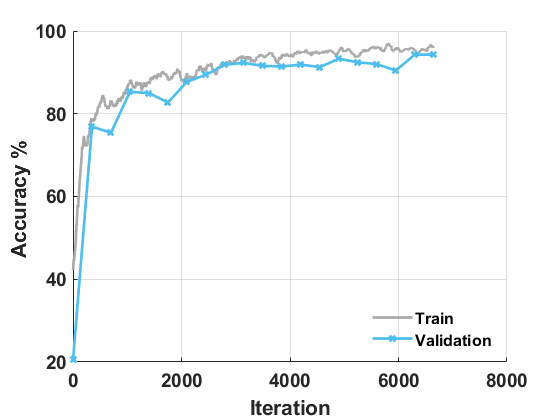} &
			\includegraphics[width=0.39\textwidth]{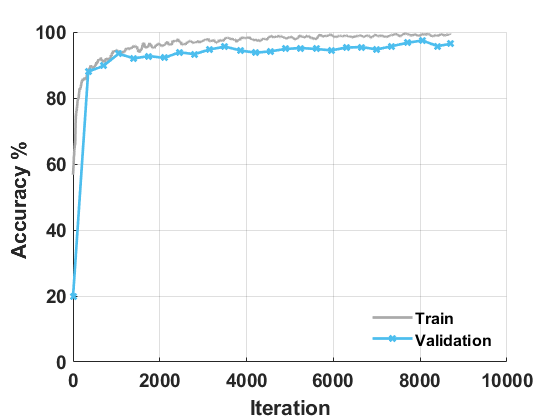} \\
			(a)&(b)\\
			\includegraphics[width=0.39\textwidth]{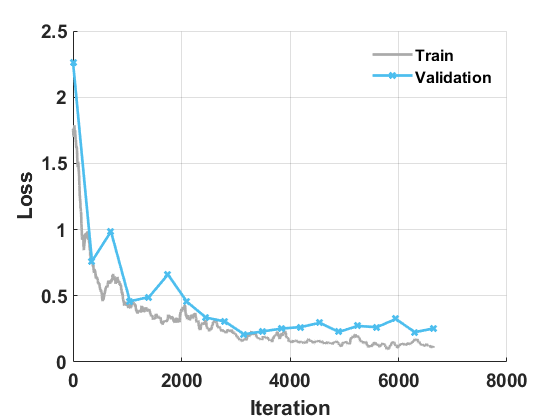} &
			\includegraphics[width=0.39\textwidth]{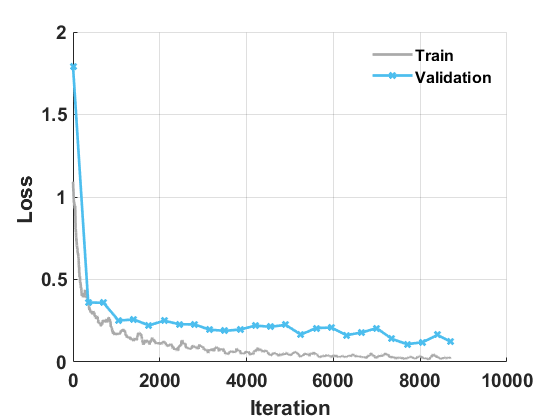} \\
			(c)&(d)\\
		\end{tabular}
		\caption{The Accuracy(Moving Mean = 100) and Loss of train and validation data. (a, c) MobileNet, (b, d) ResNet18 (+ShearLet Filters).}
		\label{fig:transferlearing}
	\end{figure*}
	
	\begin{figure*}[t!]
		\center
		\setlength{\tabcolsep}{2pt}
		\begin{tabular}{cc}
			\includegraphics[width=0.48\textwidth]{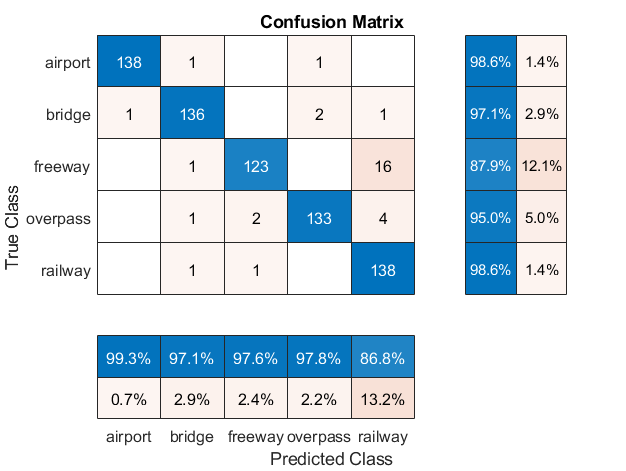} &
			\includegraphics[width=0.48\textwidth]{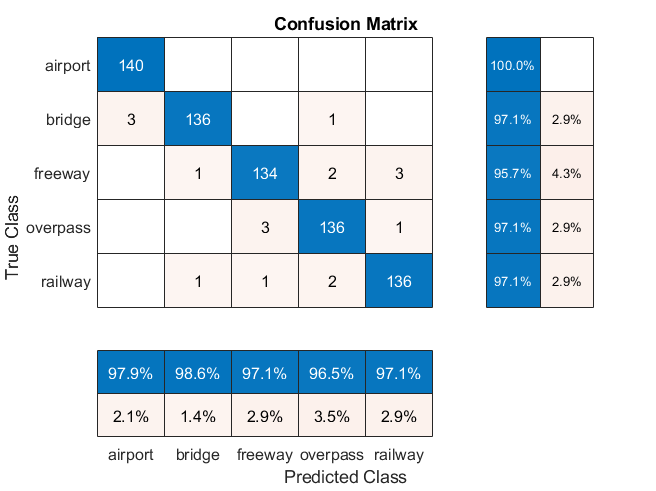} \\
			(a)&(b)\\
		\end{tabular}
		\caption{The confusion matrix of validation data. (a) MobileNet, (b) ResNet18 (+ShearLet Filters).}
		\label{fig:transferlearing2}
	\end{figure*}

	\begin{figure*}[t!]
		\center
		\setlength{\tabcolsep}{2pt}
		\begin{tabular}{cc}
			\includegraphics[width=0.49\textwidth]{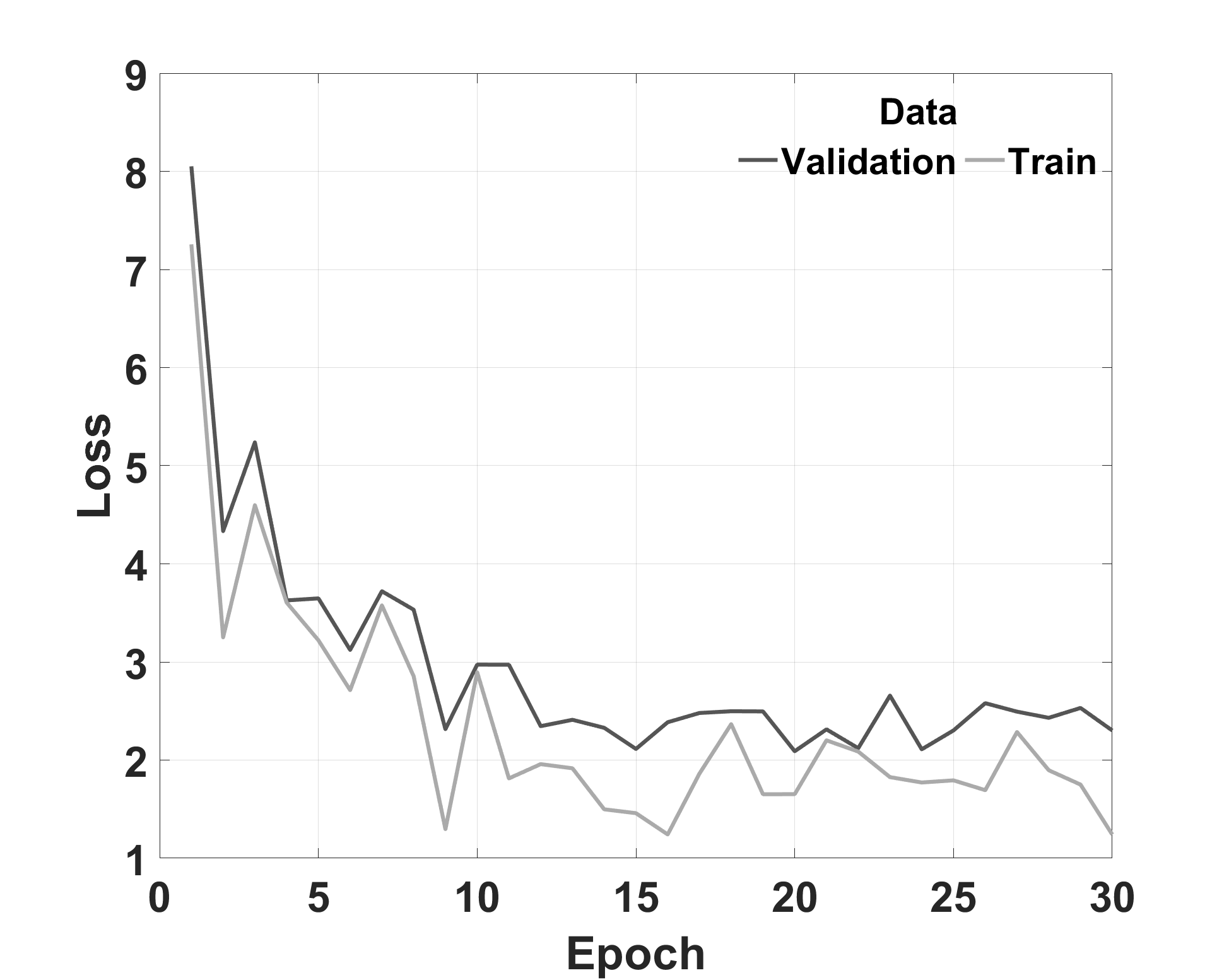} &
			\includegraphics[width=0.49\textwidth]{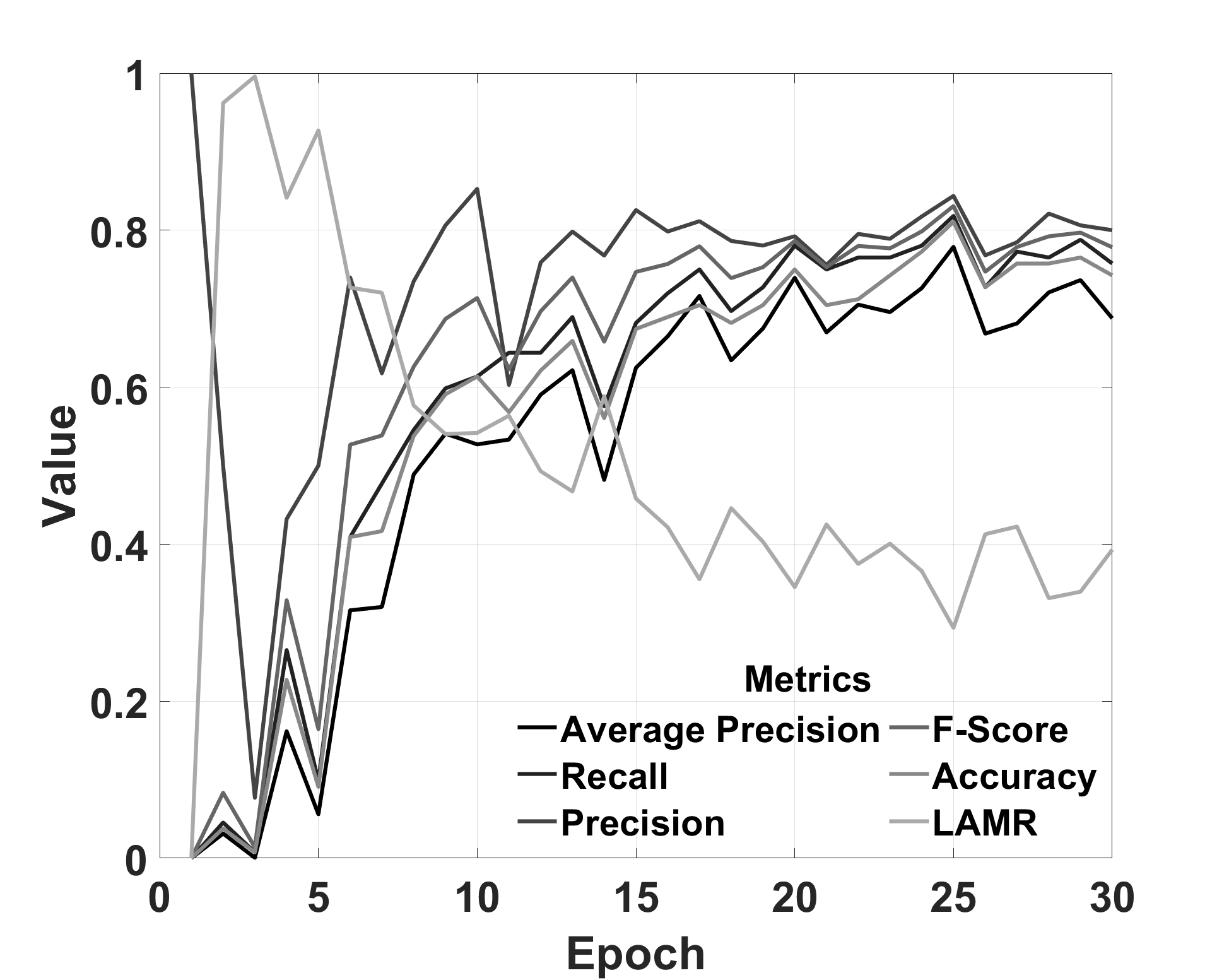}\\
			(a)&(b)\\
		\end{tabular}
		\caption{The performance of EYNet during the training phase. (a) The loss of train and validation data, (b) Quality measures on validation data.}
		\label{fig:valperf}
	\end{figure*}
	
	In EYNet, a dynamic augmentation is presented in the training phase. In this way, the WarmRestart (Cyclical) mechanism is performed to determine the probability of augmentation at the beginning and ending phase of training. The hyper-parameters of this function are similar to the Learning Rate step (Figure. \ref{fig:warm}), but the $min$ and $max$ values are zero and one, respectively. The hyper-parameter of the augmentation process named $\zeta$ is set $0.7$ to emphasize the augmented image (scaling and reflection) at the beginning of the training phase. On the other hand, the training phase is more concentrated on unmodified Mini-Batch at the final epochs. Meanwhile, the Hard (Negative) samples are trained by adjusting the color component. As demonstrated in Figure. \ref{fig:augmentation}, the generated black pixels outside the input image after down-sampling are filled by the original pixel of the corresponding images.
	
	For clarity, the pseudo-code of the augmentation process is illustrated in the Algorithm. \ref{ALG:augmentation}.

	\begin{figure*}[t!]
	\center
	\setlength{\tabcolsep}{2pt}
	\begin{tabular}{cccc}
		\includegraphics[width=0.23\textwidth]{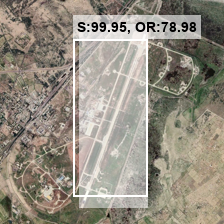} &
		\includegraphics[width=0.23\textwidth]{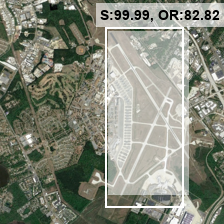} &
		\includegraphics[width=0.23\textwidth]{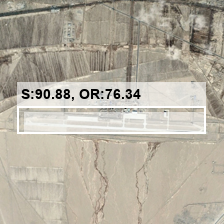} &
		\includegraphics[width=0.23\textwidth]{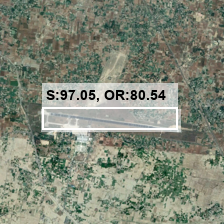}\\			
		
		\includegraphics[width=0.23\textwidth]{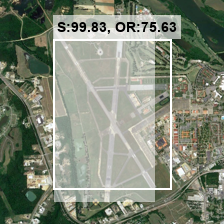} &
		\includegraphics[width=0.23\textwidth]{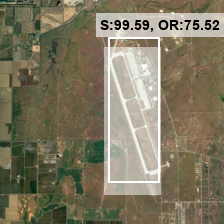} &
		\includegraphics[width=0.23\textwidth]{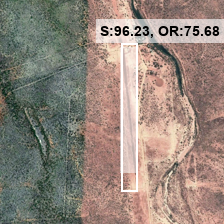} &
		\includegraphics[width=0.23\textwidth]{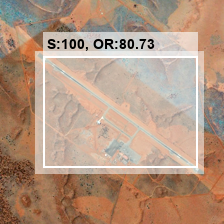}\\												
		\includegraphics[width=0.23\textwidth]{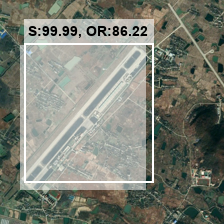} &
		\includegraphics[width=0.23\textwidth]{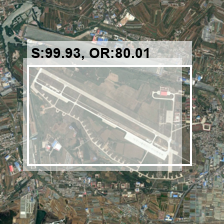} &
		\includegraphics[width=0.23\textwidth]{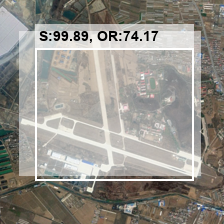} &
		\includegraphics[width=0.23\textwidth]{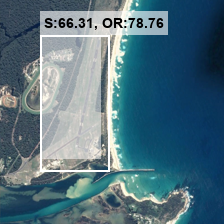}\\													
	\end{tabular}
	\caption{The airport detection result by EYNet. The bounding and highlighted boxes indicate the model outputs and grand-truth, respectively. Also, $S$ and $OR$ present the Score and Overlap Ratio, respectively.}
	\label{fig:airportdetectionvisual}
\end{figure*}

	\subsubsection{Hard (Negative) Example Mining}
	Besides the rest modification in the original YOLO, EYNet employs dynamic hard example mining to enhance the performance of models. This mechanism is introduced to reduce the redundant negative samples in the training procedure and overcome the imbalance problem between positive and negative samples. Generally, the discrimination ability of EYNet can be enhanced with the help of hard examples characterized by high loss values. 
	
	In detail, the concept of hard example mining is adapted to deform the negative samples by changing their color component. In EYNet, the hard samples are generated by Equation. \ref{eq:hard}:
	\begin{equation}
		\mu([l_{i-5}, l_{i-4}, ..., l_{i-1}]) + \sigma ([l_{i-5}, l_{i-4}, ..., l_{i-1}]) \leq l_i
		\label{eq:hard}
	\end{equation}
	where $\mu$ and $\sigma$ represent the moving mean and standard deviation, each is computed over a sliding window of length five across neighboring elements of the loss vector, respectively. Also, $l_i$ is the loss value of the current iteration.
	
	Afterward, in contrast to the mentioned augmentation, the color component of the current batch is modified by randomly adjusting the $Saturation$. In other words, only color augmentation is employed on all negative samples. Notice that the training hyperparameter is identical in this step.	
	
	The adopted modifications in the original YOLO improved the object expression capability and the training efficiency.		
	\section{Experimental Results}
	\label{sec:Experimental}	
	\subsection{Dataset and implementation details}
	In this research, two popular datasets named NWPU-RESISC45 \cite{ref36} and DIOR \cite{ref5} are utilized in transfer learning and object detection (EYNet) phases, respectively. 
	
	The first benchmark, widely employed in REmote Sensing Image Scene Classification (RESISC), contains a variety of class objects with various patterns, textures, homogeneous respect to color, etc. Each class of this dataset includes 700 images with a size of 256$\times256$ pixels in RGB color space. In the proposed scheme, five categories (including Airport, Overpass, Freeway, Bridge, and Railway) are chosen in the transfer learning phase. In other words, similar classes containing parallel lines and sharing both identical texture and structure compared to the primary problem are selected from the second dataset instead of employing a pre-trained network on irrelevant data (Ex. ImageNet). With the help of these strategies, the network's performance is boosted in terms of accuracy and time. Figure. \ref{fig:NWPU} shows six samples of each class from this benchmark. This dataset has three main characteristics: large scale, rich image variations, and high within-class diversity and between-class similarity. Moreover, the candidate images are collected under different weather, seasons, illumination, viewpoints, etc.
	
	On the other hand, EYNet employs the DIOR dataset to validate and prove the scheme's efficiency. This benchmark has 1,310 Airport samples with a size of 800$\times$800 pixels chosen through different classes in this work. Some samples of this data are illustrated in Figure. \ref{fig:dior}. As can be seen, the size of Airports is varied, which is more helpful for challenging real-world tasks. In addition, these samples demonstrate the different viewpoints, translation, illumination, background, object pose and appearance, occlusion, etc. 	The graphical distribution of airports including Locality, Spread, and Skewness is presented in Figure. \ref{fig:boxplot} using box-plot. Something which should be mentioned here is that, similar to the major dataset for remote sensing images, these datasets are gathered from Google Earth.
	
	Also, all experiments were carried out on a computer with a 3.20 GHz Intel i7 processor, 24.00 GB memory, NVIDIA 1070 Ti, and Windows 10 operating system. The programming environment was MATLAB R2022a;	
	
	 	\begin{table*}[t!]
		\footnotesize
		\caption{The average precision of various models under different
			IOU thresholds and object size.\\
			Note: The detention source of YOLOV3 is only customized in EYNet*.}
		\label{TABLE:res1}
		\renewcommand{\arraystretch}{1.5}
		\scalebox{1}{
			\begin{tabular*}{\textwidth}{@{\extracolsep{\fill}}l@{}c@{}c@{}c@{}c@{}c@{}c}
				\cline{1-7}
				Method&$AP$&$AP_{50}$&$AP_{75}$&$AP^S_{50}(54\%)$&$AP^M_{50}(34\%)$&$AP^L_{50}(12\%)$\\
				\cline{1-7}			
				\textbf{Two stage methods}&&&&&&\\								
				Faster R-CNN&0.02&0.06&0&0.01&0.12&0.16\\	
				Faster R-CNN($Epoch=100$)&0.15&0.38&0.08&0.19&0.63&0.55\\					
				\cline{1-7}			
				\textbf{One stage methods}&&&&&&\\							
				SSD&0.13&0.29&0.09&0&0.59&0.81\\	
				YOLO V2&0.22&0.58&0.14&0.23&0.89&0.92\\	
				YOLO V3&0.18&0.53&0.11&0.27&0.80&0.77\\																			
				YOLO V4 &0.20&0.53&0.09&0.31&0.77&0.81\\
				EYNet*&0.24&0.60&0.16&0.32&0.91&0.85\\
				EYNet &0.37&0.76&0.36&0.60&0.93&0.92\\						
				\cline{1-7}			
		\end{tabular*}}
	\end{table*}

	\begin{table*}[t!]
		\footnotesize
		\caption{The performance comparison of state-of-the-art methods under different criteria .\\
			Note: The detention source of YOLOV3 is only customized in EYNet*. Also, IOU thresholds are considered as $0.5$ for whole experiments.}
		\label{TABLE:res2}
		\renewcommand{\arraystretch}{1.5}
		\scalebox{1}{
			\begin{tabular*}{\textwidth}{@{\extracolsep{\fill}}l@{}c@{}c@{}c@{}c@{}c@{}c}
				\cline{1-7}
				Method&F1-Score(\%)&Recall(\%)&Precision(\%)&Accuracy(\%)&LAMR&Time(Sec)\\
				\cline{1-7}
				\textbf{Two stage methods}&&&&&&\\							
				Faster R-CNN&22.35&14.28&51.35&13.03&0.9&0.12\\									
				Faster R-CNN($Epoch=100$)&58.34&46.86&77.27&45.61&0.7&0.12\\
				\cline{1-7}
				\textbf{One stage methods}&&&&&&\\							
				SSD&46.58&31.57&88.73&31.57&0.7&0.36\\					
				YOLO V2&59.54&69.17&52.27&49.12&0.5&0.05\\	
				YOLO V3 &66.27&56.89&79.37&56.89&0.5&0.02\\																			
				YOLO V4&65.34&61.90&69.18&57.89&0.6&0.04\\		
				EYNet*&70.54&68.42&72.80&66.16&0.5&0.02\\								EYNet&81.92&78.94&85.13&78.19&0.3&0.03\\								
				\cline{1-7}
		\end{tabular*}}
	\end{table*}

	\subsection{Evaluation criteria}		
	This work employs several metrics to evaluate the performance and effectiveness of EYNet in the airport detection. These index includes Precision($P$), Recall($R$), F-Score($F$), Accuracy($Acc$), Average Precision($AP$), and Log Average Miss Rate($LAMR$). Besides, Precision against Recall, and Miss-Rate($MR$) against False Positive Per Image($FPPI$) graphs are demonstrated and compared with different state-of-the-art models.
	
	In this way, the precision and recall are defined by Equations \ref{eq:pre} and \ref{eq:rec}:
	
	\begin{equation}
		P = \frac{TP}{TP+FP}
		\label{eq:pre}
	\end{equation}
	
	\begin{equation}
		R = \frac{TP}{TP+FN}
		\label{eq:rec}
	\end{equation}	
	where $TP$ and $FP$ express the number of object which have $IOU \geq t$ and $IOU < t$, respectively. In this comparison, $t$ (usually set by 0.5) is the threshold term that plays a significant role in performance evaluation. Also, $FN$ indicates the number of the presented object by ground truth, which the model fails to detect. Notice that $TN$ is a meaningless metric that demonstrates any parts of the image where the model marked them as background.
		
In this way, $IOU$ (Intersection Over Union) is a measure to evaluate the object location accuracy. It is calculated between the detection result of the model and the matched ground-truth box. Generally, $IOU$ is calculated based on the number of pixels in the intersection and merged regions between the results and grand-truth. Obviously, the higher $IOU$ guarantees the algorithm's performance.
	
	Meanwhile, the average precision is reported to challenge the detector's performance. It is defined by Equation \ref{eq:ap}:
	
	\begin{equation}
		AP = \int_0^1 P(R)dR
		\label{eq:ap}
	\end{equation}
	where $P(R)$ is the area under curve composed of $P$ and $R$. 
	
	The other quantitative metrics, including F-Score and Accuracy, are introduced to evaluate the experimental method further. For this aim, F-Score and Accuracy are estimated by Equations \ref{eq:dr} and \ref{eq:acc}, respectively:
	\begin{equation}
		F = 2 \times \frac{P \times R}{P+R}
		\label{eq:dr}
	\end{equation}
	
	\begin{equation}
		ACC = \frac{TP}{TP+FP+FN}
		\label{eq:acc}
	\end{equation}
	Generally, F-Score is a weighted harmonic average by integrating precision and recall.
	
	 \begin{figure*}[t!]
		\center
		\setlength{\tabcolsep}{2pt}
		\begin{tabular}{cc}
			\includegraphics[width=0.49\textwidth]{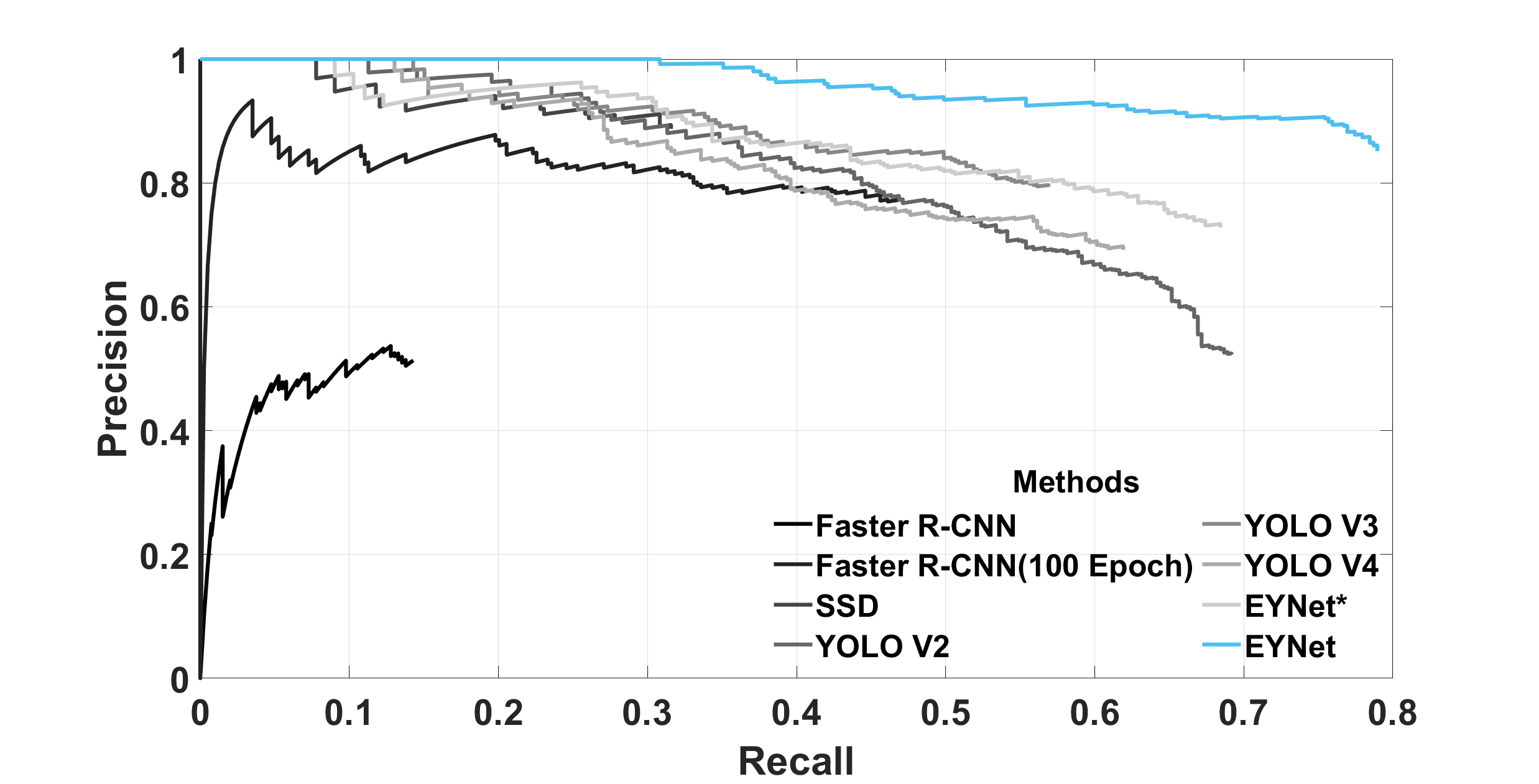} &
			\includegraphics[width=0.49\textwidth]{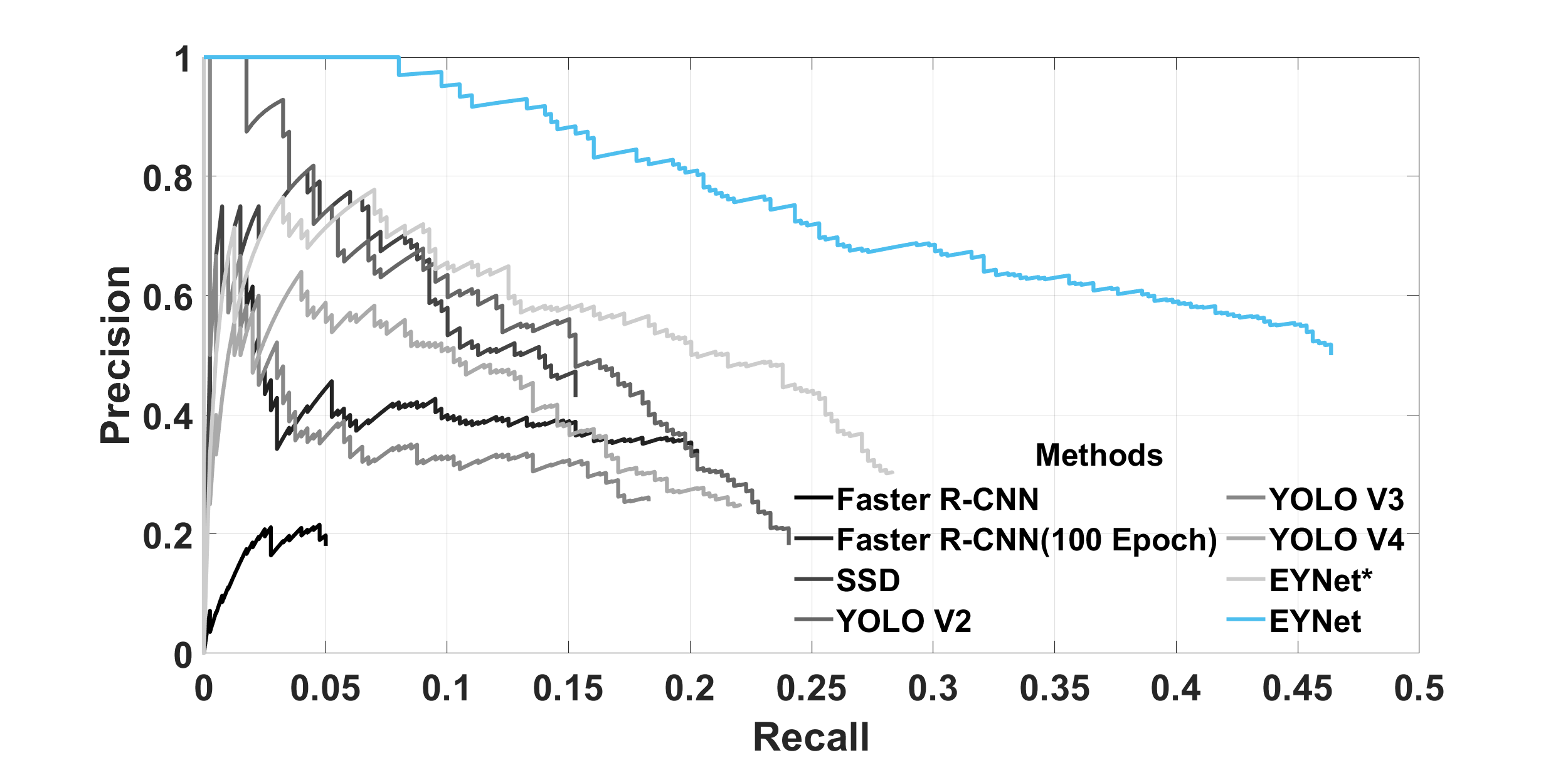} \\
			(a)&(b)\\
			\includegraphics[width=0.49\textwidth]{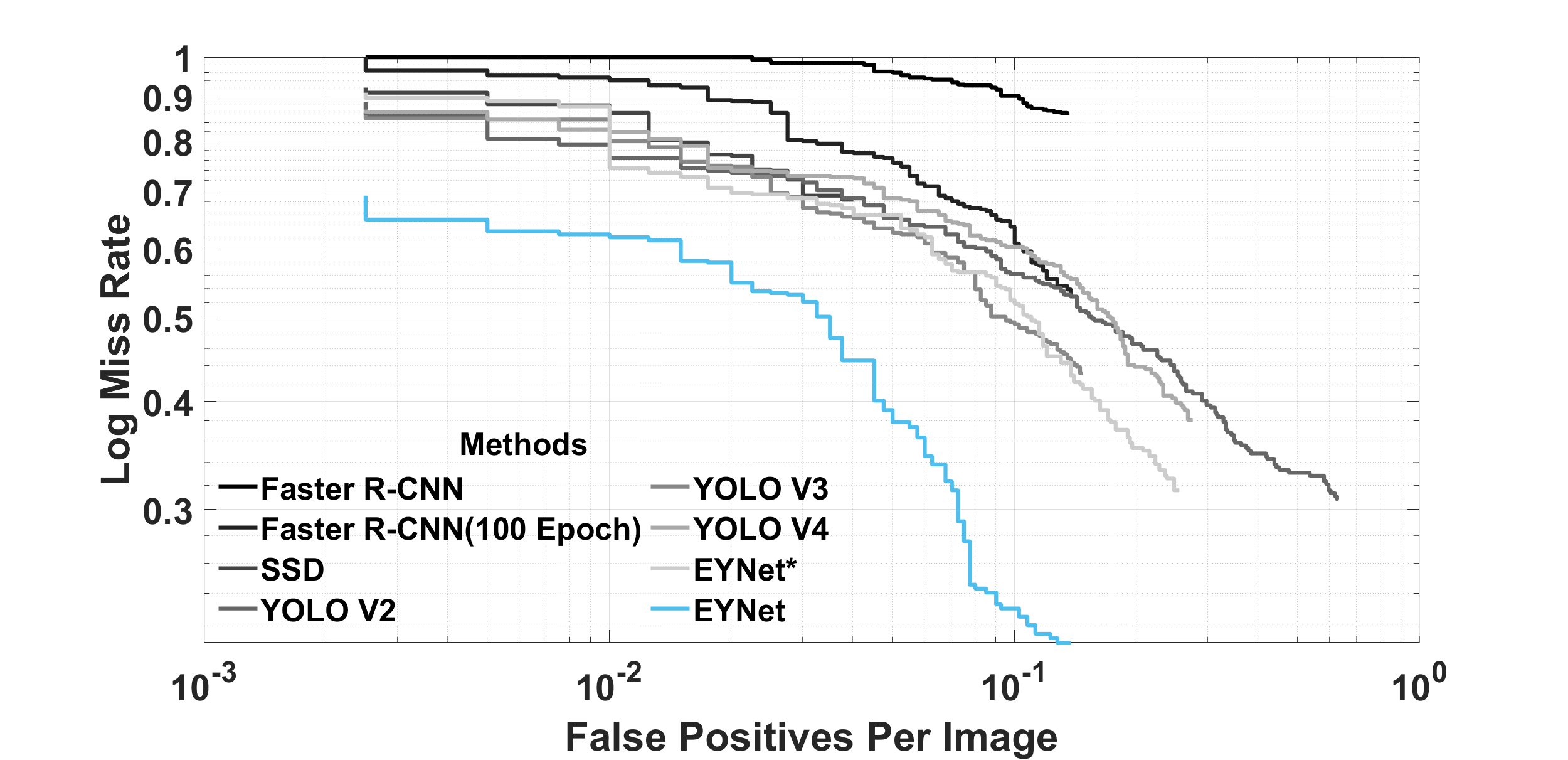} &
			\includegraphics[width=0.49\textwidth]{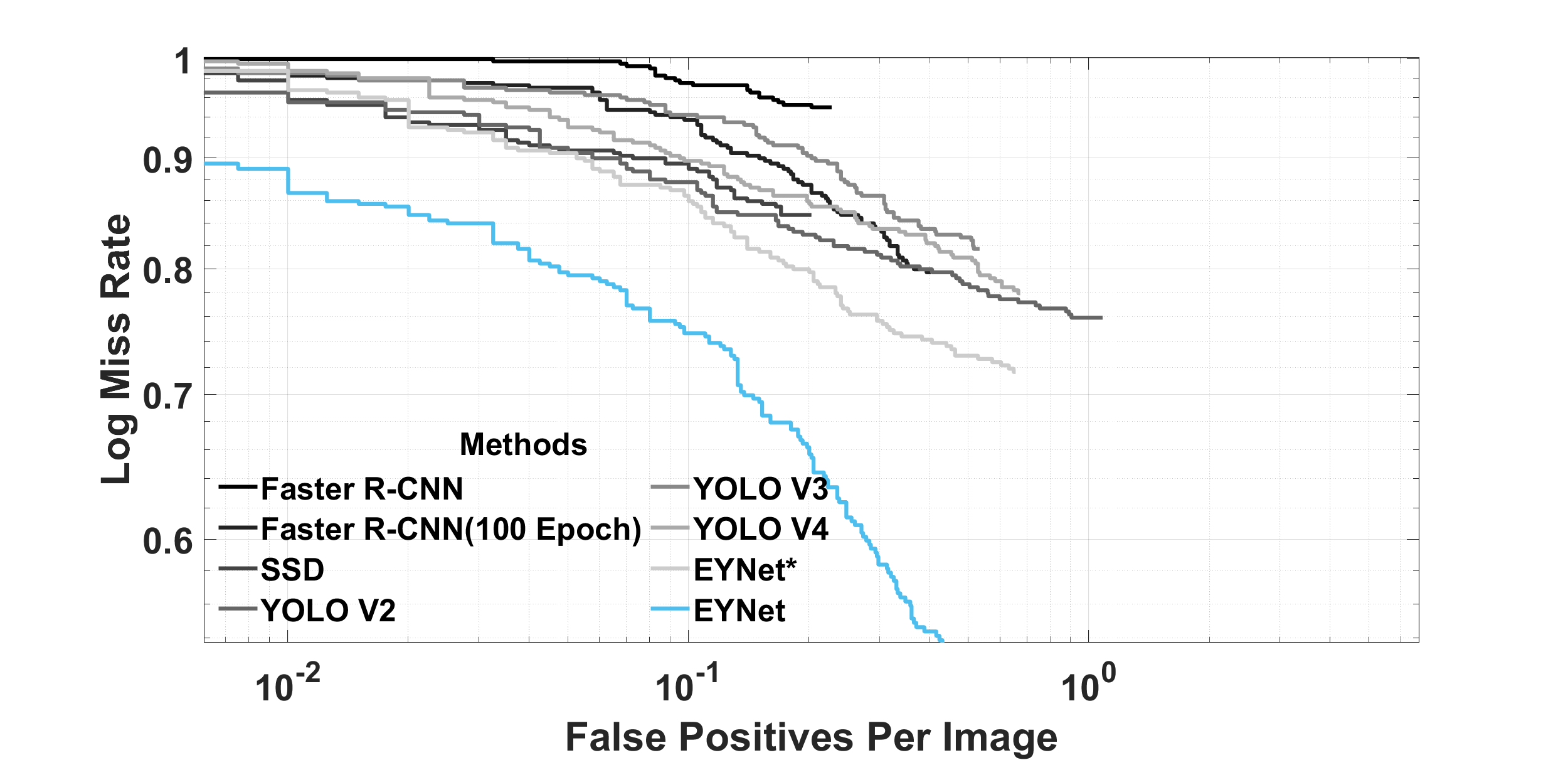} \\
			(c)&(d)\\
		\end{tabular}
		\caption{The performance comparison of state-of-the-art methods. (a, c) $IOU$ set to $0.5$, (b, d) $IOU$ set to $0.75$.}
		\label{fig:compare1}
	\end{figure*}

	In addition, the log of Average Miss-Rate is presented based in Miss-Rate and False Positive Per Image. These metrics are calculated based on Equations \ref{eq:missrate} and \ref{eq:fpr}, respectively:
	\begin{equation}
		MR = \frac{FN}{FN+TP} = 1 - R
		\label{eq:missrate}
	\end{equation}
	
	\begin{equation}
		FPPI = \frac{FP}{FP+TP} = 1 - P
		\label{eq:fpr}
	\end{equation}
	where $MR=0$ and $FPPI=0$ describe all object are detected and detected objects are correctly classified, respectively. The Miss-Rate against False Positive Per Image will be plot to prove the performance of scheme. Also, the Log-Average Miss Rate is calculated by averaging miss rate at nine $FPPI$ rates evenly spaced in log-space in the range between $10^{-2}$ and $10^0$ by Equation \ref{eq:LAMR}:	
	\begin{equation}
		LAMR =  exp\Bigg(\frac{1}{9}\sum_{f}log\bigg(MR\Big(\operatorname*{arg\,max}_{fppi(t)\leq f} \: FPPI(t)\Big)\bigg)\Bigg)
		\label{eq:LAMR}
	\end{equation}
	where detentions are taken into account which have $IOU \geq t$. Generally, $LAMR$ is a real number which summarizes the whole $MR$ against $FPPI$ curve for facilitate comparison. In detail, for each $FPPI$ as reference point the corresponding $MR$ is utilized. Also, the highest $FPPI$ is selected as new reference point when a miss-rate value for a given $f$ is absence. In sum up, decreasing $t$ leads more detection to be considered in evaluation step. 
	

	\subsection{Hyper-Parameters Settings}
	In this subsection, the details of hyper-parameters tuning are explained. These settings are distinguished for augmentation step, transfer learning, and main core phases.
	
	First, the details and types of augmentation employed in both transfer learning and EYNet phases are reported in Table \ref{TABLE:augmentation}. It should be noted that in some types of augmentation, not only is the performance not extended, they have a negative effect on the final results. Hence, this research selects beneficial and crucial augmentation types to overcome the poor samples and over-fitting problems. As can be seen, some operations are identical for both transfer learning and the main core of the proposed phases.
	
	On the other hand, the basic hyper-parameters tuning of networks are listed in Tables \ref{TABLE:hyperparameterstrans} and \ref{TABLE:hyperparameters}. These parameters are effectively specified by several experiments. 
	
	As mentioned before, the two pre-trained models, MobileNet and ResNet18 are candidated as the base networks for EYNet. The hyper-parameters setting of this phase are illustrated in Tables \ref{TABLE:hyperparameterstrans}. This phase employs a Stochastic Gradient Descent with Momentum (sgdm) solver for optimizing the network. Another noteworthy point is ShearLet filters which insert as the first convolution layers of ResNet18.
	
	Furthermore, Tables \ref{TABLE:hyperparameters} illustrate the details of EYNet hyper-parameters setting. Contrary to the transfer learning phase, the RMSprop optimizer is selected in EYNet. Also, $10\%$ of whole data are considered as a validation set for hyper-parameter setting. The epoch number is regarded as a stop criterion of the training phase. The Huber loss is adopted instead of MSE in computing offset box loss. Due to hardware limitations, this research has to decrease the mini-batch and input size of the image to $8$ and $[224\times224]$, respectively. Despite these constraints, EYNet reaches an admissible and comparable result that will be discussed in the following.
  
	\subsection{Performance evaluation of EYNet}	
	To evaluate the performance of EYNet, several criteria and analyses are reported in this subsection. To do so, first, the efficiency of the transfer learning phase and also the confusion matrix are studied in detail. Second, the various index and the loss of the main core of EYNet are demonstrated. Finally, the performance of EYNet in terms of airport detection is visually presented.

	Figure \ref{fig:transferlearing}. illustrates the accuracy and loss of both MobileNetV2 and ResNet18 networks in the transfer learning phase. It is quite evident that the candidate shallower networks can effectively reach admissible results. In other words, it is an appropriate starting point for the main core of EYNet. The stopping criteria for classification phase is validation patience. In addition, the confusion matrix of validation data for both networks is indicated in Figure. \ref{fig:transferlearing2}. As can be seen, similar classes can be significantly distinguished with the help of these models. As mentioned before, the ShearLet filters are inserted at the beginning of ResNet18. Hence, this model concentrates on high-frequency input coefficients during the training stage. On the other hand, the color image is provided to MobileNetV2 to process input in RGB space. Due to the low parameter and superior results of these networks, EYNet can be utilized in devices with poor hardware.
	 
	 In the following, the performance of EYNet during the training phase is analyzed in Figure. \ref{fig:valperf}. To do so, loss of train and validation data are plotted in Figure. \ref{fig:valperf} (a). The number of epochs is the main criteria for stopping the training stage. It can be observed, the network cannot improve the loss value after nearly 20 epochs. On the other hand, Figure. \ref{fig:valperf} (b) illustrated the several quality measures on validation data during the training stage. These quantitative evaluations indicate the performance of EYNet in terms of various essential aspects. Although the loss value reaches a stable range after 20 epochs, the quality metrics like Average Precision have been boosted after the noted epoch. Generally, EYNet is effective in the feature extraction and detection stage. The convolution kernels, especially ShearLet ones, are more efficient in extracting the features which lead airport and non-airport regions to be discriminated.
 	 
	The airport detection results for the remote sensing images are visually presented in Figure. \ref{fig:airportdetectionvisual}. Different type of airports is one of the main challenges of DIOR dataset. Some airports have prominent features, which can be marked effortlessly. However, small ones are challenging due to the scale problem. Despite this, the results proved that EYNet could effectively detect airports with different shapes, such as cross-track, single and parallel double tracks, and complex multi-track forms. Briefly, EYNet can effectively locate the airports in high spatial resolution remote sensing images.
	
	\subsection{Comparing to prior arts}
	In the end, the performance comparisons between EYNet and the rest state-of-the-art models, such as Faster R-CNN, SSD, and YOLO versions, are discussed. For this aim, the evaluation metrics including Average Precision, F-Score, Precision, Recall, Accuracy, LAMR, and detection time per image are reported. Moreover, Precision vs. Recall and Log Miss Rate vs. False Positive Per Image graphs are demonstrated for candidate models to prove the superiority of the proposed scheme. In this way, all implement details, including input data size, mini-batch, epochs number, base network, optimizer algorithm, and other hyper-parameter, are similar to EYNet for fair comparisons.
	
	Table \ref{TABLE:res1} lists Average Precision index under various $IOU$ for all mentioned models. In this table, $AP$ is average precision for $IOU$ in range $0.5$ to $0.95$. Same as the rest object detection benchmark, the airports are categorized as small, medium, and large based on their area. From the table, it can be observed that EYNet exceeds all other methods via AP criteria under the different thresholds. It should be noted that Faster R-CNN cannot reach admissible results after 30 epochs. So, the training process has been continued in 100 epochs. Moreover,  EYNet(*) is presented to prove the proposed detection structure at the top of network (instead of using the original YOLO). The rest implementation details are the same as YOLOV3 in this case. 
	
	Meanwhile, the other quantitative index is given in Table \ref{TABLE:res2}. In this experiment, the threshold of $IOU$ is set to $0.5$, which means the larger $IOU$ than the threshold is considered as a true positive. Generally, more detection objects are taken into account and enhance the false positive by decreasing the threshold. In summary, the experimental results confirm that EYNet reaches a better performance than other methods due to presenting ShearLet filters, novel augmentation, hard example mining, loss function, transfer learning, detection structure, etc. 
	
	In order to further illustrate the detection effect of EYNet, two graphs, including $PR$ and $LAMR$ under different $IOU$ thresholds, are pretested in Figure. \ref{fig:compare1}. It is quite evident that the curves of EYNet have absolute superiority compared to the rest model in whole cases. Not only EYNet overcome the other models, but also EYNet(*) reached admissible results compared to the rest models.
	
	Basically, the achieved results proved the effectiveness and significance of the constructed end-to-end scheme named EYNet. These results demonstrate strong robustness and adaptation of EYNet to different airports shapes and sizes with complicated backgrounds and similarity interference.
	\section{Conclusion and Future Works}
	\label{sec:Conclusion}
	In the last decades, with the rapid development of neural networks, object detection in remote sensing images, particularly airports, became an exciting topic in computer vision science. In this study, a practical scheme based on transfer learning, extending the YOLOV3, and ShearLet transform was proposed to address the weakness of previous methods.
	
	Generally, the original YOLO cannot reach acceptable precision when dealing with small airports, especially in complex backgrounds. Hence, an effective modification is applied to the detection sub-network of YOLOV3. Due to the visual attention mechanism based on Shear filters, as the first convolution layers of the network, the computation of the detection could be greatly reduced. In other words, this strategy enables models to focus on the critical regions in the training phase, which can significantly accelerate the convergence rate and boost the accuracy. Moreover, mining hard/negative examples and a novel augmentation strategy promote efficiency in airport detection. Also, with the help of transfer learning on similar samples, the model could recognize the airports in high accuracy and satisfactory time.
 	
	The experimental results on authentic images demonstrated that the framework increased the average precision and effectively performed in complex areas. In other words, the proposed scheme could accurately yield remarkable results in terms of average precision compared to the rest of popular object detection models.
	
	The proposed scheme will be extended to other objects such as airplanes as future work. Due to the limitations of complex conditions, there are still some rooms to improve the efficiency of the detection rate. Hence, a new or modified loss function and a novel post-processing step, filtering candidates bounding box, should be considered. Also, to encourage future works, the MATLAB source code of EYNet is available in \href{http://b.bolourian.student.um.ac.ir/}{(Link)}.
\bibliography{bibliography}

\begin{thebibliography}{10}
\expandafter\ifx\csname url\endcsname\relax
  \def\url#1{\texttt{#1}}\fi
\expandafter\ifx\csname urlprefix\endcsname\relax\def\urlprefix{URL }\fi
\expandafter\ifx\csname href\endcsname\relax
  \def\href#1#2{#2} \def\path#1{#1}\fi

\bibitem{ref1}
P.~Wang, X.~Sun, W.~Diao, K.~Fu, Fmssd: Feature-merged single-shot detection
  for multiscale objects in large-scale remote sensing imagery, IEEE
  Transactions on Geoscience and Remote Sensing 58~(5) (2020) 3377--3390.
\newblock \href {https://doi.org/10.1109/TGRS.2019.2954328}
  {\path{doi:10.1109/TGRS.2019.2954328}}.

\bibitem{ref2}
G.~Zhang, S.~Lu, W.~Zhang, Cad-net: A context-aware detection network for
  objects in remote sensing imagery, IEEE Transactions on Geoscience and Remote
  Sensing 57~(12) (2019) 10015--10024.
\newblock \href {https://doi.org/10.1109/TGRS.2019.2930982}
  {\path{doi:10.1109/TGRS.2019.2930982}}.

\bibitem{ref3}
X.~Lu, Y.~Zhang, Y.~Yuan, Y.~Feng, Gated and axis-concentrated localization
  network for remote sensing object detection, IEEE Transactions on Geoscience
  and Remote Sensing 58~(1) (2020) 179--192.
\newblock \href {https://doi.org/10.1109/TGRS.2019.2935177}
  {\path{doi:10.1109/TGRS.2019.2935177}}.

\bibitem{ref4}
Y.~Hu, X.~Li, N.~Zhou, L.~Yang, L.~Peng, S.~Xiao, A sample update-based
  convolutional neural network framework for object detection in large-area
  remote sensing images, IEEE Geoscience and Remote Sensing Letters 16~(6)
  (2019) 947--951.
\newblock \href {https://doi.org/10.1109/LGRS.2018.2889247}
  {\path{doi:10.1109/LGRS.2018.2889247}}.

\bibitem{ref5}
K.~Li, G.~Wan, G.~Cheng, L.~Meng, J.~Han,
  \href{https://www.sciencedirect.com/science/article/pii/S0924271619302825}{Object
  detection in optical remote sensing images: A survey and a new benchmark},
  ISPRS Journal of Photogrammetry and Remote Sensing 159 (2020) 296--307.
\newblock \href
  {https://doi.org/https://doi.org/10.1016/j.isprsjprs.2019.11.023}
  {\path{doi:https://doi.org/10.1016/j.isprsjprs.2019.11.023}}.
\newline\urlprefix\url{https://www.sciencedirect.com/science/article/pii/S0924271619302825}

\bibitem{ref6}
B.~Hou, Z.~Ren, W.~Zhao, Q.~Wu, L.~Jiao, Object detection in high-resolution
  panchromatic images using deep models and spatial template matching, IEEE
  Transactions on Geoscience and Remote Sensing 58~(2) (2020) 956--970.
\newblock \href {https://doi.org/10.1109/TGRS.2019.2942103}
  {\path{doi:10.1109/TGRS.2019.2942103}}.

\bibitem{ref7}
X.~Hua, X.~Wang, T.~Rui, H.~Zhang, D.~Wang,
  \href{https://www.sciencedirect.com/science/article/pii/S1568494620304348}{A
  fast self-attention cascaded network for object detection in large scene
  remote sensing images}, Applied Soft Computing 94 (2020) 106495.
\newblock \href {https://doi.org/https://doi.org/10.1016/j.asoc.2020.106495}
  {\path{doi:https://doi.org/10.1016/j.asoc.2020.106495}}.
\newline\urlprefix\url{https://www.sciencedirect.com/science/article/pii/S1568494620304348}

\bibitem{ref8}
Y.~Xu, M.~Zhu, S.~Li, H.~Feng, S.~Ma, J.~Che, End-to-end airport detection in
  remote sensing images combining cascade region proposal networks and
  multi-threshold detection networks, Remote Sensing 10~(10) (2018) 1516.

\bibitem{ref10}
G.~Cheng, J.~Han,
  \href{https://www.sciencedirect.com/science/article/pii/S0924271616300144}{A
  survey on object detection in optical remote sensing images}, ISPRS Journal
  of Photogrammetry and Remote Sensing 117 (2016) 11--28.
\newblock \href
  {https://doi.org/https://doi.org/10.1016/j.isprsjprs.2016.03.014}
  {\path{doi:https://doi.org/10.1016/j.isprsjprs.2016.03.014}}.
\newline\urlprefix\url{https://www.sciencedirect.com/science/article/pii/S0924271616300144}

\bibitem{ref11}
N.~Mohammadi, M.~M. Doyley, M.~Cetin, Ultrasound elasticity imaging using
  physics-based models and learning-based plug-and-play priors, in: ICASSP 2021
  - 2021 IEEE International Conference on Acoustics, Speech and Signal
  Processing (ICASSP), 2021, pp. 1165--1169.
\newblock \href {https://doi.org/10.1109/ICASSP39728.2021.9413652}
  {\path{doi:10.1109/ICASSP39728.2021.9413652}}.

\bibitem{ref12}
R.~Girshick, J.~Donahue, T.~Darrell, J.~Malik, Rich feature hierarchies for
  accurate object detection and semantic segmentation, in: Proceedings of the
  IEEE conference on computer vision and pattern recognition, 2014, pp.
  580--587.

\bibitem{ref13}
K.~He, X.~Zhang, S.~Ren, J.~Sun, Spatial pyramid pooling in deep convolutional
  networks for visual recognition, IEEE Transactions on Pattern Analysis and
  Machine Intelligence 37~(9) (2015) 1904--1916.
\newblock \href {https://doi.org/10.1109/TPAMI.2015.2389824}
  {\path{doi:10.1109/TPAMI.2015.2389824}}.

\bibitem{ref14}
R.~Girshick, Fast r-cnn, in: Proceedings of the IEEE international conference
  on computer vision, 2015, pp. 1440--1448.

\bibitem{ref15}
S.~Ren, K.~He, R.~Girshick, J.~Sun, Faster r-cnn: Towards real-time object
  detection with region proposal networks, Advances in neural information
  processing systems 28 (2015).

\bibitem{ref16}
J.~Redmon, S.~Divvala, R.~Girshick, A.~Farhadi, You only look once: Unified,
  real-time object detection, in: Proceedings of the IEEE conference on
  computer vision and pattern recognition, 2016, pp. 779--788.

\bibitem{ref17}
W.~Liu, D.~Anguelov, D.~Erhan, C.~Szegedy, S.~Reed, C.-Y. Fu, A.~C. Berg, Ssd:
  Single shot multibox detector, in: European conference on computer vision,
  Springer, 2016, pp. 21--37.

\bibitem{ref18}
B.~Cai, Z.~Jiang, H.~Zhang, Y.~Yao, J.~Huang, Training deep convolution neural
  network with hard example mining for airport detection, in: 2017 IEEE
  International Geoscience and Remote Sensing Symposium (IGARSS), 2017, pp.
  862--865.
\newblock \href {https://doi.org/10.1109/IGARSS.2017.8127089}
  {\path{doi:10.1109/IGARSS.2017.8127089}}.

\bibitem{ref19}
S.~Li, Y.~Xu, M.~Zhu, S.~Ma, H.~Tang, Remote sensing airport detection based on
  end-to-end deep transferable convolutional neural networks, IEEE Geoscience
  and Remote Sensing Letters 16~(10) (2019) 1640--1644.
\newblock \href {https://doi.org/10.1109/LGRS.2019.2904076}
  {\path{doi:10.1109/LGRS.2019.2904076}}.

\bibitem{ref20}
S.~Bhagavathy, B.~S. Manjunath, Modeling and detection of geospatial objects
  using texture motifs, IEEE Transactions on Geoscience and Remote Sensing
  44~(12) (2006) 3706--3715.
\newblock \href {https://doi.org/10.1109/TGRS.2006.881741}
  {\path{doi:10.1109/TGRS.2006.881741}}.

\bibitem{ref21}
G.~Tang, Z.~Xiao, Q.~Liu, H.~Liu, A novel airport detection method via line
  segment classification and texture classification, IEEE Geoscience and Remote
  Sensing Letters 12~(12) (2015) 2408--2412.
\newblock \href {https://doi.org/10.1109/LGRS.2015.2479681}
  {\path{doi:10.1109/LGRS.2015.2479681}}.

\bibitem{ref22}
m.~Budak, U.~Halıcı, A.~Şengür, M.~Karabatak, Y.~Xiao, Efficient airport
  detection using line segment detector and fisher vector representation, IEEE
  Geoscience and Remote Sensing Letters 13~(8) (2016) 1079--1083.
\newblock \href {https://doi.org/10.1109/LGRS.2016.2565706}
  {\path{doi:10.1109/LGRS.2016.2565706}}.

\bibitem{ref23}
C.~Tao, Y.~Tan, H.~Cai, J.~Tian, Airport detection from large ikonos images
  using clustered sift keypoints and region information, IEEE Geoscience and
  Remote Sensing Letters 8~(1) (2011) 128--132.
\newblock \href {https://doi.org/10.1109/LGRS.2010.2051792}
  {\path{doi:10.1109/LGRS.2010.2051792}}.

\bibitem{ref24}
D.~Zhu, B.~Wang, L.~Zhang, Airport target detection in remote sensing images: A
  new method based on two-way saliency, IEEE Geoscience and Remote Sensing
  Letters 12~(5) (2015) 1096--1100.
\newblock \href {https://doi.org/10.1109/LGRS.2014.2384051}
  {\path{doi:10.1109/LGRS.2014.2384051}}.

\bibitem{ref25}
X.~Yao, J.~Han, L.~Guo, S.~Bu, Z.~Liu,
  \href{https://www.sciencedirect.com/science/article/pii/S0925231215002623}{A
  coarse-to-fine model for airport detection from remote sensing images using
  target-oriented visual saliency and crf}, Neurocomputing 164 (2015) 162--172.
\newblock \href {https://doi.org/https://doi.org/10.1016/j.neucom.2015.02.073}
  {\path{doi:https://doi.org/10.1016/j.neucom.2015.02.073}}.
\newline\urlprefix\url{https://www.sciencedirect.com/science/article/pii/S0925231215002623}

\bibitem{ref26}
L.~Zhang, Y.~Zhang, Airport detection and aircraft recognition based on
  two-layer saliency model in high spatial resolution remote-sensing images,
  IEEE Journal of Selected Topics in Applied Earth Observations and Remote
  Sensing 10~(4) (2017) 1511--1524.
\newblock \href {https://doi.org/10.1109/JSTARS.2016.2620900}
  {\path{doi:10.1109/JSTARS.2016.2620900}}.

\bibitem{ref27}
D.~Zhao, Y.~Ma, Z.~Jiang, Z.~Shi, Multiresolution airport detection via
  hierarchical reinforcement learning saliency model, IEEE Journal of Selected
  Topics in Applied Earth Observations and Remote Sensing 10~(6) (2017)
  2855--2866.
\newblock \href {https://doi.org/10.1109/JSTARS.2017.2669335}
  {\path{doi:10.1109/JSTARS.2017.2669335}}.

\bibitem{ref28}
N.~Liu, Z.~Cui, Z.~Cao, Y.~Pi, S.~Dang, Airport detection in large-scale sar
  images via line segment grouping and saliency analysis, IEEE Geoscience and
  Remote Sensing Letters 15~(3) (2018) 434--438.
\newblock \href {https://doi.org/10.1109/LGRS.2018.2792421}
  {\path{doi:10.1109/LGRS.2018.2792421}}.

\bibitem{ref29}
J.~Tu, F.~Gao, J.~Sun, A.~Hussain, H.~Zhou, Airport detection in sar images via
  salient line segment detector and edge-oriented region growing, IEEE Journal
  of Selected Topics in Applied Earth Observations and Remote Sensing 14 (2021)
  314--326.
\newblock \href {https://doi.org/10.1109/JSTARS.2020.3036052}
  {\path{doi:10.1109/JSTARS.2020.3036052}}.

\bibitem{ref43}
X.~Dong, J.~Tian, Q.~Tian, A feature fusion airport detection method based on
  the whole scene multispectral remote sensing images, IEEE Journal of Selected
  Topics in Applied Earth Observations and Remote Sensing 15 (2022) 1174--1187.
\newblock \href {https://doi.org/10.1109/JSTARS.2021.3139926}
  {\path{doi:10.1109/JSTARS.2021.3139926}}.

\bibitem{ref30}
P.~Zhang, X.~Niu, Y.~Dou, F.~Xia, Airport detection on optical satellite images
  using deep convolutional neural networks, IEEE Geoscience and Remote Sensing
  Letters 14~(8) (2017) 1183--1187.
\newblock \href {https://doi.org/10.1109/LGRS.2017.2673118}
  {\path{doi:10.1109/LGRS.2017.2673118}}.

\bibitem{ref31}
B.~Cai, Z.~Jiang, H.~Zhang, D.~Zhao, Y.~Yao, Airport detection using end-to-end
  convolutional neural network with hard example mining, Remote Sensing 9~(11)
  (2017) 1198.

\bibitem{ref32}
Z.~Xiao, Y.~Gong, Y.~Long, D.~Li, X.~Wang, H.~Liu, Airport detection based on a
  multiscale fusion feature for optical remote sensing images, IEEE Geoscience
  and Remote Sensing Letters 14~(9) (2017) 1469--1473.
\newblock \href {https://doi.org/10.1109/LGRS.2017.2712638}
  {\path{doi:10.1109/LGRS.2017.2712638}}.

\bibitem{ref33}
F.~Chen, R.~Ren, T.~Van~de Voorde, W.~Xu, G.~Zhou, Y.~Zhou, Fast automatic
  airport detection in remote sensing images using convolutional neural
  networks, Remote Sensing 10~(3) (2018) 443.

\bibitem{ref34}
S.~Yin, H.~Li, L.~Teng, Airport detection based on improved faster rcnn in
  large scale remote sensing images, Sensing and Imaging 21~(1) (2020) 1--13.

\bibitem{ref35}
Y.~Zhong, Z.~Zheng, A.~Ma, X.~Lu, L.~Zhang, Color: Cycling, offline learning,
  and online representation framework for airport and airplane detection using
  gf-2 satellite images, IEEE Transactions on Geoscience and Remote Sensing
  58~(12) (2020) 8438--8449.
\newblock \href {https://doi.org/10.1109/TGRS.2020.2987907}
  {\path{doi:10.1109/TGRS.2020.2987907}}.

\bibitem{ref42}
N.~Li, L.~Cheng, L.~Huang, C.~Ji, M.~Jing, Z.~Duan, J.~Li, M.~Li, Framework for
  unknown airport detection in broad areas supported by deep learning and
  geographic analysis, IEEE Journal of Selected Topics in Applied Earth
  Observations and Remote Sensing 14 (2021) 6328--6338.
\newblock \href {https://doi.org/10.1109/JSTARS.2021.3088911}
  {\path{doi:10.1109/JSTARS.2021.3088911}}.

\bibitem{ref41}
B.~{Bolourian Haghighi}, A.~H. Taherinia, A.~Harati, M.~Rouhani,
  \href{https://www.sciencedirect.com/science/article/pii/S1568494620309686}{Wsmn:
  An optimized multipurpose blind watermarking in shearlet domain using mlp and
  nsga-ii}, Applied Soft Computing 101 (2021) 107029.
\newblock \href {https://doi.org/https://doi.org/10.1016/j.asoc.2020.107029}
  {\path{doi:https://doi.org/10.1016/j.asoc.2020.107029}}.
\newline\urlprefix\url{https://www.sciencedirect.com/science/article/pii/S1568494620309686}

\bibitem{ref38}
Matlab library developed (shearlab), \url{http://www.shearlab.org/}.

\bibitem{ref39}
G.~Kutyniok, W.-Q. Lim, R.~Reisenhofer,
  \href{http://doi.acm.org/10.1145/2740960}{Shearlab 3d: Faithful digital
  shearlet transforms based on compactly supported shearlets}, ACM Trans. Math.
  Softw. 42~(1) (2016) 5:1--5:42.
\newblock \href {https://doi.org/10.1145/2740960} {\path{doi:10.1145/2740960}}.
\newline\urlprefix\url{http://doi.acm.org/10.1145/2740960}

\bibitem{ref44}
J.~Redmon, A.~Farhadi, Yolo9000: better, faster, stronger, in: Proceedings of
  the IEEE conference on computer vision and pattern recognition, 2017, pp.
  7263--7271.

\bibitem{ref45}
J.~Redmon, A.~Farhadi, Yolov3: An incremental improvement, arXiv preprint
  arXiv:1804.02767 (2018).

\bibitem{ref36}
G.~Cheng, J.~Han, X.~Lu, Remote sensing image scene classification: Benchmark
  and state of the art, Proceedings of the IEEE 105~(10) (2017) 1865--1883.
\newblock \href {https://doi.org/10.1109/JPROC.2017.2675998}
  {\path{doi:10.1109/JPROC.2017.2675998}}.

\bibitem{ref40}
P.~J. Huber, Robust estimation of a location parameter, in: Breakthroughs in
  statistics, Springer, 1992, pp. 492--518.

\bibitem{ref37}
I.~Loshchilov, F.~Hutter, Sgdr: Stochastic gradient descent with warm restarts,
  arXiv preprint arXiv:1608.03983 (2016).

\end{thebibliography}
\end{document}